\documentclass{article} %
\usepackage{iclr2027_conference,times}
\newif\ifarxiv
\arxivtrue                 %
\ifarxiv\iclrfinalcopy\fi

\usepackage{amsmath,amsfonts,bm}

\def\eqref#1{equation~\ref{#1}}

\def\1{\bm{1}}

\DeclareMathAlphabet{\mathsfit}{\encodingdefault}{\sfdefault}{m}{sl}
\SetMathAlphabet{\mathsfit}{bold}{\encodingdefault}{\sfdefault}{bx}{n}

\usepackage{graphicx}
\usepackage{enumitem}
\usepackage{booktabs}
\usepackage{array}
\usepackage[table]{xcolor}
\usepackage{amsmath}
\usepackage{amssymb}
\usepackage{listings}
\usepackage{multirow}
\lstdefinestyle{promptappendix}{
    basicstyle=\ttfamily\scriptsize,
    breaklines=true,
    breakatwhitespace=false,
    columns=fullflexible,
    keepspaces=true,
    frame=single,
    numbers=none,
    xleftmargin=0pt,
    framexleftmargin=0pt,
    captionpos=b,
    literate={→}{{$\rightarrow$}}1 {—}{{---}}1
}
\usepackage{hyperref}
\usepackage{url}

\title{One Figure, Every Canvas:\\Editable Flowchart Relayout via Agentic Pipeline}

\author{Shih-Chen Tseng$^{1}$\thanks{Equal contribution.}\hspace{0.4em}, Chih-Hsuan Chen$^{1}$\footnotemark[1]\hspace{0.4em}, Ryan Yang$^{1,2}$\footnotemark[1]\hspace{0.4em}, \\
\textbf{Hsi-An Chen$^{1}$, Chun-Wei Tuan Mu$^{1}$, Yu-Lun Liu$^{1}$} \\
$^{1}$Department of Computer Science, National Yang Ming Chiao Tung University\\$^{2}$Siebel School of Computing and Data Science, University of Illinois Urbana-Champaign \\
}

\begin{document}

\maketitle
\ifarxiv\lhead{Preprint. Under review.}\fi

\begin{figure}[h]
    \centering
    \vspace{-3mm}
    \includegraphics[width=1.0\linewidth]{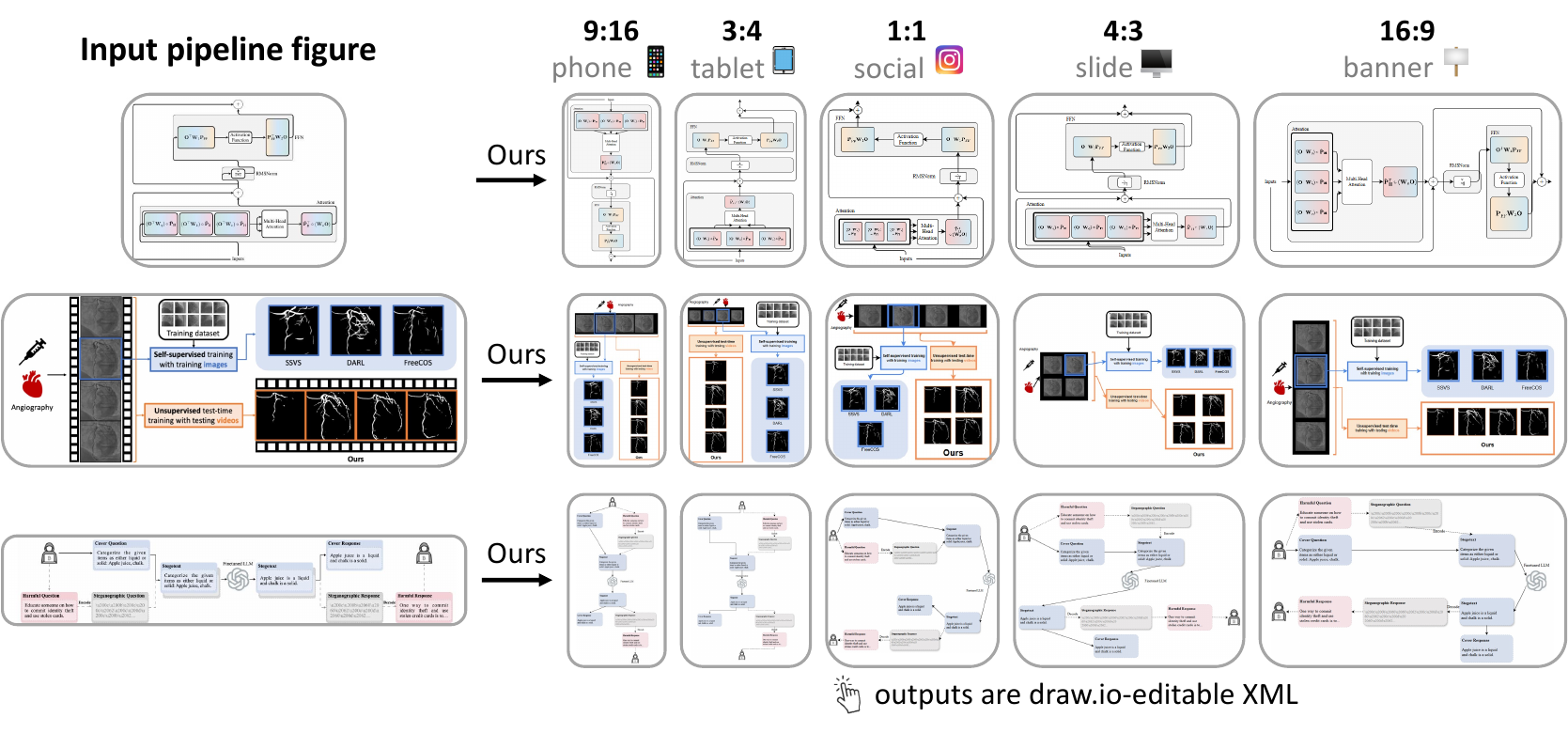}
    \vspace{-7mm}
    \caption{\textbf{Aspect-ratio-adaptive flowchart relayout.} Given a single flowchart at one fixed ratio \emph{(left)}, our method produces structure-preserving relayouts at arbitrary aspect ratios \emph{(right)}: phone, tablet, social, slide, and banner. Three different source figures (top to bottom) demonstrate generality across architecture diagrams, experimental pipelines, and process flows. Outputs are not re-rendered images but draw.io-editable XML, ready for further manual refinement.}
    \label{fig:teaser}
\end{figure}

\begin{abstract}
Pipeline figures in ML papers must be repurposed across many canvases, including paper columns, 16:9 slides, portrait posters, 1:1 social teasers, 9:16 phone previews. Each format imposes a different aspect ratio on the same computational graph, where any silently broken connection misrepresents the method. We formulate aspect-ratio-adaptive flowchart relayout as a distinct task: given a raster flowchart and a target ratio, produce a structurally faithful, hallucination-free, editable layout. Existing methods fail characteristically: image-to-image models stretch blocks and reject extreme ratios, text-to-image agentic systems hallucinate content, and parse-then-render systems mis-route edges. We propose an agentic pipeline factored into Parse, Style, and Layout stages, each pairing a main agent with a critic that combines deterministic constraint checks with VLM visual feedback so connectivity is explicitly checked and prevented from being silently broken. Outputs are draw.io-editable mxGraph XML. On a curated benchmark of 100 flowcharts at five aspect ratios, evaluated by Gemini 3.1 Pro and validated against human judgments, our method reaches 68.6\% Content Fidelity versus 11.2--41.4\% for prior work.
Project page: \url{https://onefigureeverycanvas.vercel.app/}
\end{abstract}

\begin{figure}[t]
    \centering
    \vspace{-3mm}
    \includegraphics[width=1.0\linewidth]{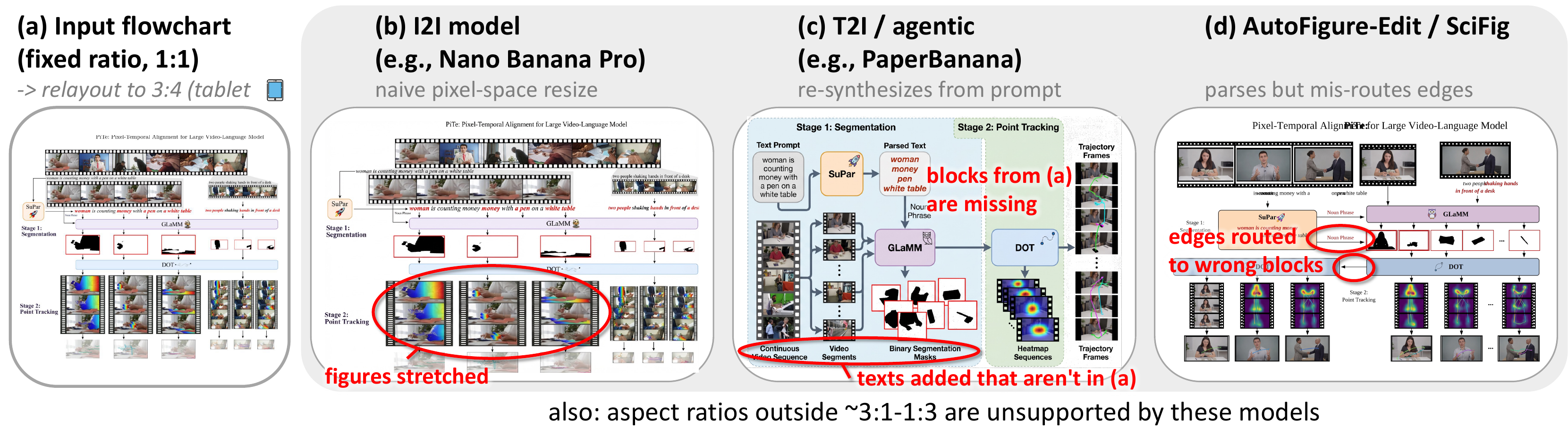}
    \vspace{-7mm}
    \caption{
    \textbf{Three categorical failure modes on flowchart relayout.} Given input (a) at 1:1 and a request for 3:2, we examine three families. (b) Image-to-image models (e.g., Nano Banana Pro) resize in pixel space, stretching embedded frames and blocks, and they also reject aspect ratios outside 3:1–1:3. (c) Text-to-image agentic systems (e.g., PaperBanana) re-synthesize from a prompt, dropping blocks present in (a) and adding labels that were not. (d) Parse-then-render systems (e.g., AutoFigure-Edit, SciFig) recover components but mis-route edges, breaking the computational graph. Red annotations mark each failure. None achieves structural fidelity, hallucination-freedom, and arbitrary resizing simultaneously. These are the three properties our task requires.
    }
    \label{fig:motivation}
    \vspace{-3mm}
\end{figure}

\vspace{-2mm}
\section{Introduction}
\vspace{-2mm}

Pipeline figures are central to how vision and ML research is communicated, yet the same diagram has to live across many canvases: paper columns, 16:9 slides, portrait posters, 1:1 social teasers, 9:16 phone previews. Each target imposes a different aspect ratio on the same computational graph, and every conversion must preserve the original blocks, edges, and flow direction. Any silently broken connection misrepresents the method. Manually re-laying out figures for each canvas is tedious and error-prone. As shown in Fig.~\ref{fig:teaser}, our goal is to produce structure-preserving relayouts at arbitrary aspect ratios from a single source figure, with each output remaining editable in draw.io for further refinement.

Recent work on automating scientific-figure creation falls into three families. Image-to-image (I2I) models such as Nano Banana Pro and GPT Image 2.0 re-synthesize the source figure at a requested resolution. Text-to-image agentic systems, PaperBanana~\citep{zhu2026paperbanana}, SciFig~\citep{huang2026scifig}, and Paper2Poster~\citep{pang2025paper2poster}, orchestrate planner, stylist, and critic agents to create a new figure from a methodology description. A third line targets editability: AutoFigure-Edit~\citep{lin2026autofigure} emits SVG, while AutomaTikZ~\citep{belouadi2023automatikz} and DeTikZify~\citep{belouadi2024detikzify} use TikZ. All target figure \emph{generation} from textual or programmatic inputs; none takes an \emph{existing} flowchart and refits it to an arbitrary canvas while keeping its computational graph intact.

When repurposed for relayout, each family exhibits a categorical failure mode (Fig.~\ref{fig:motivation}). I2I models resize in pixel space. Embedded panels and blocks are uniformly stretched, and ratios outside 3:1–1:3 are rejected. T2I agentic systems re-synthesize from a prompt, dropping source blocks and adding labels not present in the source. Parse-then-render systems recover the components but mis-route edges, breaking the directed graph that the flowchart exists to communicate. None simultaneously achieves the three properties this task requires: structural fidelity, hallucination-freedom, and arbitrary canvas resizing in a downstream-editable format. As summarized in Table~\ref{tab:qualitative}, these methods produce non-editable raster outputs, lack native flowchart-editor support, or cannot resize an existing flowchart to arbitrary canvases.

\begin{figure}[t]
    \centering
    \vspace{-3mm}
    \includegraphics[width=1.0\linewidth]{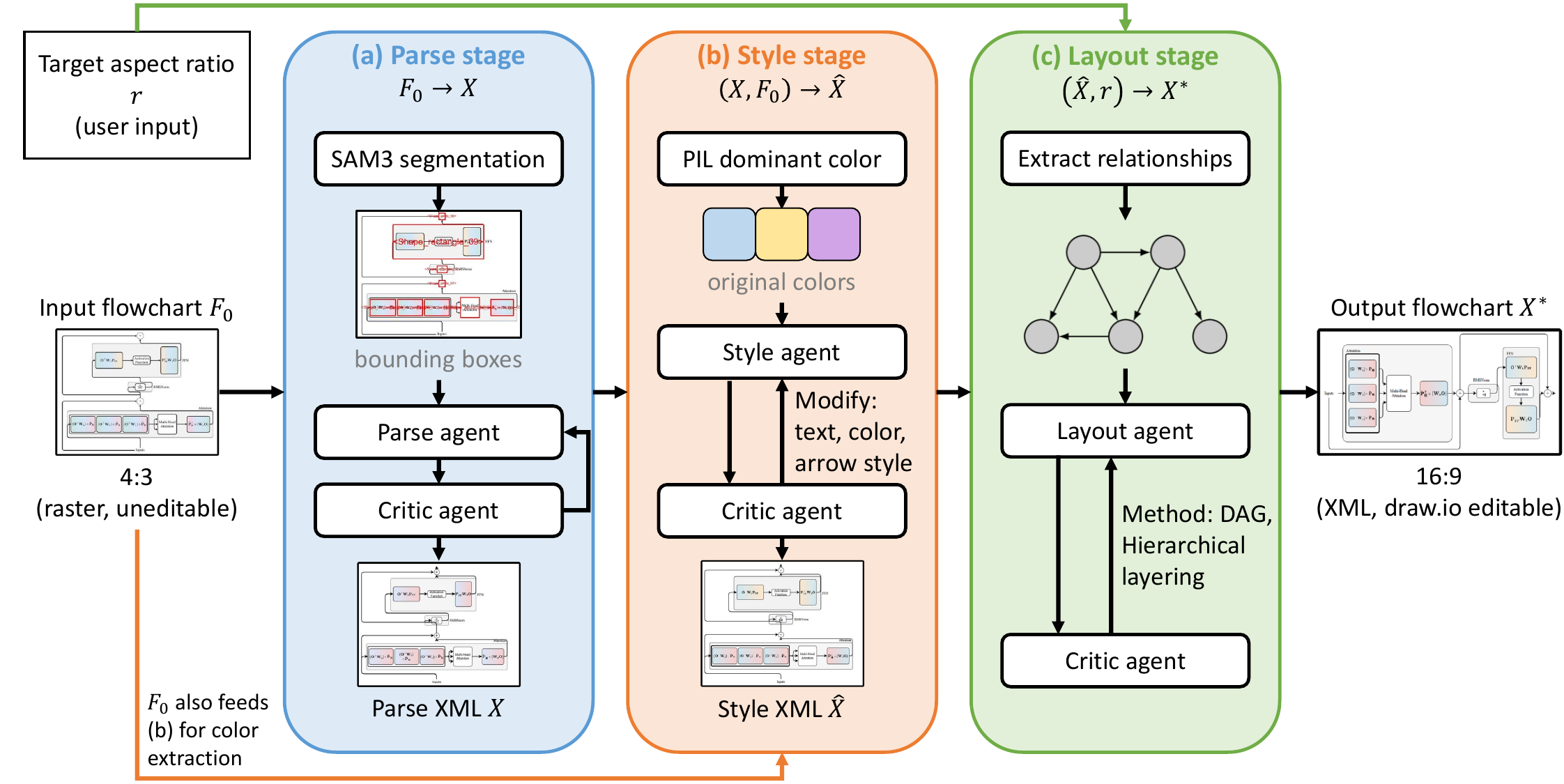}
    \vspace{-6mm}
    \caption{
    \textbf{Overview of our agentic relayout pipeline.} Given input flowchart $F_0$ (raster, here 4:3) and target aspect ratio $r$ (here 16:9), our method produces draw.io-editable XML $X^*$ in three stages, each pairing a main agent with a critic that iterates until convergence. (a) \textbf{Parse} converts $F_0$ into structured XML $X$ by combining SAM~3 instance segmentation (bounding boxes for every component) with a VLM-based parse agent. (b) \textbf{Style} produces styled XML $\hat{X}$ from $X$ and $F_0$; we extract dominant colors directly from $F_0$'s pixels with PIL, since VLM color perception is unreliable. (c) \textbf{Layout} extracts inter-block relationships from $\hat{X}$ and rearranges components into canvas $r$ via DAG-based hierarchical layering; the critic verifies no relationship is broken and no same-level components newly overlap. Output $X^*$ preserves the structure, style, and content of $F_0$ while fitting the new aspect ratio.
    }
    \label{fig:overview}
    \vspace{-3mm}
\end{figure}

We introduce an agentic pipeline for aspect-ratio-adaptive flowchart relayout. Given a raster flowchart $F_0$ and target ratio $r$, our method produces a draw.io-compatible mxGraph XML $X^*$ that preserves the structure, content, and style of $F_0$ while fitting the new canvas. Since directly resolving structure, appearance, and spatial arrangement in a single step is difficult, we decompose the problem into three stages, each pairing a main agent with a critic (Fig.~\ref{fig:overview}). The \textbf{Parse} stage recovers structured XML from $F_0$ with SAM~3~\citep{carion2025sam} spatial priors; the \textbf{Style} stage reconstructs visual appearance while extracting colors deterministically to mitigate unreliable VLM color perception; and the \textbf{Layout} stage rearranges the recovered graph for the target canvas, combining deterministic structural checks with VLM visual feedback to prevent connectivity errors. We further curate flowcharts from oral papers at top-tier venues, evaluate them at five target ratios, and propose four VLM-as-a-Judge metrics validated against human judgments. Our pipeline reaches 68.6\% Content Fidelity versus 11.2--41.4\% for prior work.

Our contributions are:
\begin{itemize}[leftmargin=*]
\vspace{-1mm}
\item We formulate \emph{aspect-ratio-adaptive flowchart relayout} as a distinct task and propose the first agentic pipeline for it, factored into Parse, Style, and Layout stages, each guarded by a critic combining VLM reasoning with deterministic verification so connectivity is explicitly checked and prevented from being silently broken.
\vspace{-1mm}
\item We construct \textsc{FlowchartRelayoutBench}, a benchmark pairing flowcharts from CVPR/ICCV/ECCV/NeurIPS/ICLR/ICML oral papers at five target ratios (9:16, 2:3, 1:1, 3:2, 16:9) with a four-metric VLM-as-a-Judge protocol (Relationship Preservation, Hallucination-free Rate, Layout Quality, Style Similarity) validated against human assessments.
\vspace{-1mm}
\item Experiments and a user study show that our pipeline substantially outperforms both monolithic VLM and existing agentic figure systems on the structural-fidelity dimensions that matter most for cross-canvas relayout.
\vspace{-1mm}
\end{itemize}

\begin{table*}[t]
\centering
\small
\vspace{-3mm}
    \caption{
    \textbf{Capability comparison with prior work.} We compare against three families of figure-generation approaches along three axes for paper-figure relayout: how style is specified, what canvas ratios are supported, and whether outputs are editable in a standard tool. Only ours achieves all three: image-reference style control, arbitrary aspect ratios, and draw.io compatibility. $\checkmark$ = full support, $\triangle$ = partial support, \texttimes = none.
    }
    \label{tab:qualitative}
    \renewcommand{\arraystretch}{0.9}
    \begin{tabular}{lccc}
    \toprule
    \textbf{Method} & \textbf{Style Control} & \textbf{Canvas Resizing} & \textbf{Editable Format} \\
    \midrule
    \textit{Image-to-image models} \\
    \quad Nano Banana Pro                      & Image reference & $\triangle$ (\textasciitilde 3:1 to 1:3) & \texttimes \\
    \quad GPT Image 2.0                      & Image reference & $\triangle$ (3:1 to 1:3)                 & \texttimes \\
    \addlinespace[2pt]
    \textit{Text-to-image / agentic systems} \\
    \quad SciFig~\citep{huang2026scifig}       & Prompt (Hard)   & \texttimes                               & \texttimes \\
    \quad PaperBanana~\citep{zhu2026paperbanana}& Prompt (Hard)  & $\triangle$ (\textasciitilde 3:1 to 1:3) & \texttimes \\
    \addlinespace[2pt]
    \textit{Parse-then-render systems} \\
    \quad AutoFigure-Edit~\citep{lin2026autofigure} & Image reference & \texttimes                          & $\triangle$ (SVG) \\
    \midrule
    \rowcolor{gray!10}
    \textbf{Ours}                             & \textbf{Image reference (Easy)} & \textbf{\checkmark\ (Arbitrary)} & \textbf{\checkmark\ (draw.io XML)} \\
    \bottomrule
    \end{tabular}%
\vspace{-3mm}
\end{table*}

\vspace{-3mm}

\section{Related Work}
\label{sec:relatedwork}
\vspace{-3mm}

\paragraph{Agentic Frameworks for Scientific Figure Generation.}
Scientific figure generation has evolved from one-shot synthesis~\citep{rodriguez2023figgen,belouadi2023automatikz,belouadi2024detikzify} to planner--critic agents for charts~\citep{dibia2023lida,yang2024matplotagent,goswami2025plotgen,han2023chartllama}, slides~\citep{zheng2025pptagent,zheng2026deeppresenter}, posters~\citep{pang2025paper2poster}, diagrams~\citep{zala2023diagrammergpt,wei2025words}, pipeline and architecture figures~\citep{zhu2026paperbanana,huang2026scifig,zhu2026autofigure,lin2026autofigure,liu2026figuresprogramsrecursivegeneration,guo2026paper2sysarch}, and layouts~\citep{lin2023parse,guo2026seeingimprovingvisualfeedback}, with dedicated benchmarks~\citep{zhang2025scimage,tang2026igenbench}. Concurrent systems improve editability through structure--style--layout decomposition~\citep{wang2026evodiagram}, PowerPoint or SVG outputs~\citep{shao2026livefigure,zhao2026crafter}, and programmatic checks~\citep{zhao2026crafter}. We similarly use multi-agent decomposition, but refit an existing raster flowchart to a target canvas while preserving structure and style.

\vspace{-3mm}
\paragraph{Structured Diagram Representation.}
Structured synthesis targets vector graphics~\citep{carlier2020deepsvg,reddy2021im2vec,rodriguez2025starvector,Zhang_2026_CVPR,cheng2026stroke,lin2026autofigure}, TikZ~\citep{belouadi2023automatikz,belouadi2024detikzify,belouadi2025tikzero}, and draw.io XML~\citep{yu2026genai}. Raster structure recovery includes parse-then-re-render methods~\citep{savva2011revision,poco2017reverse}, document/diagram understanding~\citep{kembhavi2016diagram,lee2023pix2struct,huang2022layoutlmv3,hong2024cogagent}, image-to-code benchmarks~\citep{roberts2024image2struct,yang2025chartmimic}, and flowchart-specific methods~\citep{schafer2019arrow,sun2022fr,shukla2023towards,liu2024flowmind2digital,pan2024flowlearn,singh2024flowvqa,he2025flow2code,Hu_2026_CVPR,dou2026edgeflow}. Closest to our Parse stage, concurrent work converts rasters to editable mxGraph XML~\citep{cui2025draw,su2026vcg,sivertsen2026sage}, SVG~\citep{zhao2026crafter}, or Mermaid~\citep{deka2025flowchart2mermaid}, some supporting instruction-driven edits~\citep{su2026vcg,sivertsen2026sage}. Beyond conversion, we re-lay out the recovered graph for any target aspect ratio while preserving its relationships.

\vspace{-3mm}
\paragraph{Layout Generation and Retargeting.}
Classical graph drawing includes Sugiyama layering~\citep{sugiyama1981methods,gansner1993technique}, Graphviz~\citep{gansner2000open}, ELK~\citep{domros2023eclipse}, constraint-based routing~\citep{dwyer2008dunnart,wybrow2009orthogonal}, and aspect-ratio-aware heuristics~\citep{nachmanson2007drawing}, and underpins learned layouts based on transformers~\citep{lee2020neural,gupta2021layouttransformer,jiang2023layoutformer++}, diffusion~\citep{inoue2023layoutdm,zhang2023layoutdiffusion,levi2023dlt,iwai2024layout}, LLMs~\citep{feng2023layoutgpt,lin2023layoutprompter,tang2023layoutnuwa}, and content-aware objectives~\citep{hsu2023posterlayout,li2023relation,horita2024retrieval}. Related retargeting adapts images~\citep{avidan2023seam,rubinstein2010comparative} or visualizations to different displays~\citep{hoffswell2020techniques,kim2022cicero,wu2020mobilevisfixer,kim2021automated,kim2023dupo,liu2026proteus}. These methods do not explicitly preserve directed-graph connectivity under arbitrary canvas changes. Our Layout stage combines hierarchical drawing with VLM feedback and deterministic checks for edge preservation and overlap.

\vspace{-3mm}
\paragraph{VLM-based Evaluation and Iterative Refinement.}
LLM-as-judge~\citep{liu2023g,zheng2023judging} extends to vision~\citep{chen2024mllm,lee2024prometheus,xiong2025llava}, including diagram~\citep{liang2025diagrameval} and video assessment~\citep{huang2026generative}; visual generation benchmarks include~\citep{lee2023holistic,ku2023imagenhub,jiang2024genai,huang2023t2i,xia2025chartx,xu2023chartbench,zhang-etal-2026-sciflow}. Agentic systems for figures~\citep{dibia2023lida,yang2024matplotagent,zhu2026paperbanana,huang2026scifig,pang2025paper2poster,zhu2026autofigure,zhao2026crafter} and data generation~\citep{huang2026gen} also use VLM judges, despite positional and perceptual biases~\citep{shi2025judging,ye2024justice,ma2025judging,li2025vl,yasunaga2025multimodal} and memorization~\citep{szu2025beyond}. Critic-based self-refinement~\citep{shinn2023reflexion} can be unreliable without external signals~\citep{huang2024large}, motivating execution-grounded critiques~\citep{chen2024teaching}. We thus pair VLM reasoning with deterministic edge-preservation, overlap, and canvas-containment checks.

\vspace{-3mm}
\section{Method}

\vspace{-2mm}
\subsection{Framework Overview}
\vspace{-2mm}
Given an input raster flowchart $F_0$ and a target aspect ratio $r = (W, H)$, our method produces a draw.io-compatible mxGraph XML file $X^*$ that preserves the structure, content, and style of $F_0$ while fitting the target canvas:
\begin{equation} \small
    X^* = \mathcal{F}(F_0,\, r).
\end{equation}
Since instance identity, visual appearance, and spatial arrangement are tightly coupled, direct end-to-end parametrization is ill-posed. Therefore, we decompose the transformation into three sequential stages, each handled by a main agent paired with a Critic Agent that iterates until convergence. In the \textbf{Parse Stage}, $F_0$ is converted into a structured mxGraph XML $X$ encoding every block, edge, and group. In the \textbf{Style Stage}, $X$ is refined into a styled XML $\hat{X}$ whose visual properties reflect those of $F_0$. In the \textbf{Layout Stage}, $\hat{X}$ is rearranged into the target canvas $r$, producing $X^*$ with all inter-block relationships preserved. Every intermediate representation remains in mxGraph XML and can be rendered to PNG at any stage, enabling visual verification and direct editing in draw.io. An overview is shown in Fig.~\ref{fig:overview}.

\begin{figure*}[t]
    \centering
    \vspace{-3mm}
    \includegraphics[width=1.0\linewidth]{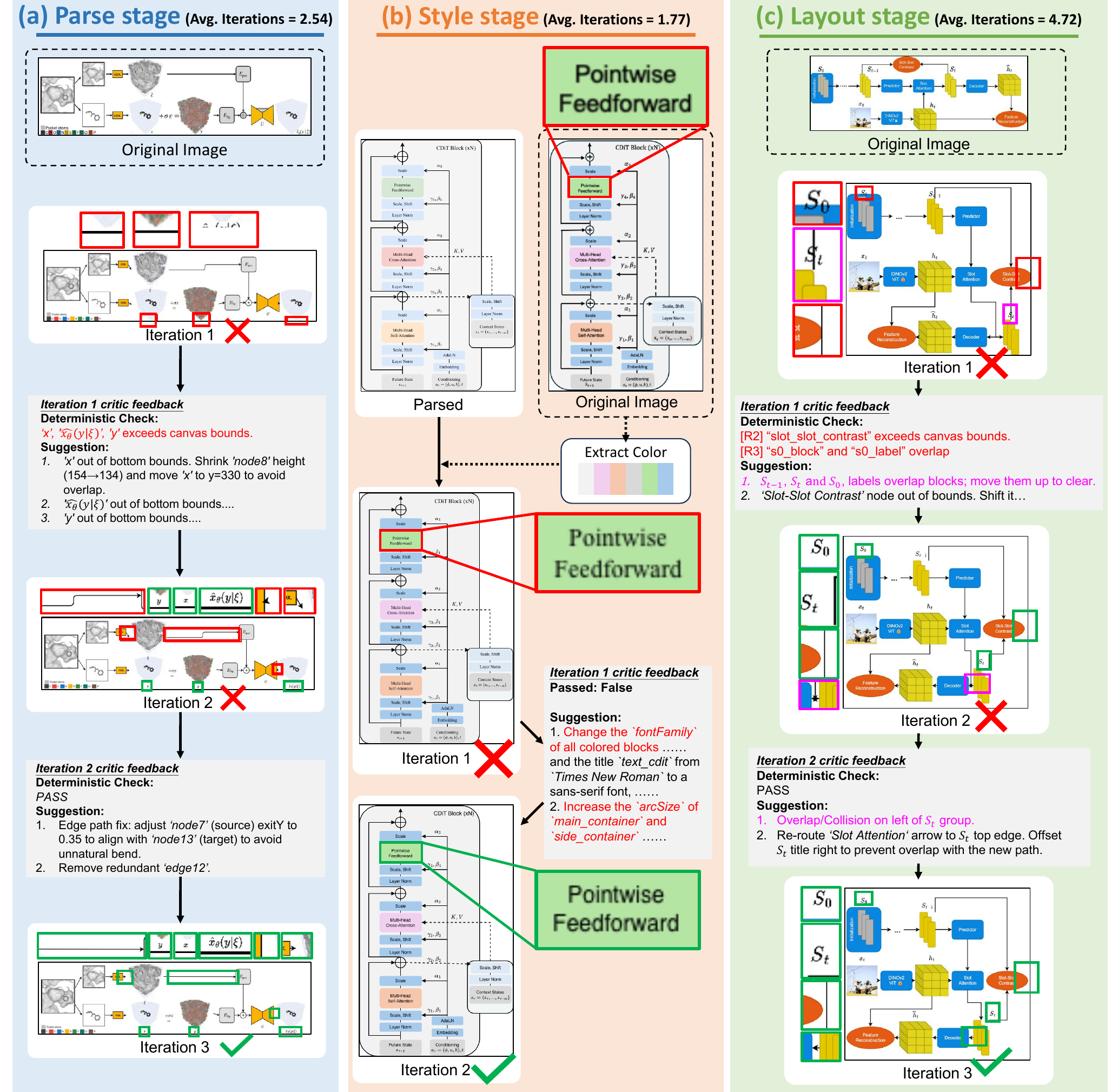}
    \vspace{-7mm}
    \caption{
    \textbf{Iterative critic refinement across pipeline stages.}
    (a) The Parse Stage combines deterministic checks and VLM feedback to progressively correct out-of-bound components and edge errors, yielding the parsed XML $X$.
    (b) The Style Stage transfers colors, typography, and corner radius, while critic feedback corrects residual font and shape mismatches.
    (c) The Layout Stage applies DAG-based hierarchical relayout and iteratively resolves boundary, overlap, and routing issues, producing the final layout $X^*$.
    }
    \vspace{-3mm}
    \label{fig:iterative_refinement}
\end{figure*}

\vspace{-3mm}
\subsection{Parse Stage}
\label{sec:parse}
\vspace{-2mm}

The Parse Stage converts the input raster flowchart $F_0$ into structured mxGraph XML $X$ encoding blocks, connectors, groups, and embedded visual assets. To reduce VLM spatial perception errors, the Parse Agent uses SAM~3~\citep{carion2025sam} to obtain instance-level bounding boxes, category priors, and spatial attributes as structured geometry context, while embedded images and icons are stored as cropped assets with XML placeholders for reinjection after the Layout Stage. After each iteration, the Critic Agent compares $F_0$, $X$, and its draw.io-rendered PNG to verify connector logic, structural consistency, and visual correspondence, returning correction instructions until the result passes or the iteration limit is reached (Fig.~\ref{fig:iterative_refinement}; Appendix~\ref{app:parse-agent} and~\ref{app:parse-critic}).

\vspace{-2mm}
\subsection{Style Stage}
\label{sec:style}
\vspace{-2mm}

The Style Stage refines $X$ into styled XML $\hat{X}$ that matches the visual appearance of $F_0$ while preserving its recovered structure and semantics. To reduce VLM color perception errors, a PIL-based Color Extraction Tool extracts per-block color assignments from $F_0$ using block positions recorded in the XML; the Style Agent then updates fill and stroke colors, text style, and arrow style while keeping block text, edge source/target references, and group membership immutable. After each iteration, the Critic Agent compares $F_0$, the rendering of $\hat{X}$, and the extracted color assignments to verify visual correspondence, returning correction instructions until the result passes or the iteration limit is reached (Fig.~\ref{fig:iterative_refinement}; Appendix~\ref{app:style-agent} and~\ref{app:style-critic}).

\vspace{-2mm}
\subsection{Layout Stage}
\label{sec:layout}
\vspace{-2mm}

The Layout Stage rearranges $\hat{X}$ onto the target canvas $r=(W,H)$ to produce $X^*$ while preserving inter-block relationships. The Layout Agent extracts edge relationships from the XML and performs directed-graph-based hierarchical relayout, first within containers and then across containers, preserving connectivity and keeping related blocks spatially close. After each iteration, the Critic Agent combines deterministic checks for overlap avoidance, relationship preservation, and canvas containment with visual comparison of the draw.io-rendered PNG and $F_0$ to identify routing, spacing, or arrangement issues, returning correction instructions until the layout passes or the iteration limit is reached (Fig.~\ref{fig:iterative_refinement}; Appendix~\ref{app:layout-agent}, \ref{app:layout-rules}, and~\ref{app:layout-critic}).

\vspace{-2mm}
\subsection{Style Transfer}
\label{sec:style_transfer}
\vspace{-2mm}
Our framework supports reference-guided style transfer as an optional extension while preserving diagram structure and semantic content; details and qualitative results are provided in Appendix~\ref{app:style-transfer}.

\vspace{-2mm}
\subsection{Editable Diagram Representation}
\label{sec:editable}
\begin{figure}[t]
    \centering
    \vspace{-3mm}
    \includegraphics[width=\linewidth]{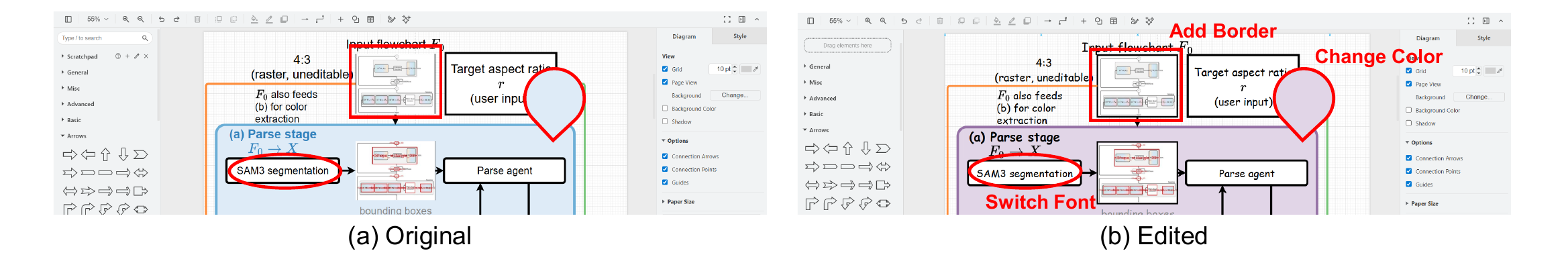}
    \vspace{-7mm}
    \caption{
    \textbf{Editable XML output in draw.io.}
    (a) The generated $X^*$ can be opened directly in draw.io, where each block, label, connector, and style attribute remains editable.
    (b) Users can modify fonts, borders, colors, and component positions independently, while explicit source/target references preserve connector relationships after editing.
    }
      \label{fig:editable_representation}
    \vspace{-3mm}
\end{figure}

\vspace{-2mm}
All intermediate and final representations are maintained in draw.io-compatible mxGraph XML, which explicitly encodes blocks, edges, and group relationships. Each edge stores \texttt{source} and \texttt{target} references, allowing connectors to update when blocks move. This enables direct editing of the parsed $X$, styled $\hat{X}$, and final $X^*$ without regenerating the entire diagram, while also supporting more reliable critic checks of structure, style, and connectivity, as illustrated in Fig.~\ref{fig:editable_representation}.

\vspace{-3mm}
\section{Experiments}

\vspace{-3mm}
\paragraph{Implementation Details.}
All agents (Parse, Style, Layout, and their Critic Agents) are built on Gemini 3.1 Pro Preview. Draw.io-compatible mxGraph XML serves as the unified intermediate representation, and the draw.io CLI renders it as PNG at each stage for Critic visual comparison. The Parse, Style, and Layout stages run for at most 3, 2, and 5 iterations, respectively, with early termination once the Critic issues a passing judgment. Full agent prompts are listed in Appendix~\ref{app:pipeline-prompts}.

\vspace{-2mm}
\subsection{Experimental Setup}
\vspace{-2mm}
We introduce \textsc{FlowchartRelayoutBench}, a benchmark pairing 100 source flowcharts at five target ratios with a four-metric VLM-as-a-Judge protocol under hierarchical scoring.

\vspace{-2mm}
\paragraph{Dataset.}
The dataset comprises 100 flowcharts from oral-level and above papers at top-tier AI/CV venues over the past three years: ECCV (30), CVPR (30), NeurIPS (12), ICLR (10), ICCV (9), and ICML (9). To cover heterogeneous source-diagram structures, we deliberately curate examples spanning three
predefined structural-complexity levels: 25 easy, 40 medium, and 35 hard flowcharts. The benchmark contains 15.5 nodes and 13.6 edges per figure on average, with both quantities generally increasing with structural complexity. Detailed definitions, statistics, and performance breakdowns for the three levels are provided in Appendix~\ref{app:complexity}.
Each source is evaluated at five baseline-supported aspect ratios (9:16, 2:3, 1:1, 3:2, 16:9), yielding 500 relayout tasks.

\vspace{-2mm}
\paragraph{Evaluation Metrics.}
We adopt a VLM-as-a-Judge protocol for reference-based evaluation. For each sample, the judge model compares the relayout result against the original flowchart under the target canvas size, and assesses whether the candidate preserves the original structure, content, layout adaptation, and visual style (Appendix~\ref{app:bench-relationship}, \ref{app:bench-hallucination}, \ref{app:bench-layout}, and~\ref{app:bench-style}). We independently validate agreement between these VLM judgments and human assessments in Sec.~\ref{sec:human_vlm_agreement}.

\vspace{-3mm}
\paragraph{Evaluation Dimensions.}
Inspired by PaperBanana's evaluation protocol~\citep{zhu2026paperbanana}, we evaluate relayout results from two perspectives: \textbf{Content Fidelity} and \textbf{Visual Quality}.

\begin{itemize}[leftmargin=*]
\vspace{-1mm}
    \item \textbf{Content Fidelity.} 
    Content Fidelity has two binary checks. \textbf{Relationship Preservation} verifies that connector logic (source, target, and flow direction) matches the original. \textbf{Hallucination-free Rate} checks that no elements are added or omitted. Each is pass/fail; the score is the pass ratio.
\vspace{-1mm}
    \item \textbf{Visual Quality.}
    Visual Quality contains two ranking-based metrics. \textbf{Layout Quality} ranks valid candidates by canvas utilization and overall arrangement, while penalizing trivial scaling, stretching, or simply rotating a landscape layout to fit a portrait canvas. \textbf{Style Similarity} ranks by preservation of original block colors, text styles, and connector styles. The judge jointly ranks all methods per sample; we report the average rank (lower is better).
\vspace{-1mm}
\end{itemize}

\vspace{-3mm}
\paragraph{Hierarchical Scoring.}
Hierarchical scoring prioritizes content correctness over visual quality. Candidates are first ranked by the number of passed Content Fidelity checks, with two passes ranked above one pass and one pass above none. Ties are then ordered by the combined Layout Quality and Style Similarity. Finally, rank $k$ among $N$ candidates is mapped to $(N-k)/(N-1)\times100$.

\vspace{-3mm}
\paragraph{Baselines.}
We compare against four baselines in three categories. \textbf{Image-to-image}: Nano Banana Pro and GPT Image 2.0 directly generate raster output from the source. \textbf{T2I agentic}: PaperBanana~\citep{zhu2026paperbanana}, adapted via text descriptions of each source. \textbf{Parse-then-render}: AutoFigure-Edit~\citep{lin2026autofigure}, with its VLM settings adapted to target the requested ratio (it does not natively support resizing). All baselines are conditioned on the target ratio and source.

\begin{table*}[t]
\centering
\vspace{-5mm}
\caption{
\textbf{Quantitative comparison on \textsc{FlowchartRelayoutBench}.}
Five methods across three families are evaluated under Correctness, Visual Quality, and Overall, a hierarchical score prioritizing Content Fidelity over Visual Quality. Our method achieves the best Correctness metrics and Overall score (81.85, 1.15$\times$ the next-best). PaperBanana ranks best on Layout despite weak correctness, while Nano Banana Pro and GPT Image 2.0 rank well on Style by largely preserving the original appearance. \textbf{Bold} = best, \underline{underline} = second-best, \textit{italic} = third-best.
}
\label{tab:comparison}
\footnotesize
\setlength{\tabcolsep}{4pt}
\renewcommand{\arraystretch}{0.9}
\begin{tabular}{lccc cc c}
\toprule
& \multicolumn{3}{c}{\textbf{Correctness} (\%) $\uparrow$}
& \multicolumn{2}{c}{\textbf{Visual Quality} $\downarrow$}
& \\
\cmidrule(lr){2-4} \cmidrule(lr){5-6}
\textbf{Method} & \textbf{Relationship} & \textbf{Hallu.-free} & \textbf{Content} 
& \textbf{Layout} & \textbf{Style} & \textbf{Overall}$\uparrow$ \\
\midrule
\multicolumn{7}{l}{\textit{Image-to-image}} \\
\quad Nano Banana Pro      & \textit{49.8} & \textit{69.4} & \underline{41.4} & \textit{2.85} & \textbf{1.35} & \underline{71.10} \\
\quad GPT Image 2.0      & \underline{50.2} & \underline{75.4} & \textit{40.2} & 3.17 & \underline{2.08} & \underline{71.10} \\
\addlinespace[2pt]
\multicolumn{7}{l}{\textit{T2I / agentic}} \\
\quad PaperBanana~\citep{zhu2026paperbanana}   & 38.4 & 18.6 & 11.2 & \textbf{2.14} & 4.34 & 39.05 \\
\addlinespace[2pt]
\multicolumn{7}{l}{\textit{Parse-then-render}} \\
\quad AutoFigure-Edit~\citep{lin2026autofigure} & 41.8 & 41.4 & 24.8 & 3.63 & 4.16 & \textit{47.90} \\
\midrule
\rowcolor{gray!10}
\textbf{Ours}            & \textbf{74.2} & \textbf{90.6} & \textbf{68.6} & \underline{2.66} & \textit{2.98} & \textbf{81.85} \\
\bottomrule
\end{tabular}%
\vspace{-3mm}
\end{table*}

\begin{figure*}[t]
\vspace{-3mm}
    \centering
    \small
    \setlength{\tabcolsep}{2.5pt}
    \renewcommand{\arraystretch}{1.0}

    \newcommand{\imgcell}[1]{%
    \begin{minipage}[c][0.12\textheight][c]{\linewidth}
    \centering
    \setlength{\fboxsep}{0pt}%
    \setlength{\fboxrule}{0.4pt}%
    \fcolorbox{gray}{white}{%
    \includegraphics[width=0.95\linewidth,height=0.11\textheight,keepaspectratio]{#1}%
    }%
    \end{minipage}
    }

    \newcommand{\imgcellwide}[1]{%
    \begin{minipage}[c][0.09\textheight][c]{\linewidth}
    \centering
    \setlength{\fboxsep}{0pt}%
    \setlength{\fboxrule}{0.4pt}%
    \fcolorbox{gray}{white}{%
    \includegraphics[width=0.95\linewidth,height=0.08\textheight,keepaspectratio]{#1}%
    }%
    \end{minipage}
    }

    \resizebox{\textwidth}{!}{%
    \begin{tabular}{
        >{\centering\arraybackslash}m{1.1cm}
        |>{\centering\arraybackslash}m{3.0cm}
        |>{\centering\arraybackslash}m{3.0cm}
        |>{\centering\arraybackslash}m{3.0cm}
        |>{\centering\arraybackslash}m{3.0cm}
        |>{\centering\arraybackslash}m{3.0cm}
        |>{\centering\arraybackslash}m{3.0cm}
    }
    \toprule
    \textbf{Ratio} 
    & \textbf{Original flowchart}
    & \textbf{Nano Banana Pro}
    & \textbf{GPT Image 2.0}
    & \textbf{PaperBanana}
    & \textbf{AutoFigure-Edit}
    & \textbf{Ours} \\
    \midrule

    \textbf{3:2}
    & \imgcellwide{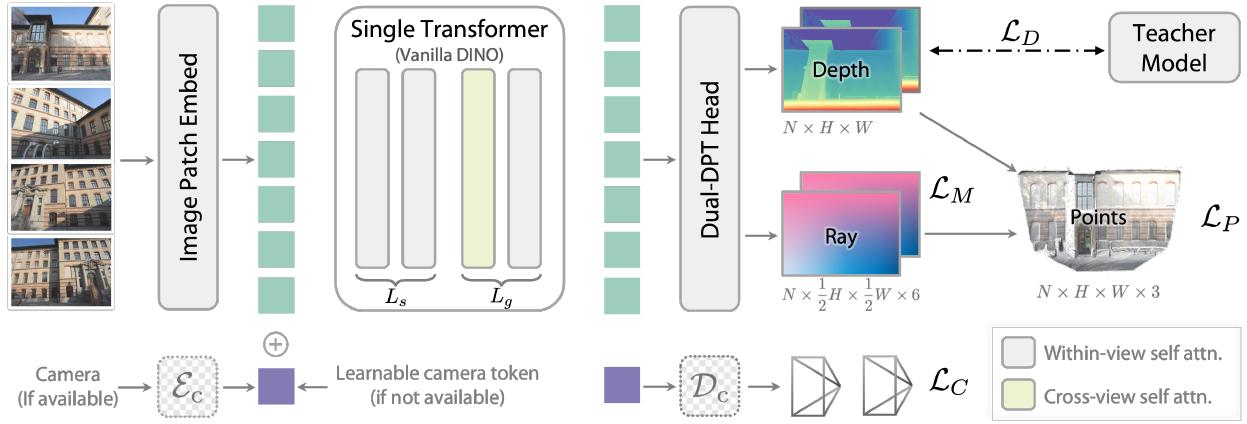}
    & \imgcellwide{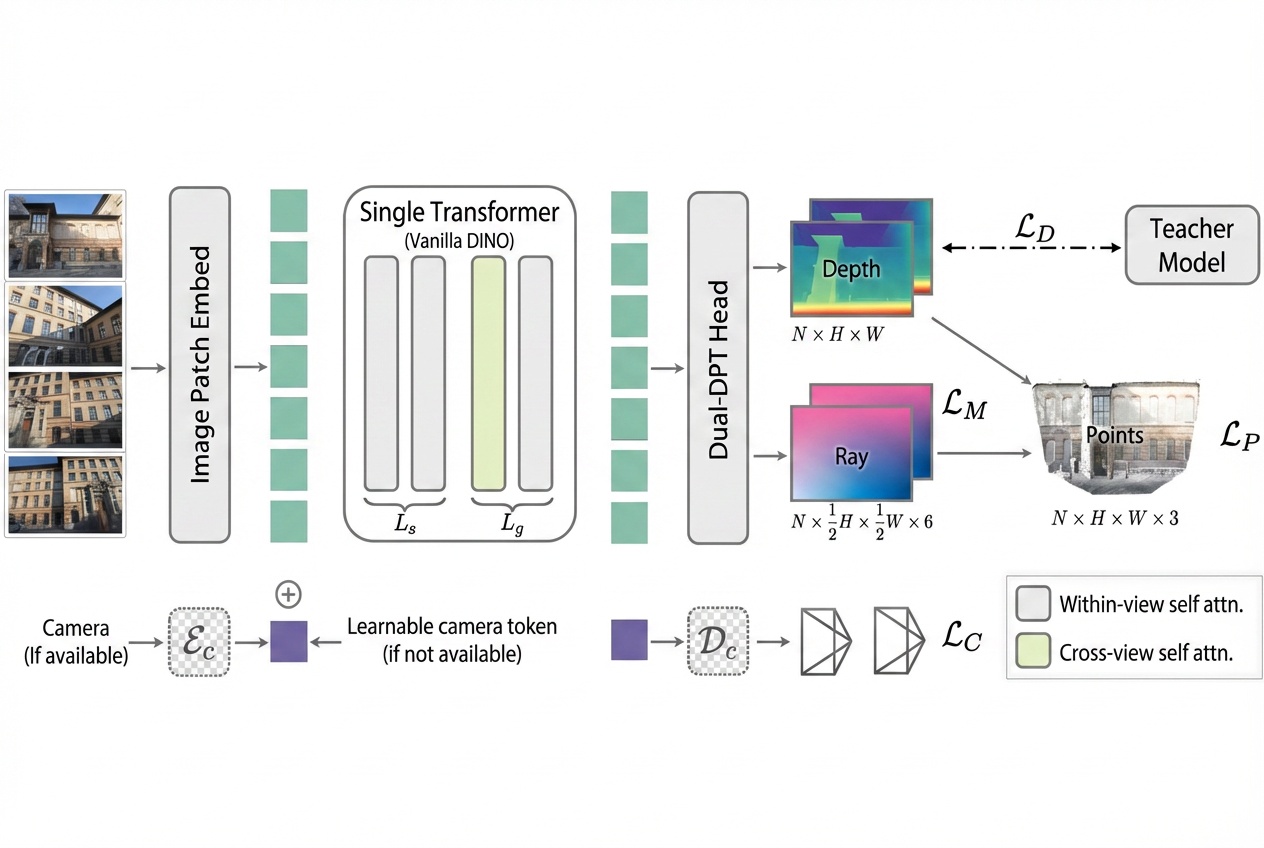}
    & \imgcellwide{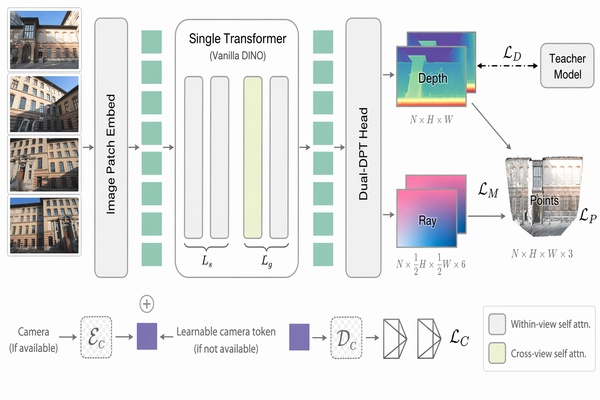}
    & \imgcellwide{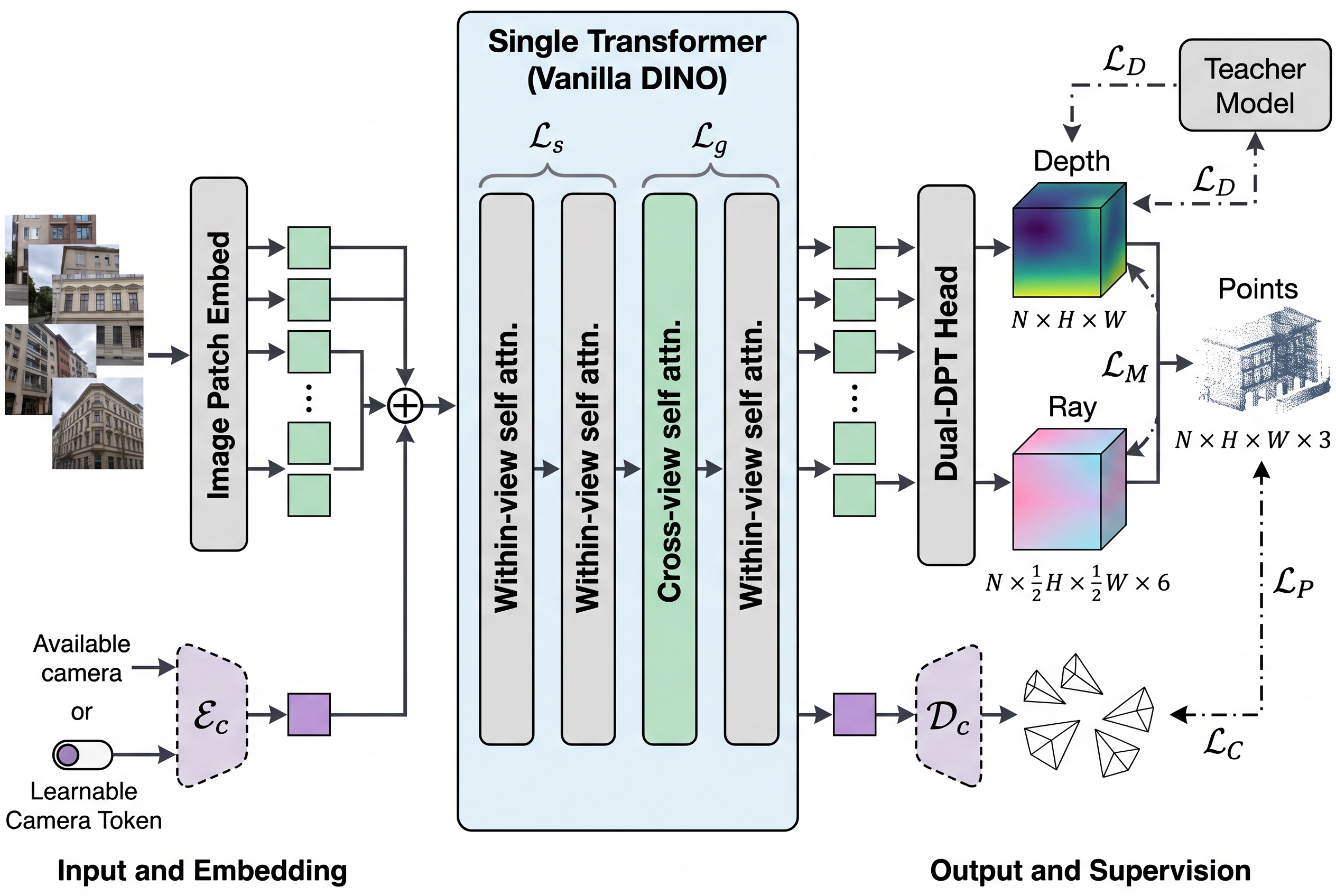}
    & \imgcellwide{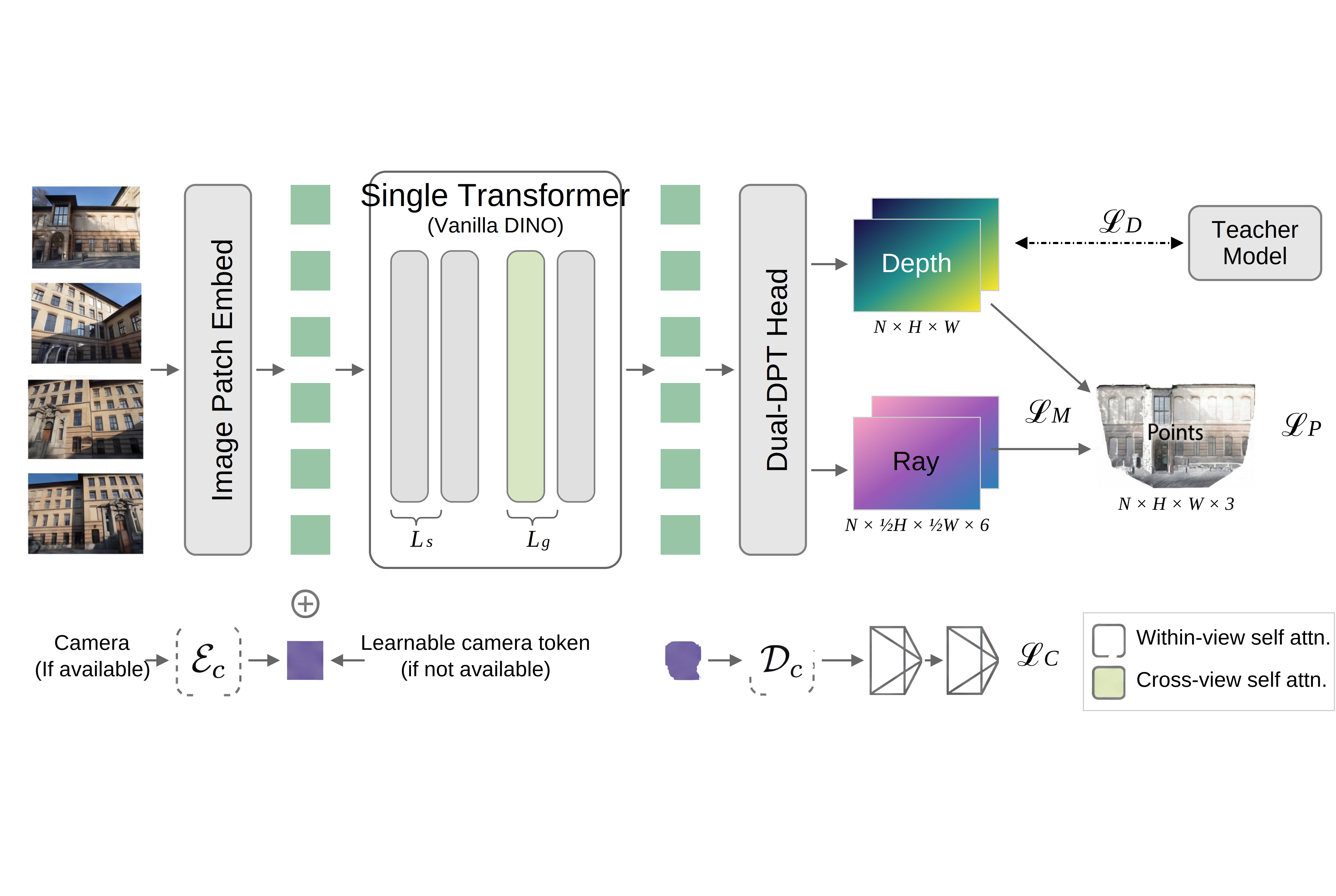}
    & \imgcellwide{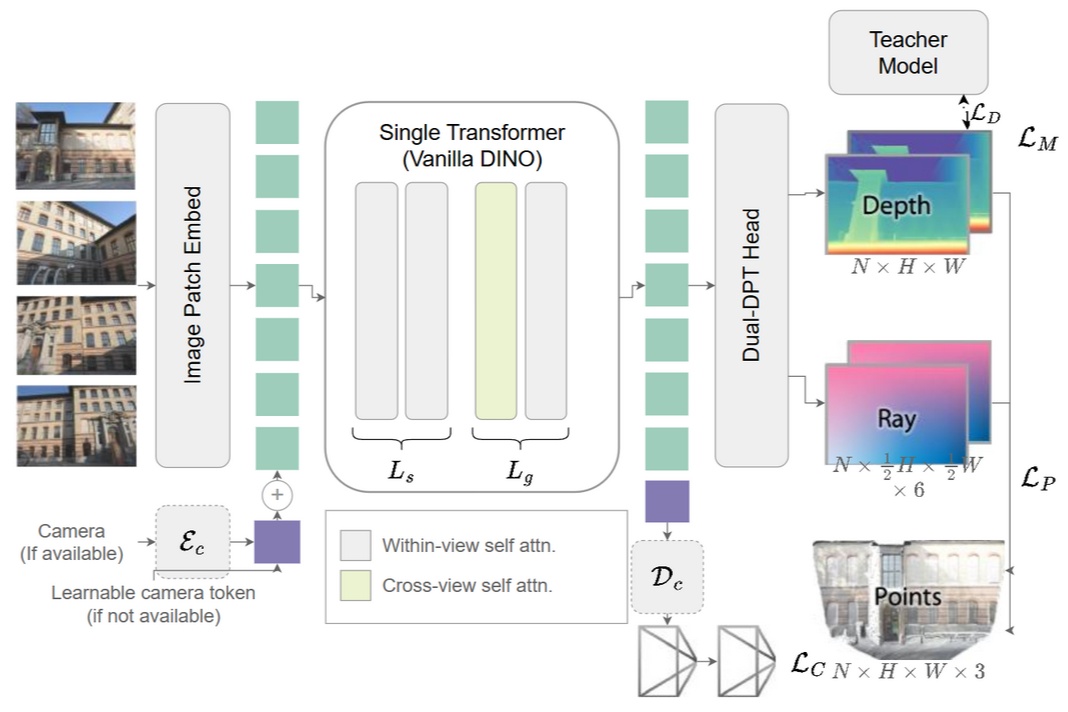} \\
    \hline

    \textbf{1:1}
    & \imgcell{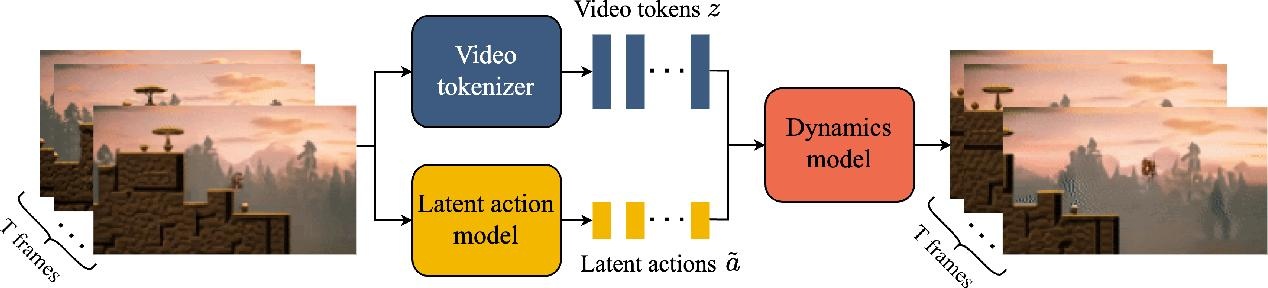}
    & \imgcell{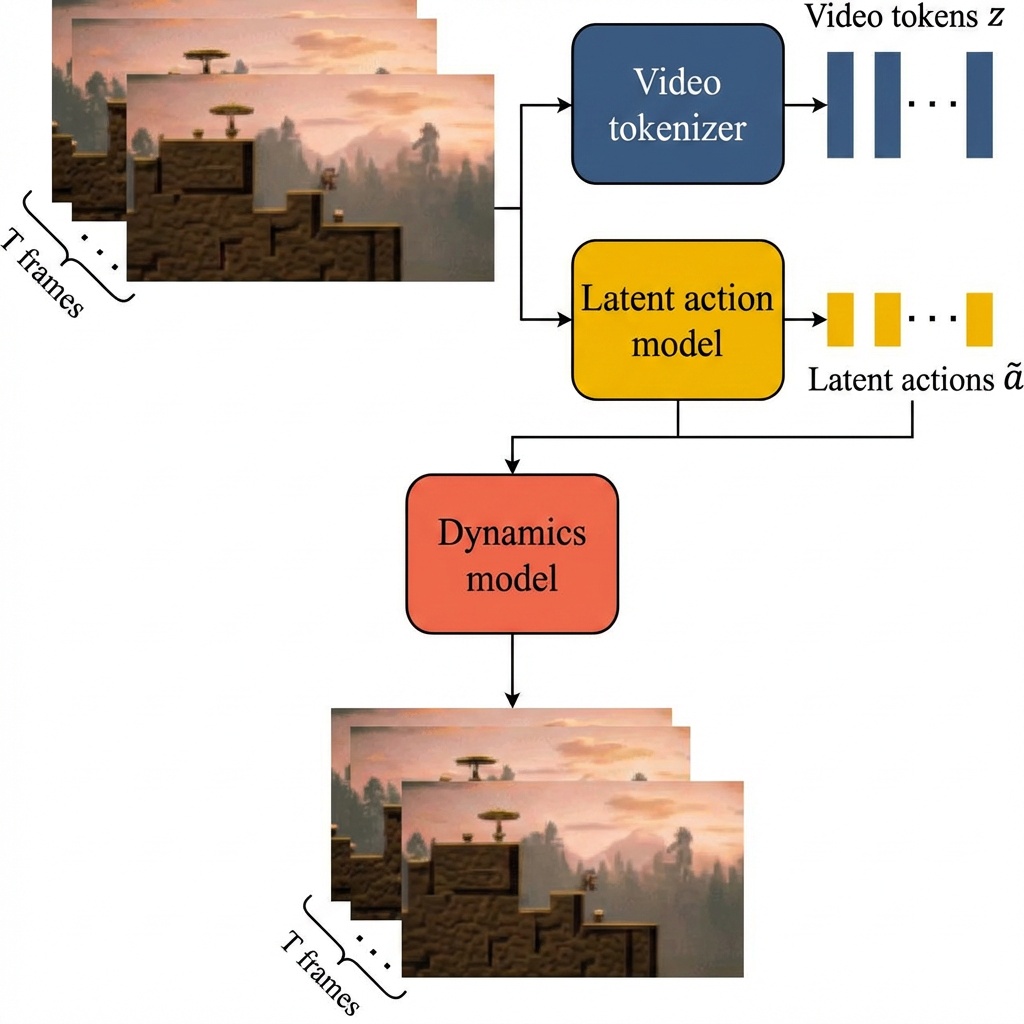}
    & \imgcell{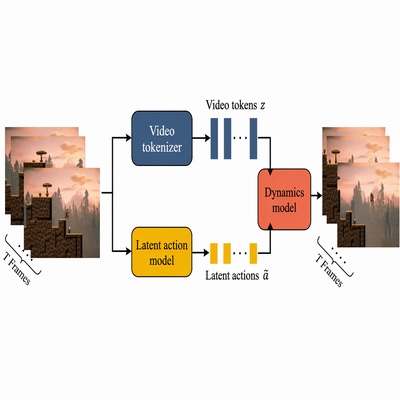}
    & \imgcell{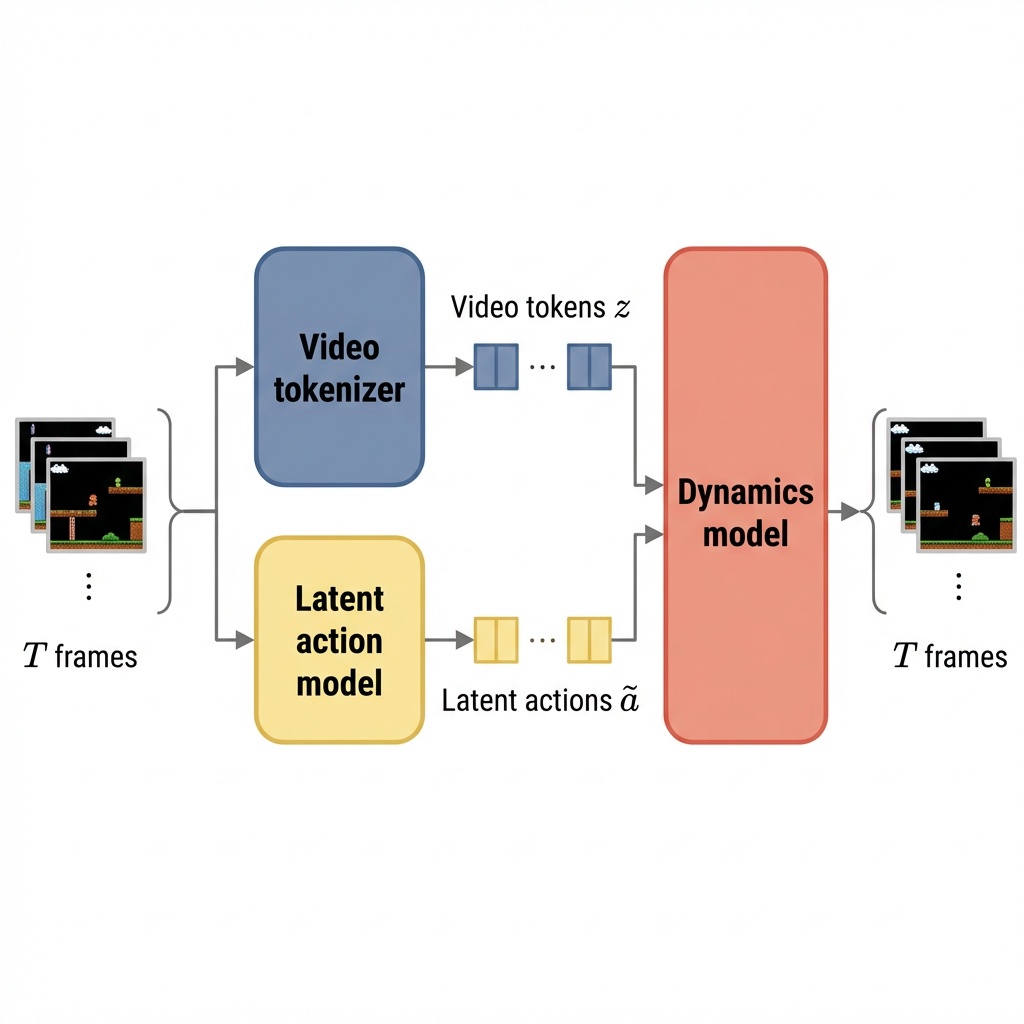}
    & \imgcell{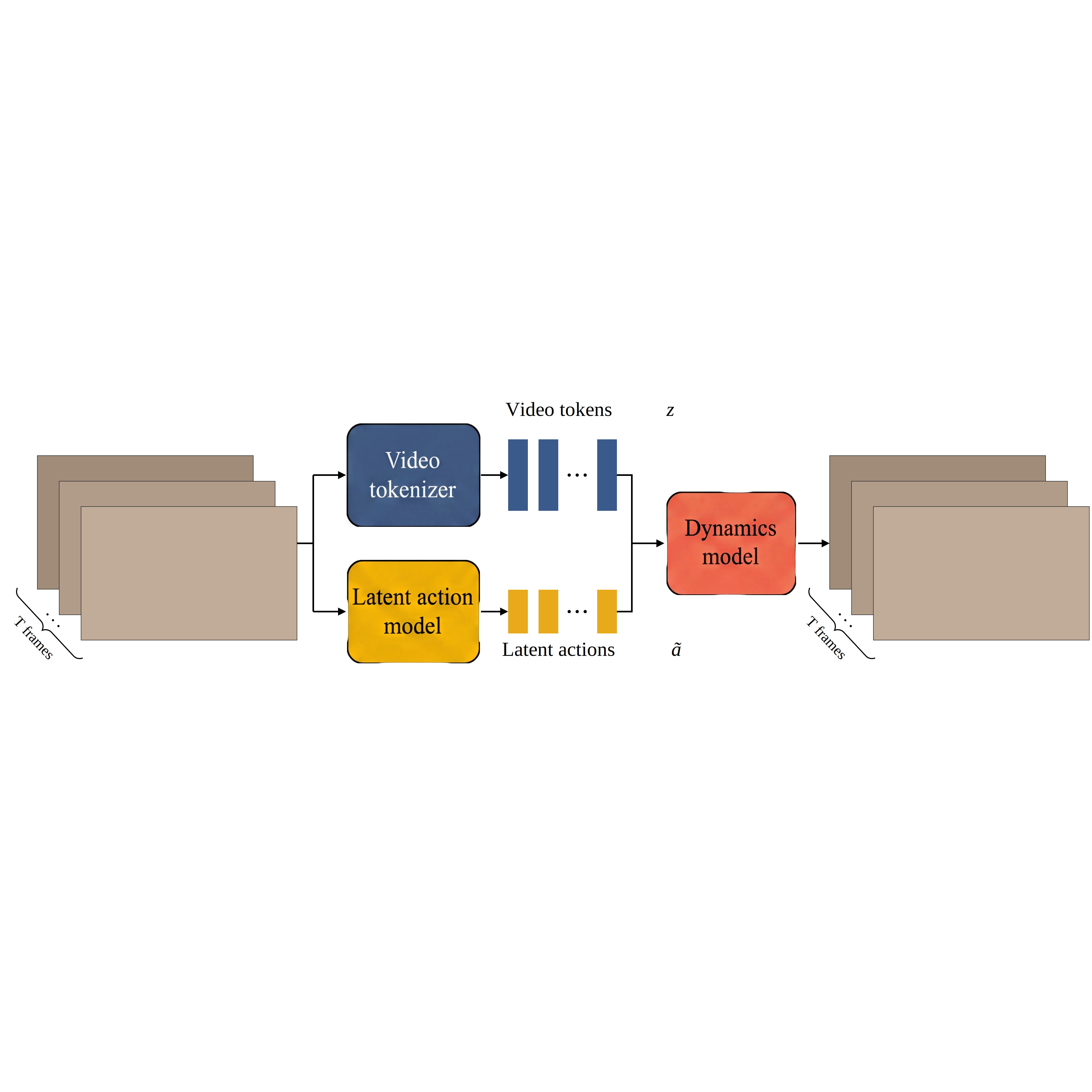}
    & \imgcell{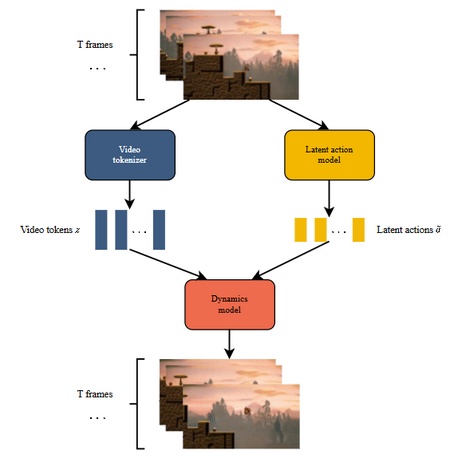} \\
    \hline

    \textbf{2:3}
    & \imgcell{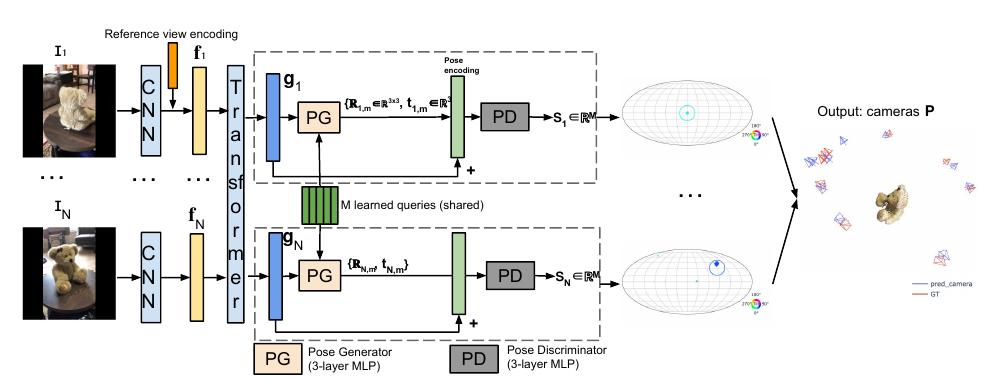}
    & \imgcell{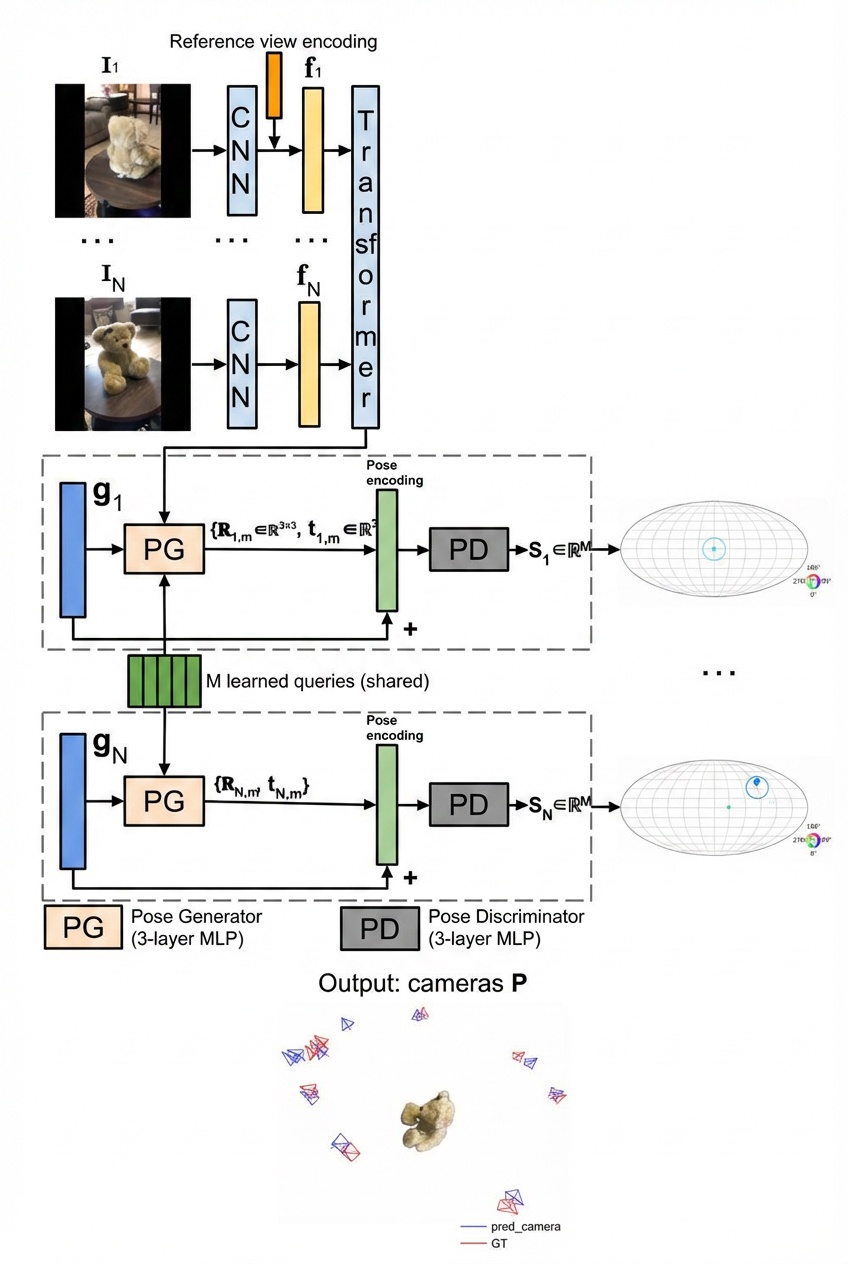}
    & \imgcell{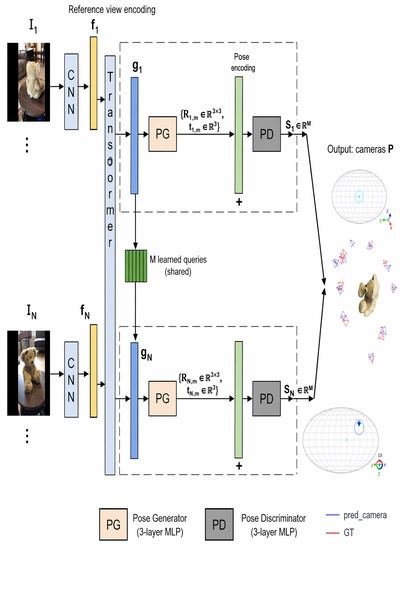}
    & \imgcell{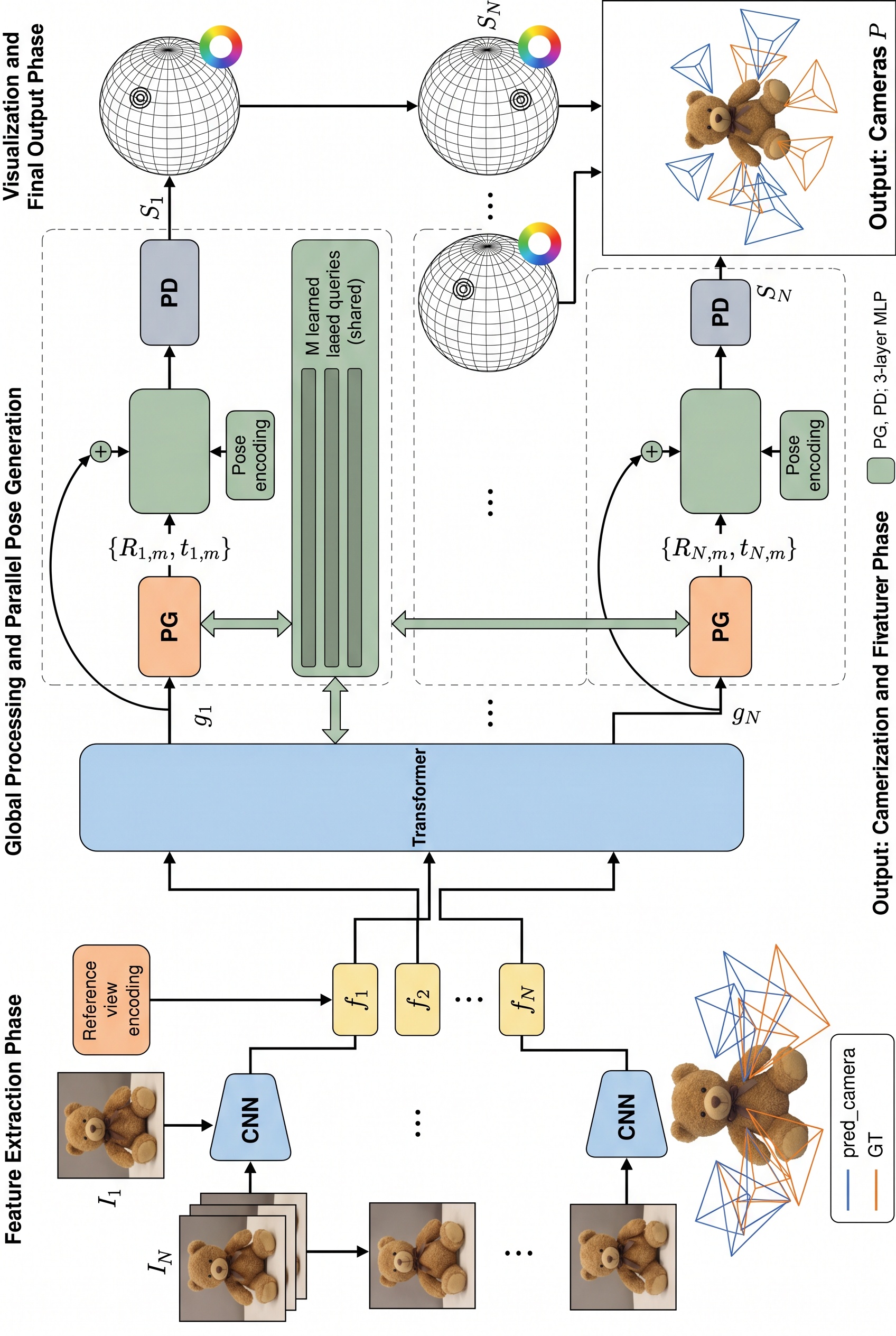}
    & \imgcell{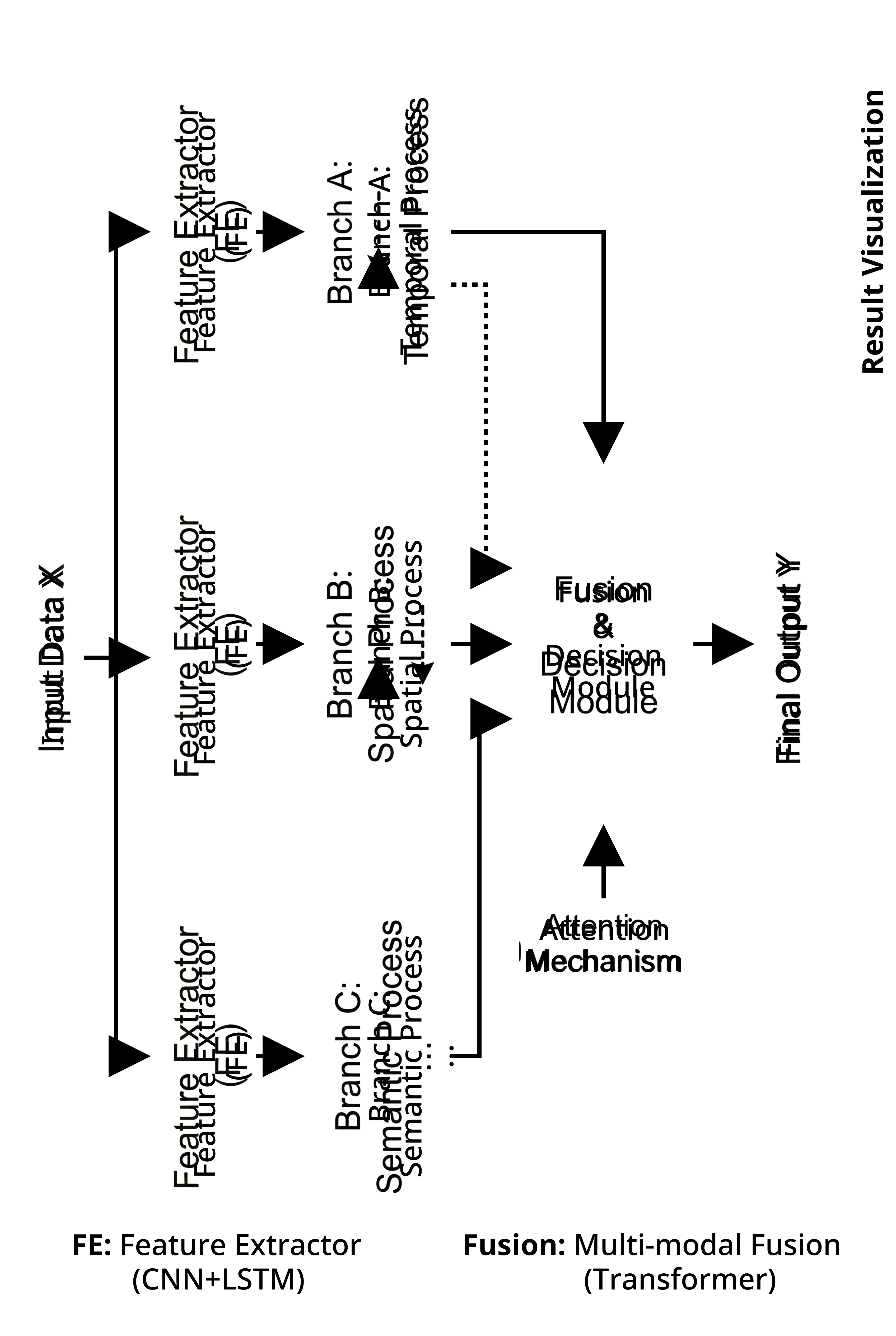}
    & \imgcell{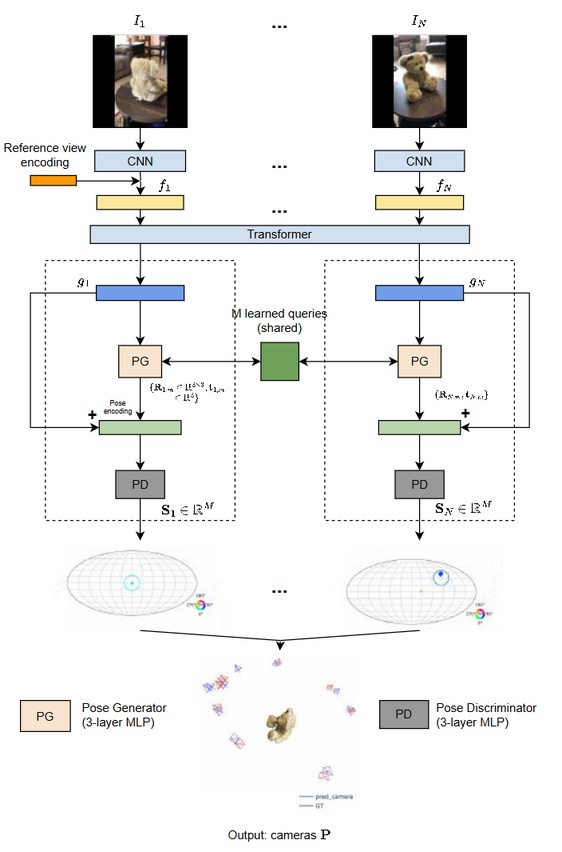} \\
    \bottomrule
    \end{tabular}
    }

    \vspace{-2mm}
    \caption{
    \textbf{Qualitative comparison under different target aspect ratios.}
    We compare our method with four baselines at three target ratios: 1:1, 2:3, and 3:2. 
    \textbf{GPT Image 2.0} typically compresses or crops the original; \textbf{Nano Banana Pro} attempts relayout more often but tends to break relationships. 
    \textbf{PaperBanana} rotates horizontal layouts to fit portrait canvases, and \textbf{AutoFigure-Edit} may introduce hallucinated elements or inconsistent relationships. 
    \textbf{Ours} preserves relationships, avoids hallucinations, uses the canvas effectively, and produces editable XML. Please zoom in for details.
    }
    \label{fig:qualitative_comparison}
    \vspace{-3mm}
\end{figure*}

\vspace{-2mm}
\subsection{Comparisons}
\label{sec:comparisons}
\vspace{-2mm}

Table~\ref{tab:comparison} reports quantitative comparisons against all baselines. Our method achieves the best Relationship Preservation (74.2\%), Hallucination-free Rate (90.6\%), Content Fidelity (68.6\%), and Overall score (81.85), showing stronger structural and content preservation. Qualitative comparisons are shown in Fig.~\ref{fig:qualitative_comparison}. For more visual results, please refer to Appendix~\ref{app:qualitative}.

\vspace{-3mm}
\paragraph{Content Fidelity.}
I2I models often preserve relationships through padding or stretching, but can still distort or hallucinate elements. AutoFigure-Edit performs worse because its SVG lacks explicit source and target connectivity, while PaperBanana is further affected by hallucination and relationship mismatches from text-driven regeneration.

\vspace{-3mm}
\paragraph{Visual Quality.}
I2I models rank better in Style Similarity by largely preserving the original pixels, while our XML reconstruction introduces a modest style gap. PaperBanana ranks best in Layout Quality because its text-driven generation does not preserve the original block geometry or exact connectivity, allowing freer visual arrangement.

\vspace{-3mm}
\paragraph{Overall.}
Our method achieves the highest Overall score (81.85), reflecting a better balance between content correctness and visual quality. Baselines with strong visual appearance score lower when content or connectivity is not preserved.

\vspace{-3mm}
\subsection{User Study}
 
\vspace{-3mm}
\paragraph{Experiment Setup.}
We conduct a user study on 50 cases, each with one source flowchart and five candidate relayouts (ours and four baselines), yielding 250 outputs in total. Each participant evaluates 4 randomly assigned cases, providing binary judgments on Relationship Preservation and Hallucination-free Rate, and ranking candidates 1st–5th for Layout Quality and Style Similarity. Participants are recruited from CS backgrounds, many with CV/ML expertise; detailed criteria reduce inter-annotator variability.
 
\begin{figure*}[t]
    \centering
    \begin{minipage}[t]{0.48\textwidth}
        \centering
        \includegraphics[width=\linewidth]{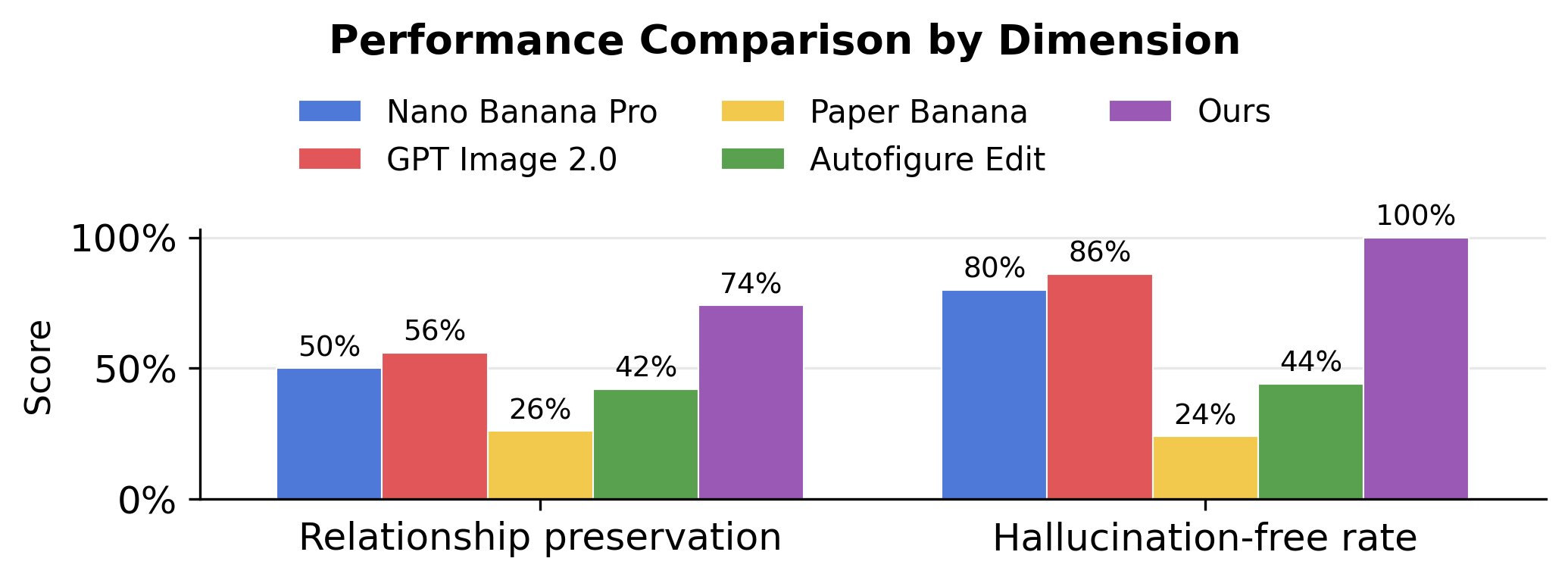}\\
        \small (a) Content Fidelity
    \end{minipage}
    \hfill
    \begin{minipage}[t]{0.48\textwidth}
        \centering
        \includegraphics[width=\linewidth]{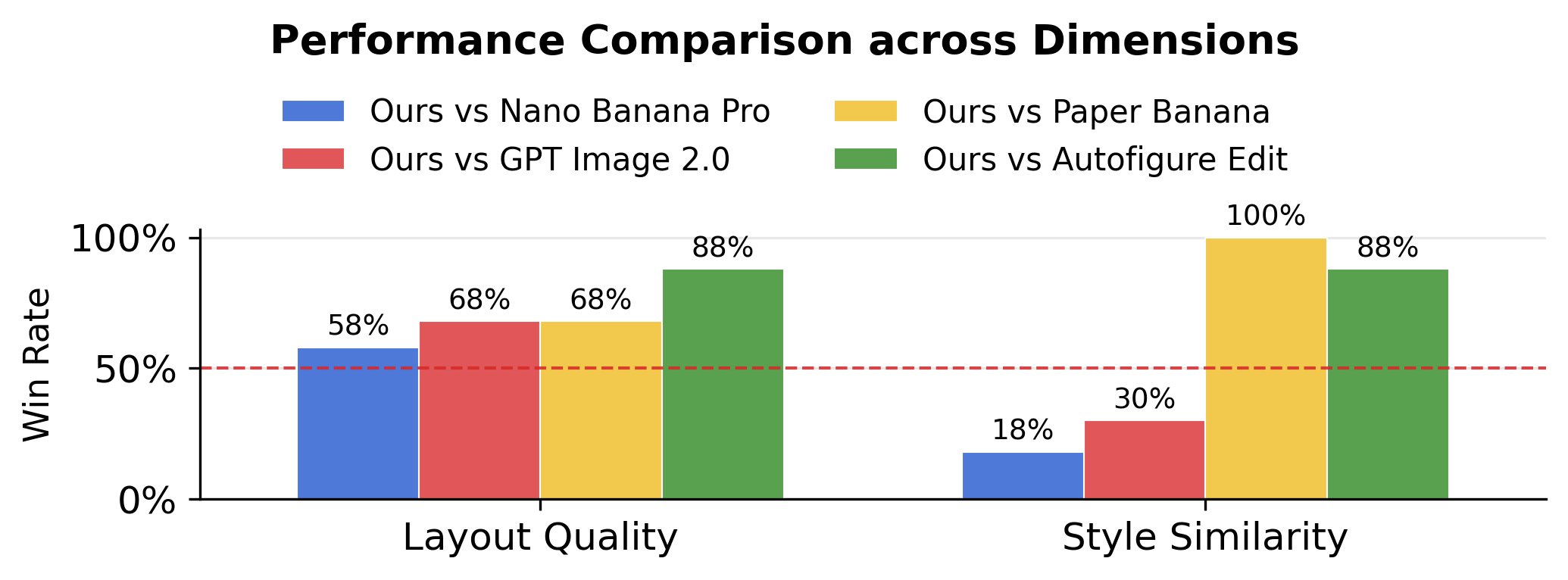}\\
        \small (b) Visual Quality
    \end{minipage}
    \vspace{-3mm}
    \caption{\textbf{User study results.}
    We report human evaluation results for content-related metrics and visual-quality metrics. Our method achieves the best performance on Relationship Preservation and Hallucination-free Rate, and shows strong pairwise win rates in Layout Quality and Style Similarity.}
    \label{fig:user_study_results}
\vspace{-3mm}
\end{figure*}

\vspace{-3mm}
\paragraph{Result Analysis.}
We collect 600 image-level evaluations from 30 participants (Fig.~\ref{fig:user_study_results}). Our method performs best on Relationship Preservation, Hallucination-free Rate, and Layout Quality, while I2I baselines remain stronger in Style Similarity. Despite their near-verbatim appearance preservation, users still find our editable XML reconstruction visually competitive.

\vspace{-3mm}
\subsection{Agreement Between User Study and VLM-as-a-Judge}
\label{sec:human_vlm_agreement}
\vspace{-3mm}

\begin{table}[t]
\vspace{-5mm}
\centering
\caption{\textbf{Agreement between VLM-as-a-Judge and human evaluation.}
We report Phi coefficient ($\phi$) and classification agreement for binary Relationship Preservation and Hallucination-free judgments, and Spearman's rank correlation ($\rho$) for Layout Quality and Style Similarity. All correlations are significant at $p<0.001$.}
\label{tab:human_vlm_agreement}
\small
\begin{tabular}{lcccc}
\toprule
\textbf{Metric} & \textbf{Statistic} & \textbf{Corr.} & \textbf{Agreement} & \textbf{95\% CI} \\
\midrule
Relationship   & $\phi$ & 0.671 & 83.2\% & [0.560, 0.753] \\
Hallucination  & $\phi$ & 0.593 & 82.0\% & [0.486, 0.693] \\
Layout         & $\rho$ & 0.620 & -- & [0.513, 0.714] \\
Style          & $\rho$ & 0.757 & -- & [0.704, 0.807] \\
\bottomrule
\end{tabular}
\end{table}

Since our benchmark relies on VLM-as-a-Judge, we evaluate its agreement with human judgments from the user study using Phi coefficient for binary metrics and Spearman's rank correlation for ranking-based metrics (Table~\ref{tab:human_vlm_agreement}).

\vspace{-3mm}
\subsection{Ablation Study}
\label{sec:ablation}
\vspace{-3mm}
We conduct quantitative ablations on a complexity-balanced subset of 30 flowcharts, with 10 easy, 10 medium, and 10 hard examples. We examine: (1) multi-stage decomposition versus direct prompting, (2) the contributions of the Style Stage and Critic Agents, and (3) replacing our Layout Stage with a conventional graph-layout algorithm. All variants use the same 30 flowcharts and evaluation protocol. Visual comparisons are provided in Appendix~\ref{app:ablation}.

\begin{table*}[t]
\centering
\small
\vspace{-4mm}
\caption{\textbf{Quantitative ablation study on 30 flowcharts.}
We compare direct prompting, Stage/Critic removal, and Graphviz DOT replacement. Layout, Style, and Overall are computed jointly within the ablation and are therefore not directly comparable with Table~\ref{tab:comparison}. \textbf{Bold} = best, \underline{underline} = second-best, \textit{italic} = third-best.
}
\label{tab:ablation}
\renewcommand{\arraystretch}{0.9}
\setlength{\tabcolsep}{2.1pt}
\begin{tabular}{lcccccc}
\toprule
\textbf{Method}
& \textbf{Rel.}$\uparrow$
& \textbf{Hallu.-free}$\uparrow$
& \textbf{Content}$\uparrow$
& \textbf{Layout}$\downarrow$
& \textbf{Style}$\downarrow$
& \textbf{Overall}$\uparrow$ \\
\midrule

\multicolumn{7}{l}{\textit{Direct prompting / decomposition}} \\

\quad Direct Prompt Parse Stage
& 53.3
& 60.0
& 33.3
& 5.00
& 7.67
& 50.74 \\

\quad Direct Prompt Layout Stage
& \textit{70.0}
& \underline{93.3}
& \textit{66.7}
& \underline{4.07}
& 5.03
& \textit{69.26} \\

\quad Direct Prompt VLM-to-XML (Gemini 3.1 Pro)
& 43.3
& 50.0
& 33.3
& 5.23
& 6.87
& 41.11 \\

\quad Direct Prompt VLM-to-XML (GPT-5.5)
& 33.3
& 80.0
& 30.0
& 4.97
& 7.77
& 49.63 \\

\addlinespace[2pt]
\multicolumn{7}{l}{\textit{Stage / Critic removal}} \\

\quad w/o All Critic Agents
& 60.0
& 80.0
& 46.7
& 4.60
& 5.50
& 61.85 \\

\quad w/o Parse Critic
& \underline{73.3}
& 80.0
& 60.0
& 4.83
& \textit{3.83}
& 63.33 \\

\quad w/o Layout Critic
& 66.7
& \textit{83.3}
& 60.0
& 4.77
& 4.40
& 60.00 \\

\quad w/o Style Stage
& \underline{73.3}
& \underline{93.3}
& \underline{70.0}
& \textit{4.33}
& \underline{2.53}
& \underline{76.30} \\

\addlinespace[2pt]
\multicolumn{7}{l}{\textit{Conventional layout algorithm}} \\

\quad w/ Graphviz (DOT) Layout
& 43.3
& 36.7
& 20.0
& 6.77
& 9.67
& 35.19 \\

\addlinespace[2pt]
\multicolumn{7}{l}{\textit{Full pipeline}} \\
\midrule
\rowcolor{gray!10}
\quad \textbf{Ours}
& \textbf{76.7}
& \textbf{96.7}
& \textbf{76.7}
& \textbf{3.90}
& \textbf{1.73}
& \textbf{87.04} \\

\bottomrule
\end{tabular}
\vspace{-4mm}
\end{table*}

\vspace{-3mm}
\paragraph{Direct prompting vs.\ multi-stage decomposition.}
Replacing the Parse or Layout Stage with direct prompting reduces Content Fidelity from 76.7\% to 33.3\% and 66.7\%, respectively, while one-shot VLM-to-XML generation drops it to 33.3\% with Gemini~3.1 Pro and 30.0\% with GPT-5.5. The full pipeline also achieves the best Layout and Style ranks, showing that task-specific decomposition improves both structural fidelity and visual quality beyond the underlying VLM alone.

\vspace{-3mm}
\paragraph{Contribution of Style Stages and Critic Agents.}
Removing all Critic Agents reduces Content Fidelity from 76.7\% to 46.7\% and Overall from 87.04 to 61.85. Removing either the Parse or Layout Critic also degrades Content Fidelity, while removing the Style Stage mainly worsens Style Similarity and lowers Overall to 76.30. These results confirm the complementary roles of stage-specific critics in correcting structural and layout errors and of the Style Stage in preserving appearance.

\vspace{-3mm}
\paragraph{Comparison with conventional graph layout.}
Replacing our Layout Stage with Graphviz DOT reduces the Overall score. Its topology-driven layout lacks explicit target-canvas awareness, often leaving substantial unused space, and fails to preserve relationships involving container structures.
 
\vspace{-3mm}
\section{Conclusions}
\vspace{-3mm}
We present an agentic pipeline for aspect-ratio-adaptive flowchart relayout, refitting a raster flowchart to an arbitrary target canvas while preserving structure, content, and visual style in a draw.io-editable output. The pipeline decomposes the problem into three stages: Parse, Style, and Layout. Each stage is guarded by a Critic Agent that pairs deterministic verification with VLM reasoning, so connectivity is explicitly checked and prevented from being silently broken. Experiments on \textsc{FlowchartRelayoutBench} and a user study confirm substantial gains over image-generation and agentic figure-generation baselines, particularly on structural and Content Fidelity.
Limitations are discussed in Appendix~\ref{app:limitations}.

\newpage

\bibliographystyle{iclr2027_conference}
\bibliography{iclr2027_conference}

\newpage
\appendix

\makeatletter\@addtoreset{lstlisting}{section}\makeatother
\renewcommand{\thelstlisting}{\thesection.\arabic{lstlisting}}
\section*{Appendix Overview}

This appendix provides additional details on benchmark construction,
limitations, ablation studies, open-weight model transfer, style transfer,
qualitative results, and implementation. We also provide an interactive HTML
viewer showcasing baseline comparisons and relayouts at arbitrary aspect ratios.

Appendix~\ref{app:complexity} defines the easy, medium, and hard
structural-complexity levels used to construct
\textsc{FlowchartRelayoutBench}, reports node and edge statistics, and
provides performance breakdowns across the three levels. Appendix~\ref{app:limitations} discusses limitations of the current method
across style fidelity, VLM dependency, convergence, edge routing, and runtime
cost. Appendix~\ref{app:ablation} provides qualitative comparisons complementing
the quantitative ablation study in Section~\ref{sec:ablation}, visualizing
representative ablation variants and the conventional Graphviz DOT layout. Appendix~\ref{app:openweight} evaluates model transfer by replacing the
proprietary VLM with an open-weight Qwen model and compares the full pipeline
against direct XML generation and Qwen-based image editing. Appendix~\ref{app:style-transfer} presents our optional reference-guided
style transfer extension and qualitative examples. Appendix~\ref{app:qualitative} presents extended qualitative comparisons of
our method against all four baselines across multiple flowchart examples and
target aspect ratios. Appendix~\ref{app:pipeline-prompts} provides the implementation prompts used
by all six pipeline agents, while Appendix~\ref{app:benchmark-prompts}
provides the prompts used by the four VLM-based benchmark evaluators.

\section{Benchmark Structural Complexity}
\label{app:complexity}

We deliberately curate \textsc{FlowchartRelayoutBench} to cover source
flowcharts with different levels of structural complexity. Rather than
defining complexity only by diagram size, we consider how clearly diagram
components and their relationships can be recovered from the raster input.
We use three predefined levels: easy, medium, and hard.

\paragraph{Easy.}
Easy flowcharts contain clearly identifiable node boundaries, with connectors
terminating unambiguously at their corresponding nodes. Relationships and flow
directions are explicitly represented, with little ambiguity in either
component identity or connectivity.

\paragraph{Medium.}
Medium flowcharts retain clear logical relationships, but contain mild visual
ambiguity. For example, nodes may have irregular or weakly defined boundaries,
or some connectors may be visually abbreviated or omitted while their source,
target, and direction remain readily inferable from the surrounding structure.

\paragraph{Hard.}
Hard flowcharts contain substantial structural ambiguity. Typical cases include
dense or crossing connectors, node boundaries that are difficult to delineate,
ambiguous logical relationships, or omitted connectors that make source--target
relationships or flow directions difficult to infer.

As summarized in Table~\ref{tab:complexity_results}, the final benchmark contains 25 easy, 40 medium, and 35 hard flowcharts. The structural statistics are consistent with increasing complexity: easy, medium, and hard examples contain on average 10.1, 14.3, and 20.8 nodes, respectively, and 9.4, 13.4, and 16.8 edges. Across the full benchmark, flowcharts contain 15.5 nodes and 13.6 edges on average. These statistics provide an additional quantitative characterization of the structural diversity represented in the benchmark.

\subsection{Performance by Structural Complexity}
\label{app:complexity_results}

We additionally evaluate all methods separately on the easy, medium, and hard subsets to examine how source-diagram complexity affects relayout performance. As shown in Table~\ref{tab:complexity_results}, Relationship Preservation and Content Fidelity generally decrease as structural complexity increases, indicating that the three subsets represent progressively more challenging cases. Our method achieves the highest Relationship Preservation, Hallucination-free Rate, Content Fidelity, and Overall score at every complexity level, while maintaining a lead of more than 20 percentage points in Content Fidelity over the next-best method across all three subsets.

\begin{table*}[t]
\centering
\caption{
\textbf{Performance across source-diagram structural complexity.}
Results are reported on the easy, medium, and hard subsets of
\textsc{FlowchartRelayoutBench}. Overall follows the hierarchical scoring in
Table~\ref{tab:comparison}. Our method achieves the highest Content Fidelity and
Overall score at all three complexity levels.
\textbf{Bold} = best, \underline{underline} = second-best, and
\textit{italic} = third-best within each level.
}
\label{tab:complexity_results}
\vspace{2mm}
\footnotesize
\renewcommand{\arraystretch}{0.95}
\setlength{\tabcolsep}{3.5pt}
\begin{tabular}{llccc cc c}
\toprule
& & \multicolumn{3}{c}{\textbf{Correctness} (\%) $\uparrow$}
& \multicolumn{2}{c}{\textbf{Visual Quality} $\downarrow$}
& \\
\cmidrule(lr){3-5} \cmidrule(lr){6-7}
\textbf{Complexity} & \textbf{Method} & \textbf{Relationship} & \textbf{Hallu.-free} & \textbf{Content}
& \textbf{Layout} & \textbf{Style} & \textbf{Overall}$\uparrow$ \\
\midrule

\multirow{5}{*}{\shortstack{\textbf{Easy}\\[2pt]\footnotesize Nodes: 10.1$\pm$3.4\\\footnotesize Edges: 9.4$\pm$3.3}}
& Nano Banana Pro & \textit{62.4} & \textit{77.6} & \textit{56.0} & \textit{2.78} & \textbf{1.54} & \textit{69.40} \\
& GPT Image 2.0 & \underline{66.4} & \underline{83.2} & \underline{56.8} & 3.01 & \underline{2.09} & \underline{70.00} \\
& PaperBanana & 52.0 & 25.6 & 18.4 & \textbf{2.36} & 4.50 & 33.20 \\
& AutoFigure-Edit & 61.6 & 48.8 & 39.2 & 3.62 & 3.88 & 45.00 \\
\cmidrule(lr){2-8}
\rowcolor{gray!10}
& \textbf{Ours} & \textbf{93.6} & \textbf{98.4} & \textbf{92.8} & \underline{2.46} & \textit{2.86} & \textbf{80.60} \\
\midrule

\multirow{5}{*}{\shortstack{\textbf{Medium}\\[2pt]\footnotesize Nodes: 14.3$\pm$6.5\\\footnotesize Edges: 13.4$\pm$7.4}}
& Nano Banana Pro & \textit{48.0} & \textit{72.0} & \textit{39.0} & \textit{2.86} & \textbf{1.24} & \textit{71.50} \\
& GPT Image 2.0 & \underline{52.5} & \underline{77.5} & \underline{41.0} & 3.17 & \underline{2.08} & \underline{73.25} \\
& PaperBanana & 43.5 & 18.5 & 10.0 & \textbf{2.08} & 4.33 & 41.63 \\
& AutoFigure-Edit & 39.0 & 42.0 & 23.5 & 3.65 & 4.23 & 46.25 \\
\cmidrule(lr){2-8}
\rowcolor{gray!10}
& \textbf{Ours} & \textbf{70.0} & \textbf{87.5} & \textbf{62.5} & \underline{2.68} & \textit{3.03} & \textbf{79.75} \\
\midrule

\multirow{5}{*}{\shortstack{\textbf{Hard}\\[2pt]\footnotesize Nodes: 20.8$\pm$11.3\\\footnotesize Edges: 16.8$\pm$8.0}}
& Nano Banana Pro & \underline{42.9} & \textit{60.6} & \underline{33.7} & \textit{2.90} & \textbf{1.34} & \underline{71.86} \\
& GPT Image 2.0 & \textit{36.0} & \underline{67.4} & \textit{27.4} & 3.29 & \underline{2.08} & \textit{69.43} \\
& PaperBanana & 22.9 & 13.7 & 7.4 & \textbf{2.05} & 4.25 & 40.29 \\
& AutoFigure-Edit & 30.9 & 35.4 & 16.0 & 3.63 & 4.29 & 51.86 \\
\cmidrule(lr){2-8}
\rowcolor{gray!10}
& \textbf{Ours} & \textbf{65.1} & \textbf{88.6} & \textbf{58.3} & \underline{2.79} & \textit{3.01} & \textbf{85.14} \\

\bottomrule
\end{tabular}
\vspace{-3mm}
\end{table*}

\subsection{Refinement Iterations by Structural Complexity}
\label{app:complexity_iterations}

To examine how structural complexity affects refinement effort, we report the average number of iterations performed by each pipeline stage on the easy, medium, and hard subsets. As shown in Table~\ref{tab:complexity_iterations}, total iterations increase with structural complexity, from 8.23 for easy to 9.08 for hard examples. The increase is most pronounced in the Layout Stage, while Parse iterations remain relatively stable, suggesting that structural complexity primarily increases the refinement required for spatial rearrangement.

\begin{table}[t]
\centering
\caption{
\textbf{Refinement Iterations by Structural Complexity.}
Average refinement iterations (mean $\pm$ std.) for the Parse, Style, and Layout stages and the full pipeline. The maximum iteration budgets are 3, 2, and 5, respectively.
}
\label{tab:complexity_iterations}
\small
\begin{tabular}{lcccc}
\toprule
\textbf{Complexity}
& \textbf{Parse}
& \textbf{Style}
& \textbf{Layout}
& \textbf{Total} \\
\midrule
Easy
& $2.33 \pm 0.80$
& $1.63 \pm 0.48$
& $4.42 \pm 1.088$
& $8.38 \pm 1.68$ \\

Medium
& $2.56 \pm 0.59$
& $1.82 \pm 0.38$
& $4.74 \pm 0.78$
& $9.13 \pm 1.04$ \\

Hard
& $2.65 \pm 0.67$
& $1.81 \pm 0.39$
& $4.89 \pm 0.45$
& $9.35 \pm 0.96$ \\
\bottomrule
\end{tabular}
\vspace{-3mm}
\end{table}

\section{Limitations}
\label{app:limitations}

\paragraph{Style gap.}
Our XML-based reconstruction framework prioritizes editability over exact pixel preservation, which introduces an inherent visual style gap relative to methods that directly manipulate original image pixels. Although the generated diagrams remain structurally faithful and fully editable, subtle discrepancies in typography, icon rendering, spacing, or low-level visual appearance may still occur. Closing this gap while preserving editable graph structure remains an important direction for future work.

\paragraph{Closed-source VLM dependency.}
Our main experiments rely on Gemini 3.1 Pro Preview, a proprietary multimodal model, which introduces limitations in reproducibility, accessibility, deployment cost, and privacy-sensitive usage scenarios. Future model updates outside our control may also affect output behavior. In Appendix~\ref{app:openweight}, we replace Gemini with the open-weight Qwen3.6-27B model. The proposed pipeline continues to outperform direct generation under the same VLM, but its substantially lower absolute structural fidelity shows that current performance remains sensitive to the capability of the underlying model.

\paragraph{Convergence not guaranteed.}
The proposed critic-guided refinement mechanism is heuristic rather than formally guaranteed. Each stage terminates after either receiving a passing judgment or reaching a fixed maximum number of iterations. Although the deterministic validators substantially reduce structural failures, the overall iterative process does not guarantee convergence or globally optimal layouts. In rare cases, outputs may still contain subtle visual artifacts or aesthetically suboptimal arrangements that are not detected by the VLM critic.

\paragraph{Edge routing failures.}
Our deterministic checker verifies the integrity of source--target connectivity relationships but does not explicitly optimize edge-routing aesthetics. As a result, diagrams with high node density, complex graph topology, or extreme aspect-ratio transformations may still exhibit unnecessary edge crossings, uneven spacing, or visually unnatural routing despite preserving correct connectivity.

\paragraph{Runtime and API cost.}
The multi-stage agentic pipeline introduces substantially higher inference cost and latency than direct one-shot generation methods. Each relayout instance requires approximately 10--20 VLM API calls across the three stages, in addition to one SAM~3 inference pass. Runtime is dominated by remote API latency rather than local computation. While this overhead is manageable for individual editing tasks, reproducing the full benchmark evaluation may require sustained API throughput and incur non-trivial monetary cost under current commercial API pricing.

\section{Qualitative Ablation}
\label{app:ablation}

\begin{figure*}[t]
\centering
\includegraphics[
    width=\textwidth,
    keepaspectratio
]{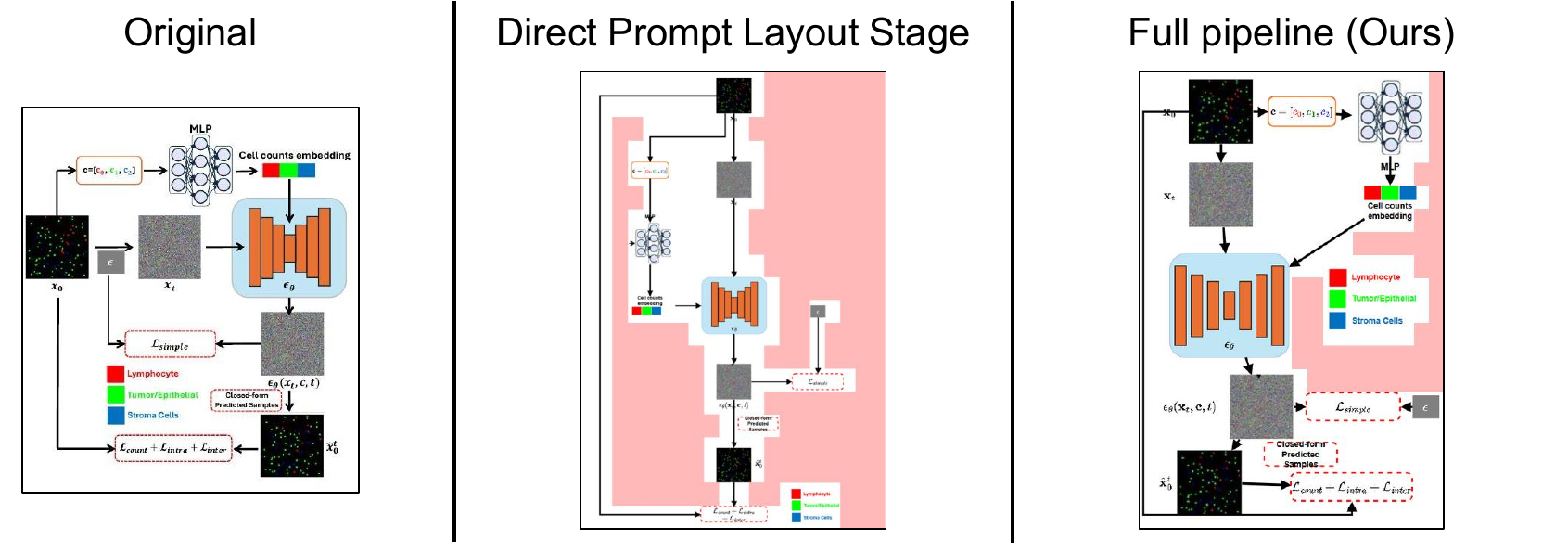}
\caption{
\textbf{Direct Prompt Layout Stage.}
Direct prompting leaves substantial unused space and adapts to the target canvas by excessively shrinking components. Our method makes more effective use of the available canvas while largely preserving component sizes and maintaining clear connector flow.
}
\label{fig:ablation_direct_layout}
\vspace{-3mm}
\end{figure*}

\begin{figure*}[t]
\centering
\includegraphics[
    width=\textwidth,
    keepaspectratio
]{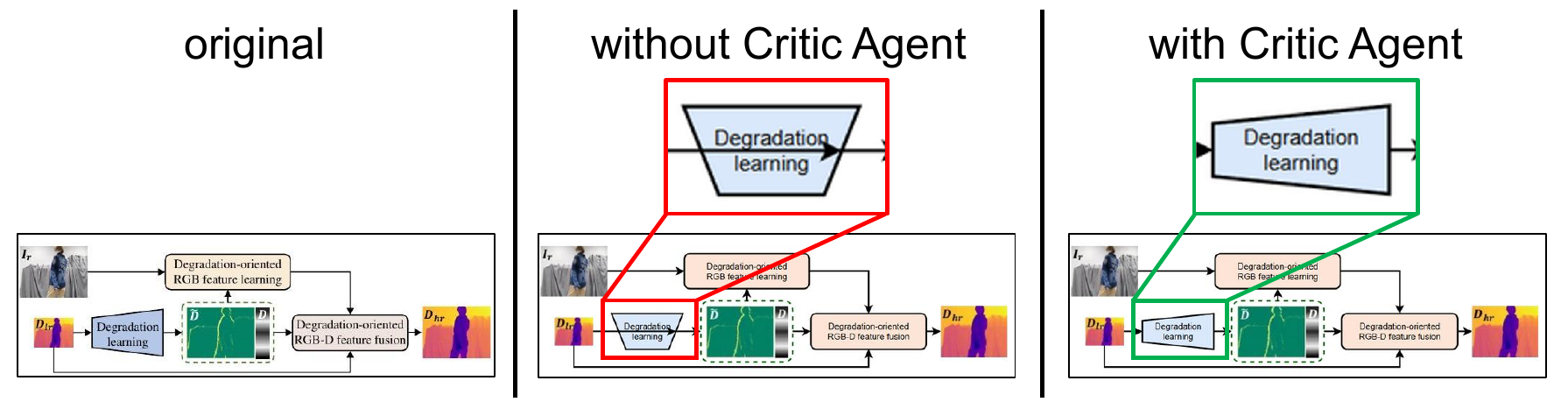}
\caption{
\textbf{Without Parse Critic.}
Structural errors from the initial parse remain uncorrected, including incorrect shape types and mismatched connector attachment points. Iterative Critic feedback improves both component geometry and edge anchoring.
}
\label{fig:ablation_parse_critic}
\vspace{-3mm}
\end{figure*}

\begin{figure*}[t]
\centering
\includegraphics[
    width=\textwidth,
    keepaspectratio
]{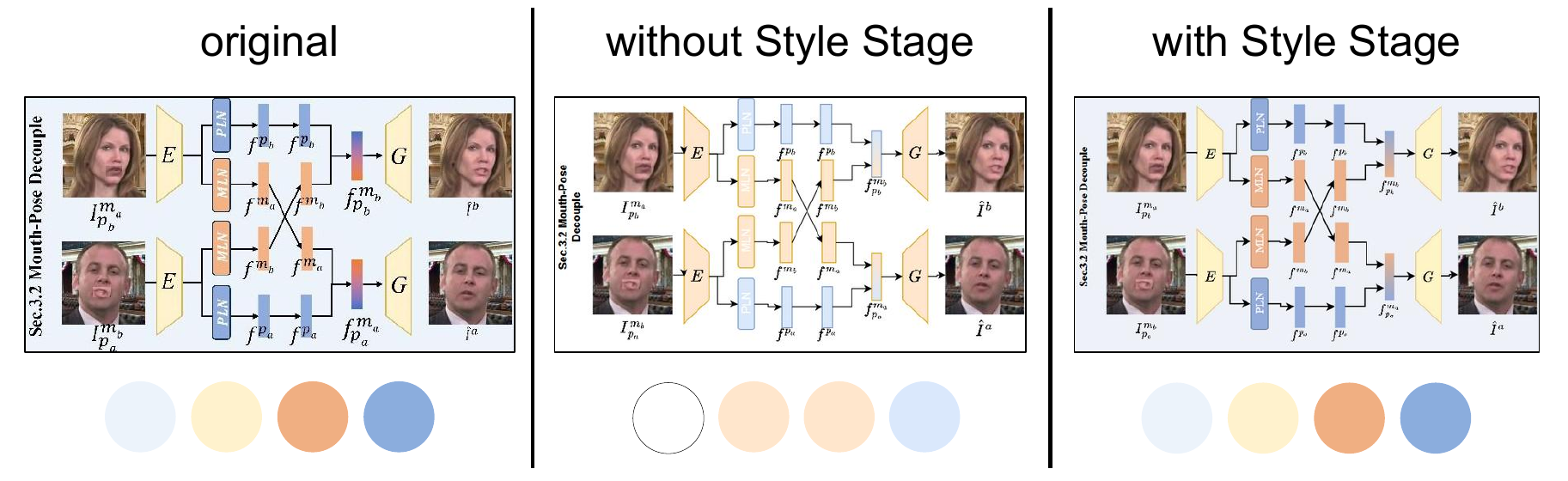}
\caption{
\textbf{Without Style Stage.}
The reconstructed diagram exhibits mismatched block colors, missing or simplified containers, inconsistent connector styles, and typography differences. The Style Stage substantially improves visual correspondence while preserving the recovered structure.
}
\label{fig:style_stage_comparison}
\vspace{-3mm}
\end{figure*}

\begin{figure*}[t]
\centering
\includegraphics[
    width=\textwidth,
    keepaspectratio
]{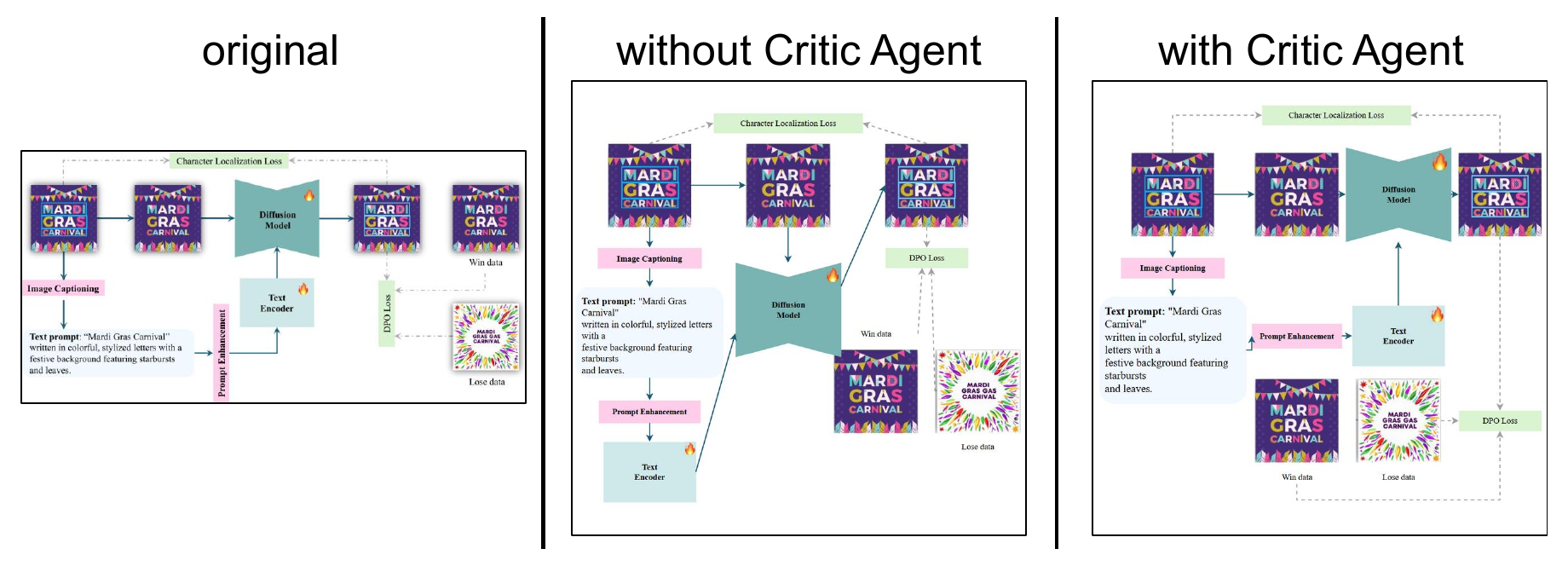}
\caption{
\textbf{Without Layout Critic.}
The single-pass result exhibits weaker space utilization, less natural arrow routing, and poorer visual balance. Iterative deterministic and visual feedback improves placement, compactness, and readability while preserving connectivity.
}
\label{fig:ablation_layout_critic}
\vspace{-3mm}
\end{figure*}

\begin{figure*}[t]
\centering
\includegraphics[
    width=\textwidth,
    keepaspectratio
]{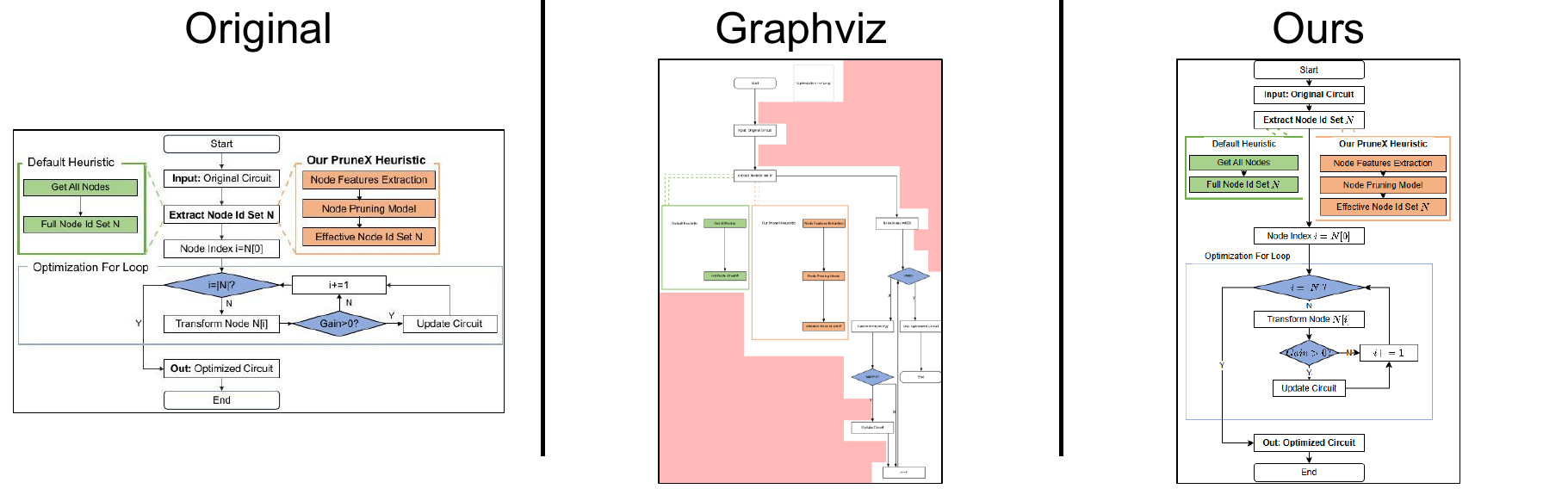}
\caption{
\textbf{Replacing Layout Stage with Graphviz DOT.}
The topology-driven layout often leaves substantial unused space and may fail to preserve relationships involving container structures. Our Layout Stage better utilizes the target canvas while maintaining these structural relationships.
}
\label{fig:graphviz_layout_comparison}
\vspace{-3mm}
\end{figure*}

This section complements the quantitative ablation study in
Section~\ref{sec:ablation} with representative qualitative comparisons.
We visualize direct prompting for the Layout Stage
(Fig.~\ref{fig:ablation_direct_layout}), removal of the Parse Critic
(Fig.~\ref{fig:ablation_parse_critic}), removal of the Layout Critic
(Fig.~\ref{fig:ablation_layout_critic}), removal of the Style Stage
(Fig.~\ref{fig:style_stage_comparison}), and replacement of our Layout
Stage with Graphviz DOT (Fig.~\ref{fig:graphviz_layout_comparison}).
These examples illustrate the characteristic failure modes underlying
the quantitative results in Table~\ref{tab:ablation}.

\section{Open-Weight VLM Transfer}
\label{app:openweight}

To test whether our pipeline generalizes beyond Gemini~3.1 Pro Preview, we replace the underlying VLM with the open-weight Qwen3.6-27B while keeping the Parse--Style--Layout pipeline unchanged, and compare against direct XML generation using Qwen3.6-27B and direct image editing using Qwen-Image-Edit-2511. As shown in Table~\ref{tab:qwen_transfer}, despite the lower absolute performance of Qwen3.6-27B, our pipeline outperforms direct XML generation with the same VLM, reducing XML-format failures from 16.0\% to 0.0\% and improving Content Fidelity from 0.0\% to 20.0\%. Direct image editing often retains the original elongated layout with limited structural rearrangement while introducing visible artifacts and inconsistencies in text and diagram blocks, as illustrated in Fig.~\ref{fig:qwen_qualitative}. These results indicate that the pipeline transfers across model families, although absolute performance remains dependent on the underlying VLM.
\begin{table}[t]
\centering
\caption{
\textbf{Transfer to an open-weight VLM.}
We replace Gemini~3.1 Pro Preview with Qwen3.6-27B while keeping the proposed
pipeline unchanged. Although absolute structural fidelity decreases with the
open-weight model, our pipeline substantially improves over direct XML
generation using the same VLM and eliminates XML-format failures.  Overall is computed within this three-method comparison and is not
directly comparable with Overall scores in Tables~\ref{tab:comparison}
and~\ref{tab:ablation}.
\textbf{Bold} = best, \underline{underline} = second-best,
\textit{italic} = third-best.
}
\label{tab:qwen_transfer}
\vspace{2mm}
\footnotesize
\renewcommand{\arraystretch}{0.95}

\resizebox{\linewidth}{!}{%
\begin{tabular}{lccccccc}
\toprule
\textbf{Method}
& \textbf{XML Fail.}$\downarrow$
& \textbf{Rel.}$\uparrow$
& \textbf{Hallu.-free}$\uparrow$
& \textbf{Content}$\uparrow$
& \textbf{Layout}$\downarrow$
& \textbf{Style}$\downarrow$
& \textbf{Overall}$\uparrow$ \\
\midrule

Direct VLM-to-XML (Qwen3.6-27B)
& 16.0
& \textit{8.0}
& \underline{12.0}
& \textit{0.0}
& \textit{2.36}
& \textit{2.28}
& \textit{38.0} \\

Direct I2I (Qwen-Image-Edit-2511)
& \textbf{0.0}
& \textbf{24.0}
& \textit{8.0}
& \underline{8.0}
& \underline{2.16}
& \underline{2.20}
& \underline{40.0} \\

\midrule
\rowcolor{gray!10}
\textbf{Ours (Qwen3.6-27B)}
& \textbf{0.0}
& \underline{20.0}
& \textbf{52.0}
& \textbf{20.0}
& \textbf{1.40}
& \textbf{1.44}
& \textbf{90.0} \\

\bottomrule
\end{tabular}%
}
\end{table}
\begin{figure*}[t]
    \centering
    \small
    \setlength{\tabcolsep}{3pt}
    \renewcommand{\arraystretch}{1.05}

    \newcommand{\QwenRowHeight}{0.14\textheight}
    \newcommand{\QwenImageHeight}{0.13\textheight}
    \newcommand{\QwenMethodWidth}{3.3cm}

    \newcommand{\qwenimgcell}[1]{%
        \begin{minipage}[c][\QwenRowHeight][c]{\linewidth}
            \centering
            \setlength{\fboxsep}{0pt}%
            \setlength{\fboxrule}{0.5pt}%
            \fcolorbox{black}{white}{%
                \includegraphics[
                    width=0.94\linewidth,
                    height=\QwenImageHeight,
                    keepaspectratio
                ]{#1}%
            }%
        \end{minipage}
    }

    \begin{tabular}{
        >{\centering\arraybackslash}m{\QwenMethodWidth}
        |>{\centering\arraybackslash}m{\QwenMethodWidth}
        |>{\centering\arraybackslash}m{\QwenMethodWidth}
        |>{\centering\arraybackslash}m{\QwenMethodWidth}
    }

    \shortstack[c]{
        \textbf{Original Image}
    }
    &
    \shortstack[c]{
        \textbf{Direct VLM-to-XML}\\
        \textbf{(Qwen3.6-27B)}
    }
    &
    \shortstack[c]{
        \textbf{Direct I2I}\\
        \textbf{(Qwen-Image-Edit-2511)}
    }
    &
    \shortstack[c]{
        \textbf{Ours}\\
        \textbf{(Qwen3.6-27B)}
    }
    \\

    \hline

    \qwenimgcell{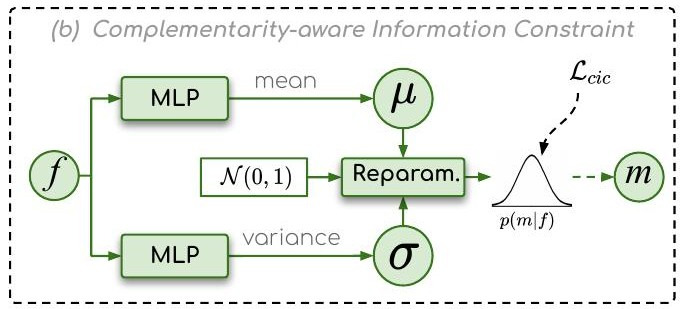}
    &
    \qwenimgcell{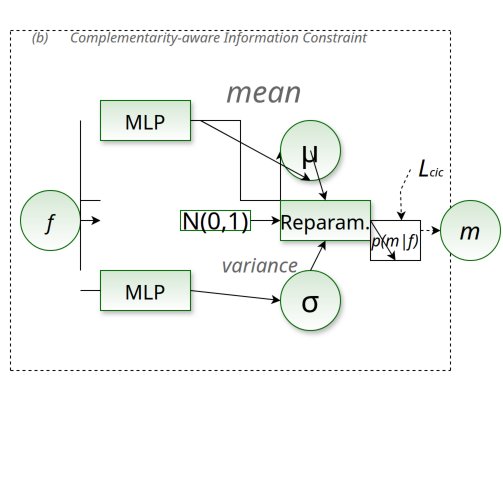}
    &
    \qwenimgcell{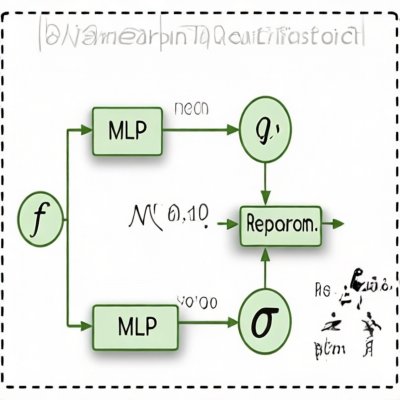}
    &
    \qwenimgcell{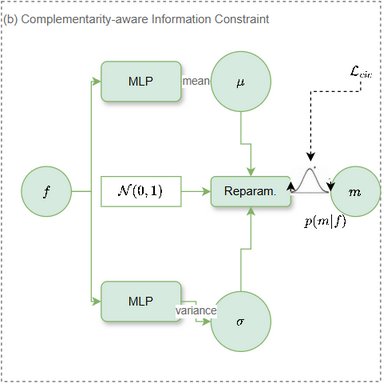}
    \\

    \end{tabular}

     \caption{
    \textbf{Qualitative comparison using open-weight Qwen models under the 1:1 target aspect ratio.}
    Direct VLM-to-XML introduces structural and connectivity errors, including overlapping blocks.
    Direct I2I preserves most visual elements but exhibits missing components, distorted text, and substantial geometric stretching.
    Our method instead produces a coherent relayout with less distortion while better preserving the original structure and connectivity.
    }
    \label{fig:qwen_qualitative}
    \vspace{-3mm}
\end{figure*}

\section{Style Transfer}
\label{app:style-transfer}

\begin{figure*}[t]
    \centering
    \vspace{-3mm}
    \includegraphics[
        width=\textwidth,
        keepaspectratio
    ]{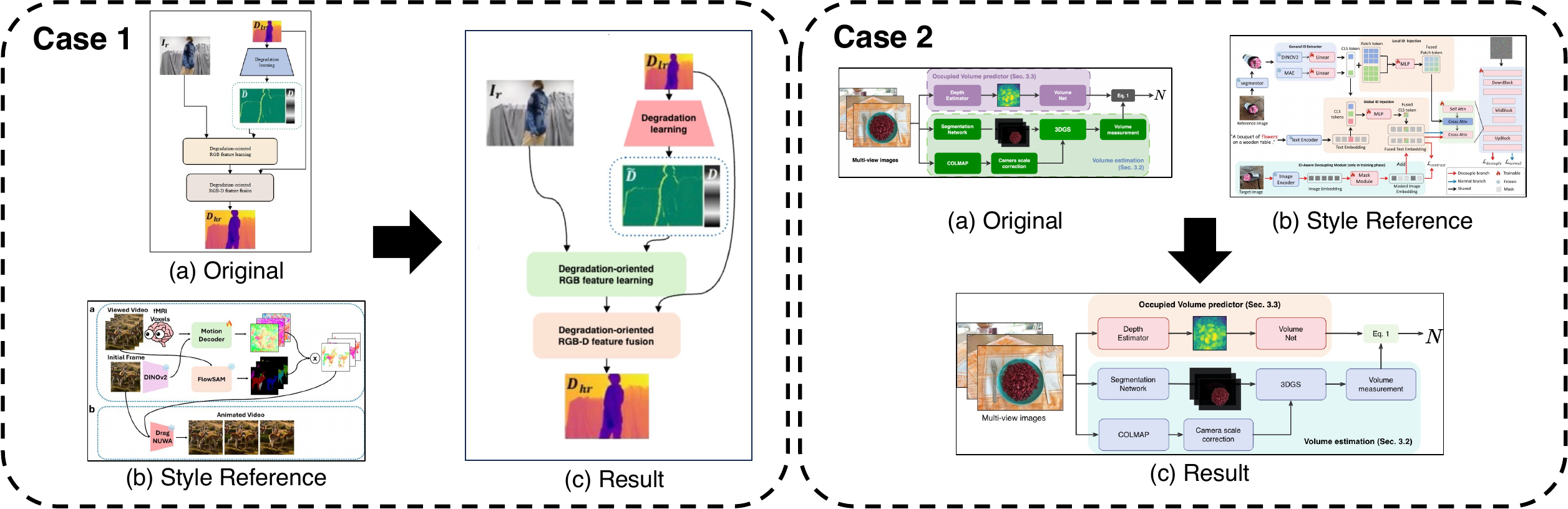}
    \vspace{-6mm}
    \caption{
    \textbf{Style transfer results.}
    Given a completed relayout (a) and an unrelated reference $R$ (b), our Style Transfer extension produces $X^*_R$ (c) that adopts $R$'s fill colors, typography, corner style, and connectors while preserving (a)'s block connectivity and semantic content. Colors are transferred via semantic role assignment (background, primary fill, connector) rather than pixel matching, since (a) and $R$ share no spatial correspondence.
    }
    \label{fig:style_transfer}
    \vspace{-3mm}
\end{figure*}

This appendix details the optional reference-guided Style Transfer extension introduced in Section~\ref{sec:style_transfer}. Given a completed relayout output $X^*$ and an unrelated reference image $R$, the extension transfers the visual style of $R$ to produce a restyled XML $X^*_R$ without altering structural or semantic content. Because $R$ and $X^*$ have no spatial correspondence, a Color Extraction Tool first quantizes colors in LAB space to obtain separate fill and stroke palettes from $R$, after which a VLM assigns each palette entry to semantic roles such as background, primary fill, text, and connector color. A Style Agent then iteratively rewrites the style attributes of $X^*$ under Critic Agent feedback, while inferring non-color properties including corner roundness, font family, and arrow style directly from $R$. A qualitative example is shown in Fig.~\ref{fig:style_transfer}.

\section{Additional Visual Comparisons}
\label{app:qualitative}

\begin{figure*}[t]
    \centering
    \small
    \setlength{\tabcolsep}{2pt}
    \renewcommand{\arraystretch}{1.05}

    \newcommand{\AppRowHeight}{0.11\textheight}    %
    \newcommand{\AppImageHeight}{0.10\textheight}  %
    \newcommand{\AppLabelWidth}{0.75cm}            %
    \newcommand{\AppCaseWidth}{3.2cm}              %

    \newcommand{\appimgcell}[1]{%
    \begin{minipage}[c][\AppRowHeight][c]{\linewidth}
    \centering
    \setlength{\fboxsep}{0pt}%
    \setlength{\fboxrule}{0.5pt}%
    \fcolorbox{black}{white}{%
    \includegraphics[
        width=0.97\linewidth,
        height=\AppImageHeight,
        keepaspectratio
    ]{#1}%
    }%
    \end{minipage}
    }

    \newcommand{\methodlabel}[1]{%
    \rotatebox[origin=c]{90}{\textbf{#1}}%
    }

    \resizebox{\textwidth}{!}{%
    \begin{tabular}{
        >{\centering\arraybackslash}m{\AppLabelWidth}
        |>{\centering\arraybackslash}m{\AppCaseWidth}
        |>{\centering\arraybackslash}m{\AppCaseWidth}
        |>{\centering\arraybackslash}m{\AppCaseWidth}
    }
    \hline

    \methodlabel{Ratio}
    & \textbf{1:1}
    & \textbf{1:1}
    & \textbf{1:1}
    \\
    \hline

    \methodlabel{Original flowchart}
    & \appimgcell{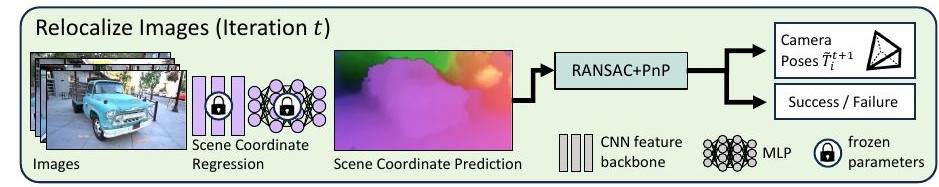}
    & \appimgcell{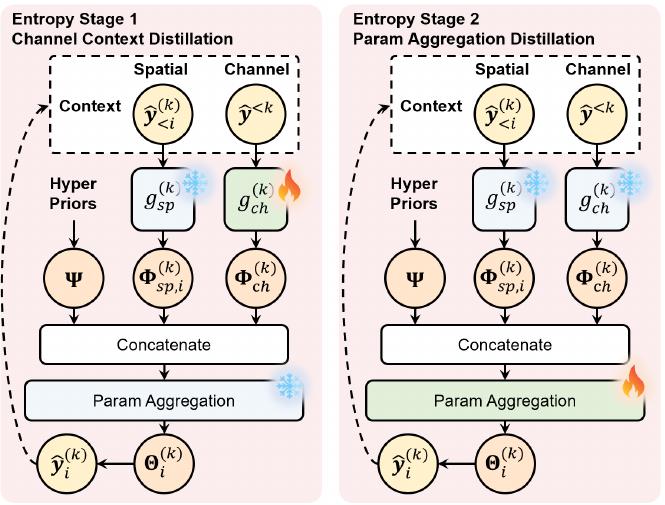}
    & \appimgcell{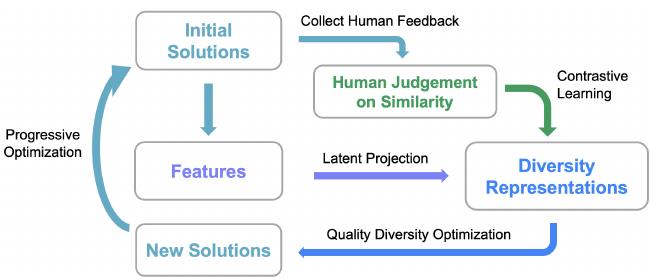}
    \\
    \hline

    \methodlabel{Nano Banana Pro}
    & \appimgcell{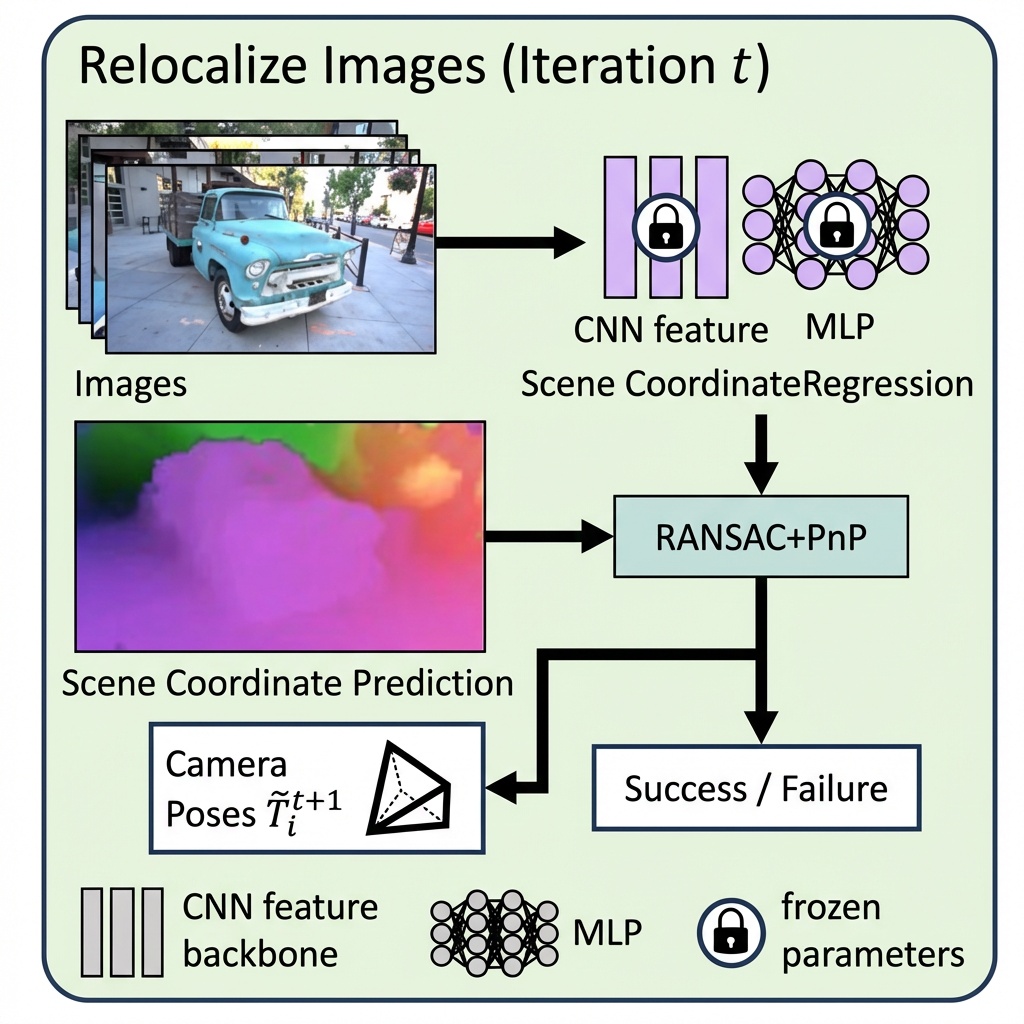}
    & \appimgcell{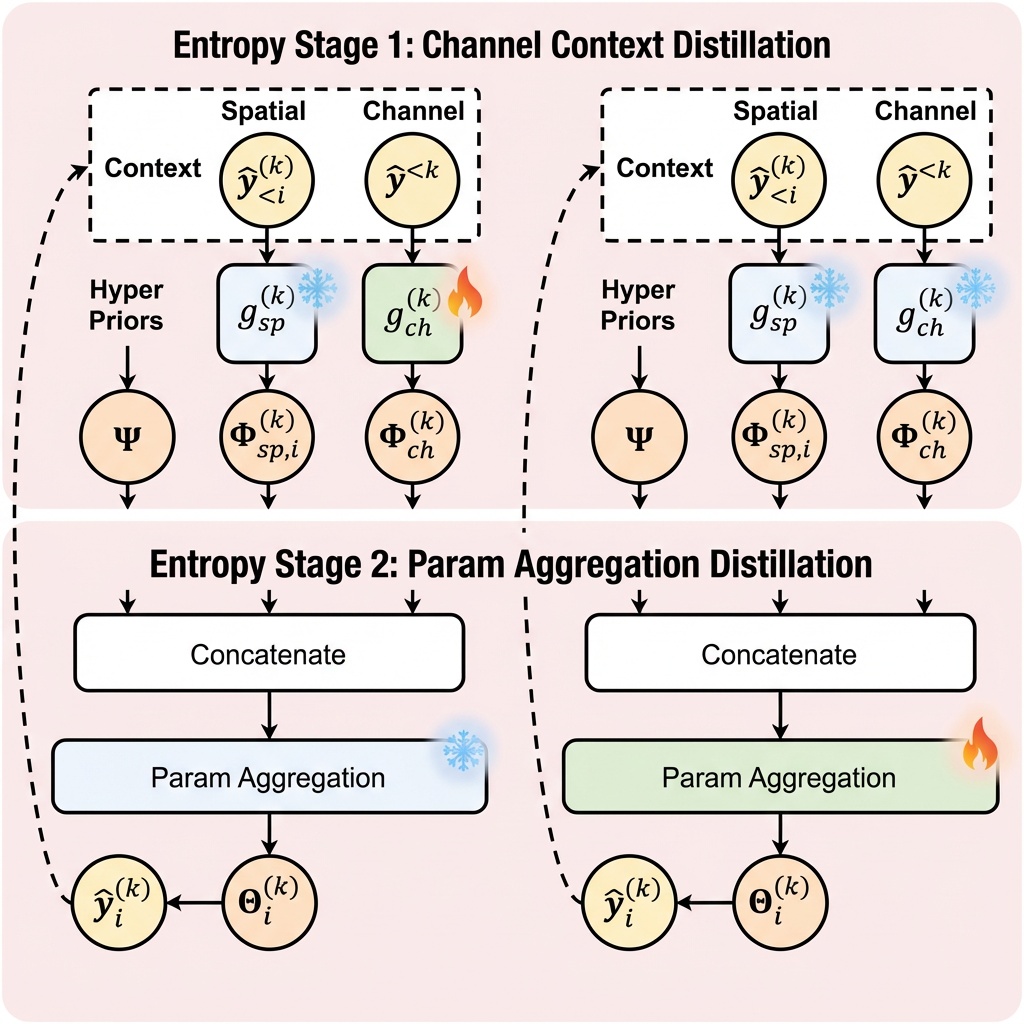}
    & \appimgcell{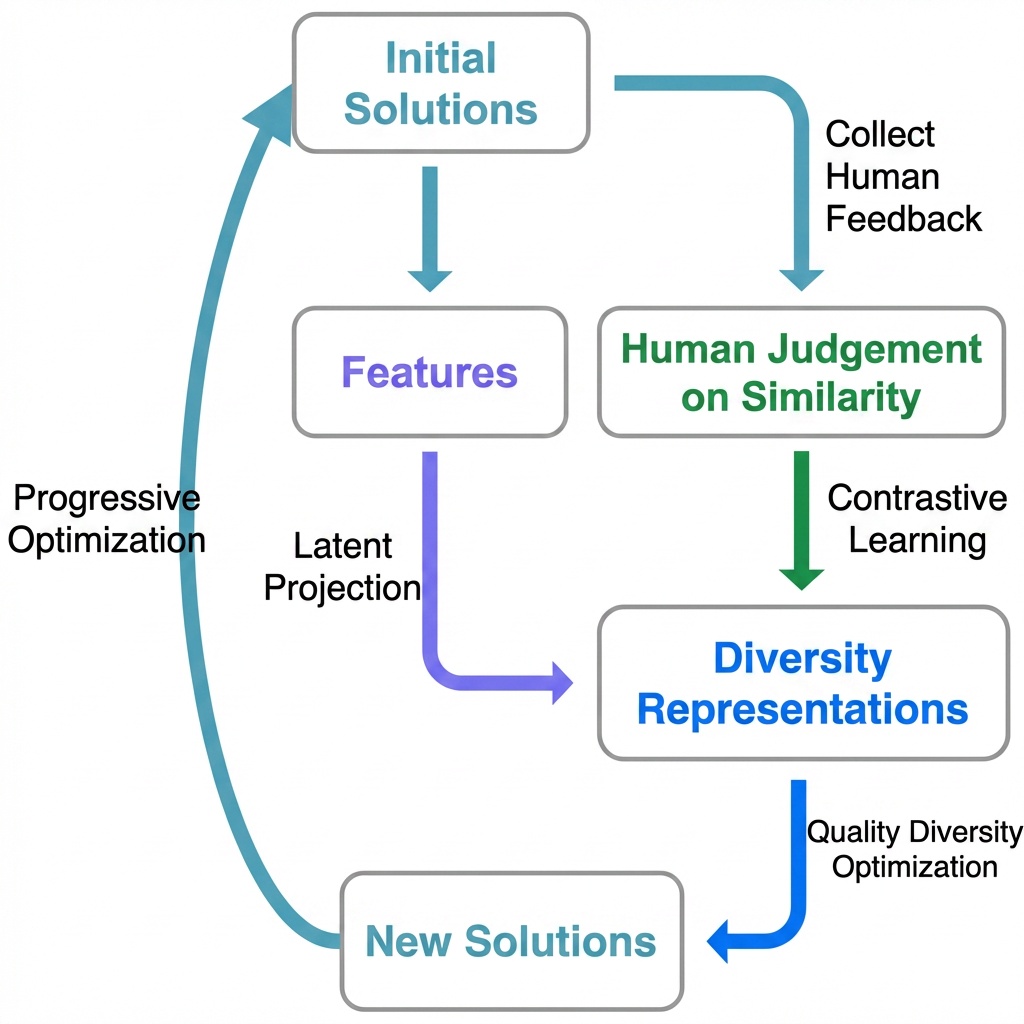}
    \\
    \hline

    \methodlabel{GPT Image 2.0}
    & \appimgcell{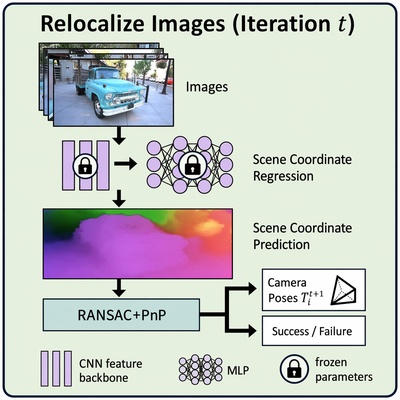}
    & \appimgcell{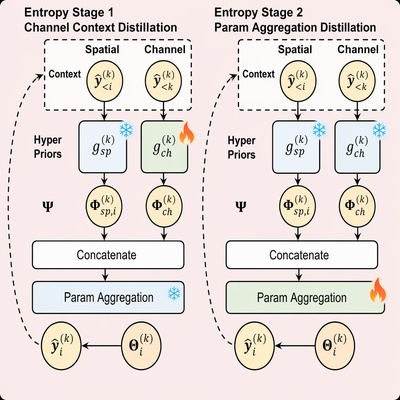}
    & \appimgcell{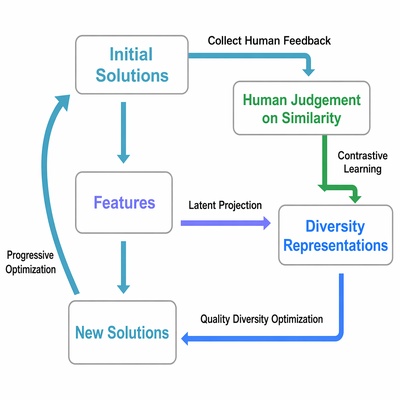}
    \\
    \hline

    \methodlabel{PaperBanana}
    & \appimgcell{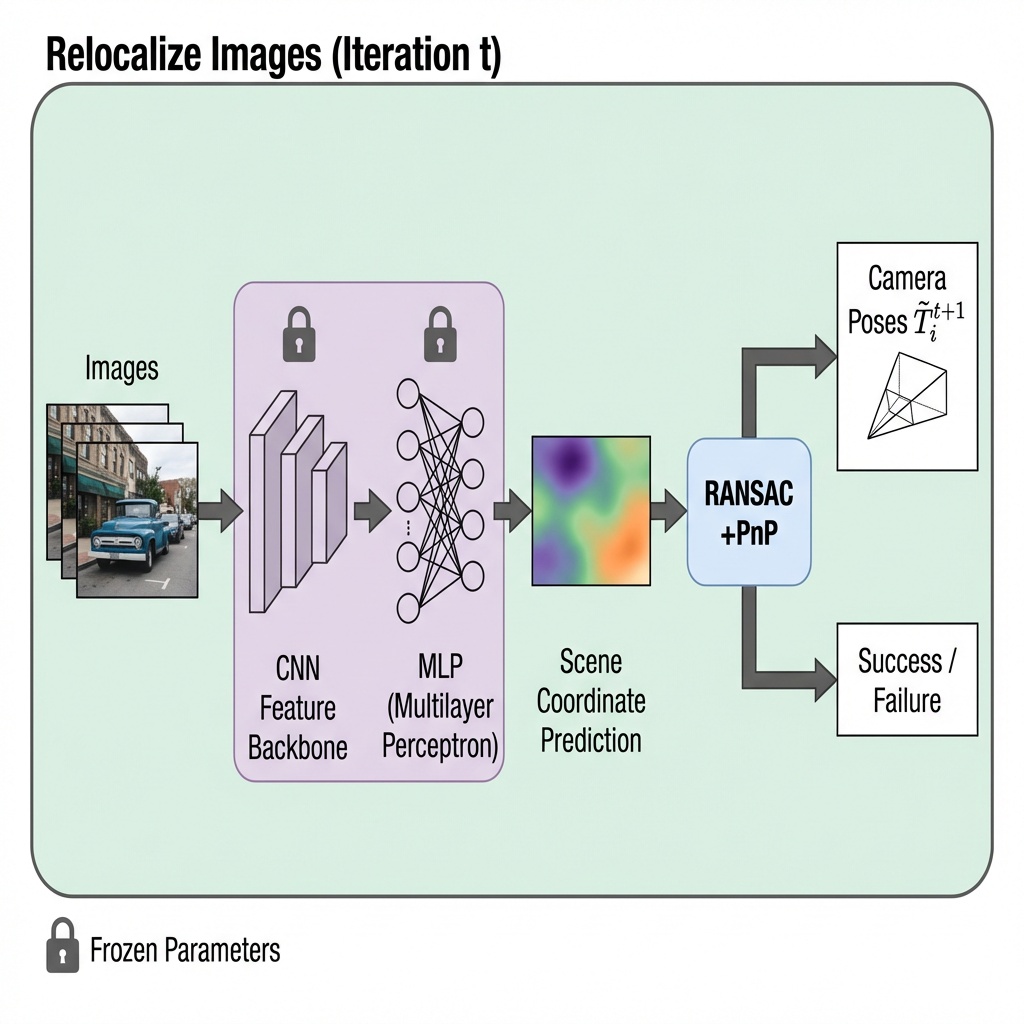}
    & \appimgcell{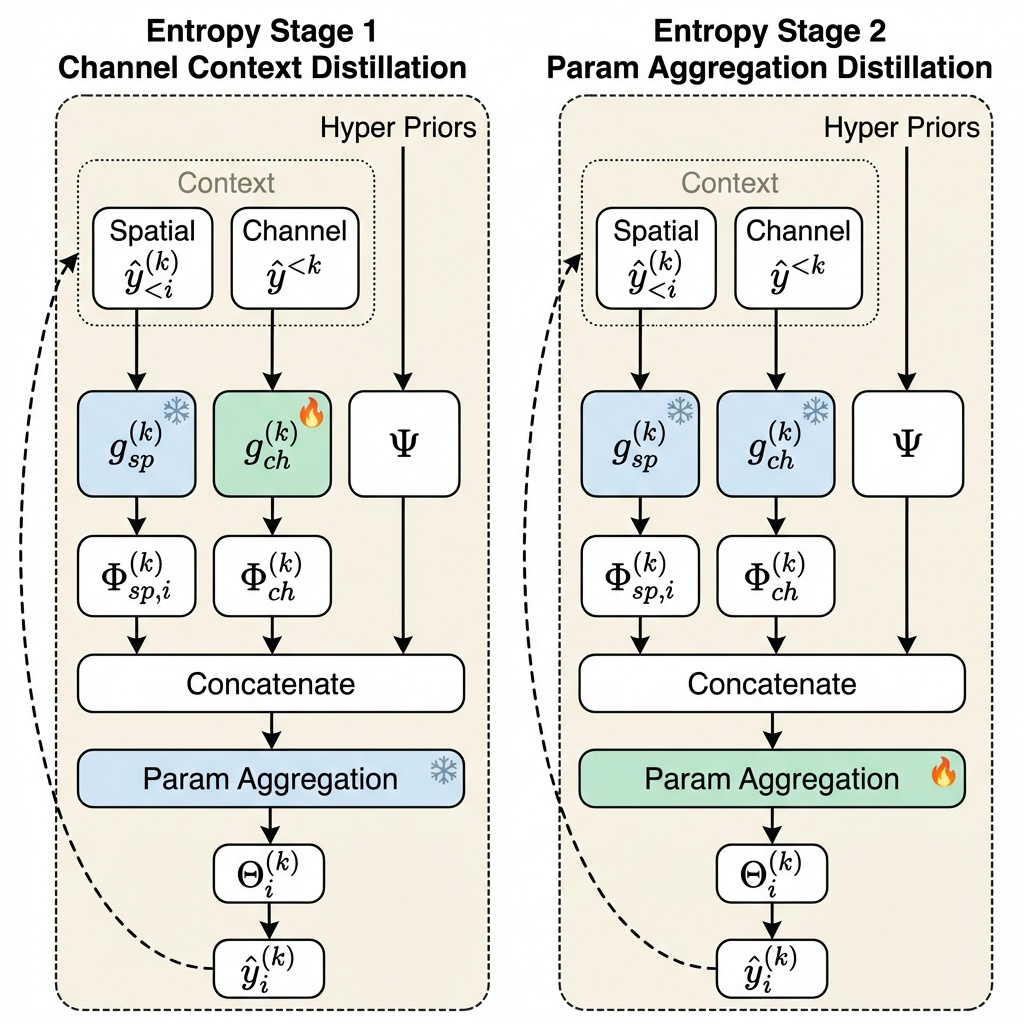}
    & \appimgcell{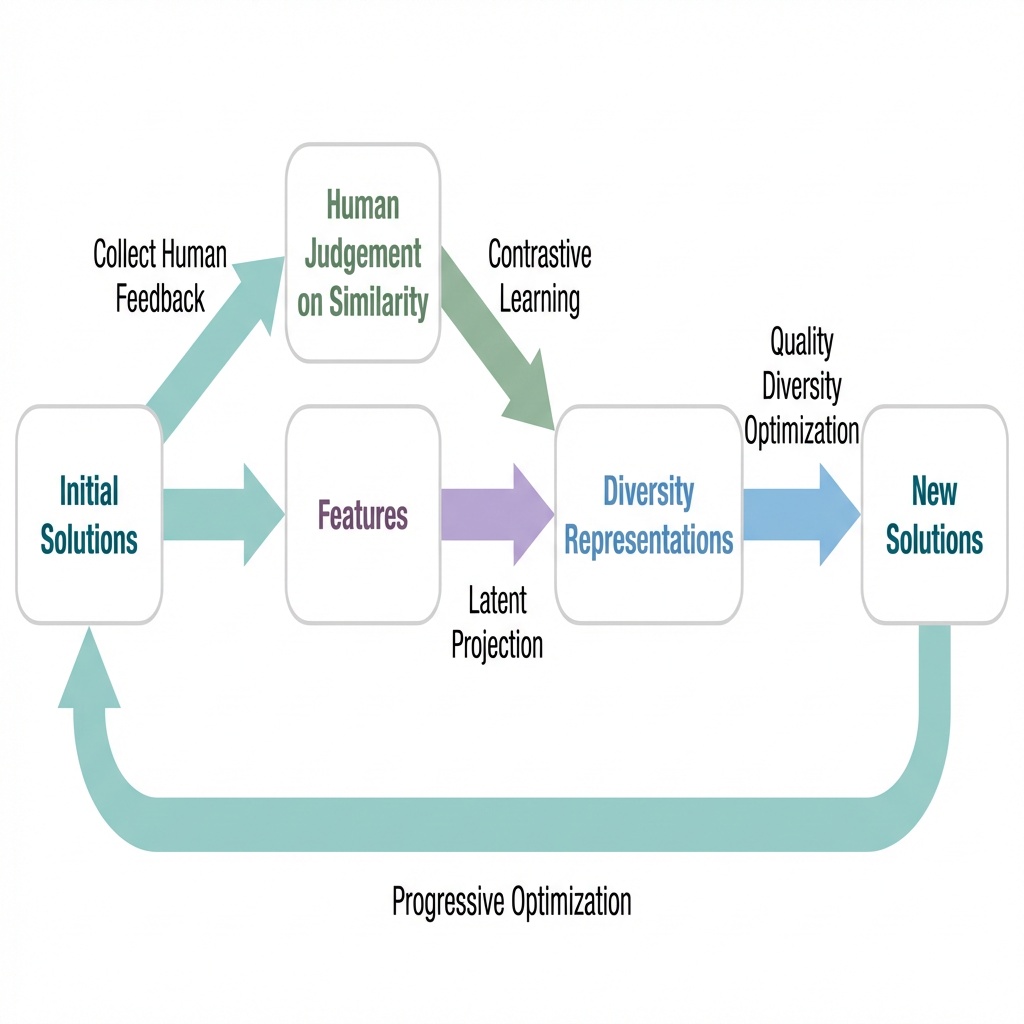}
    \\
    \hline

    \methodlabel{AutoFigure-Edit}
    & \appimgcell{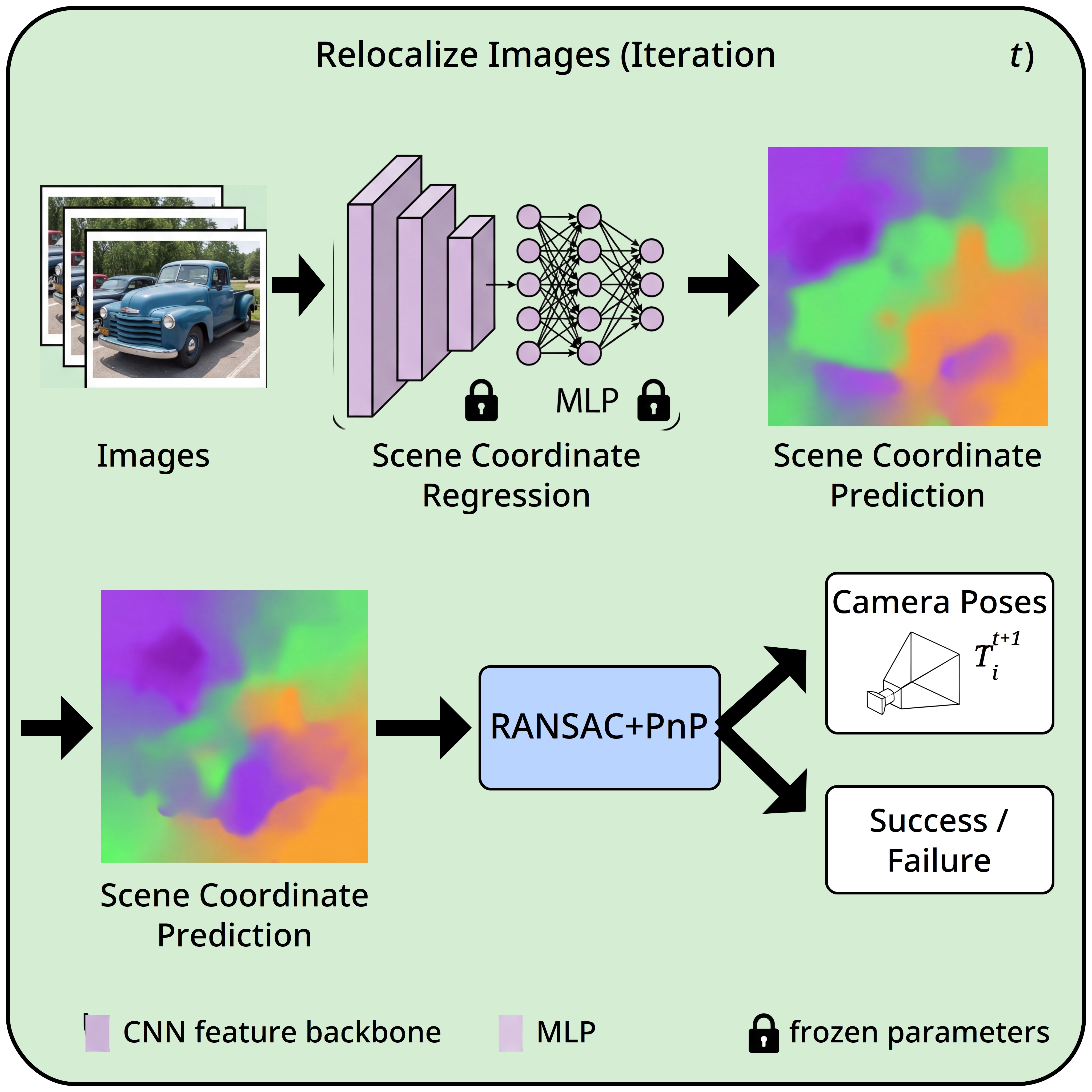}
    & \appimgcell{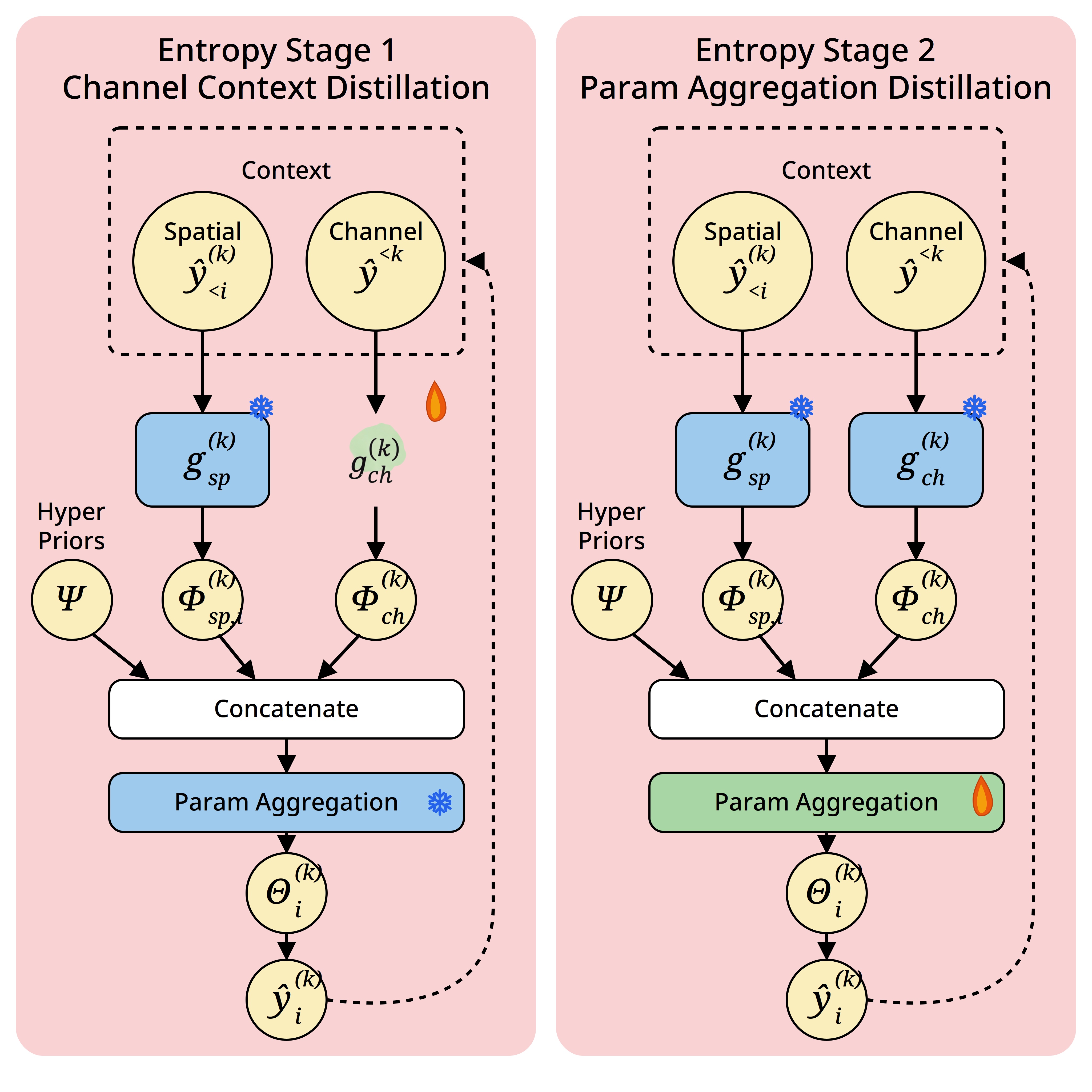}
    & \appimgcell{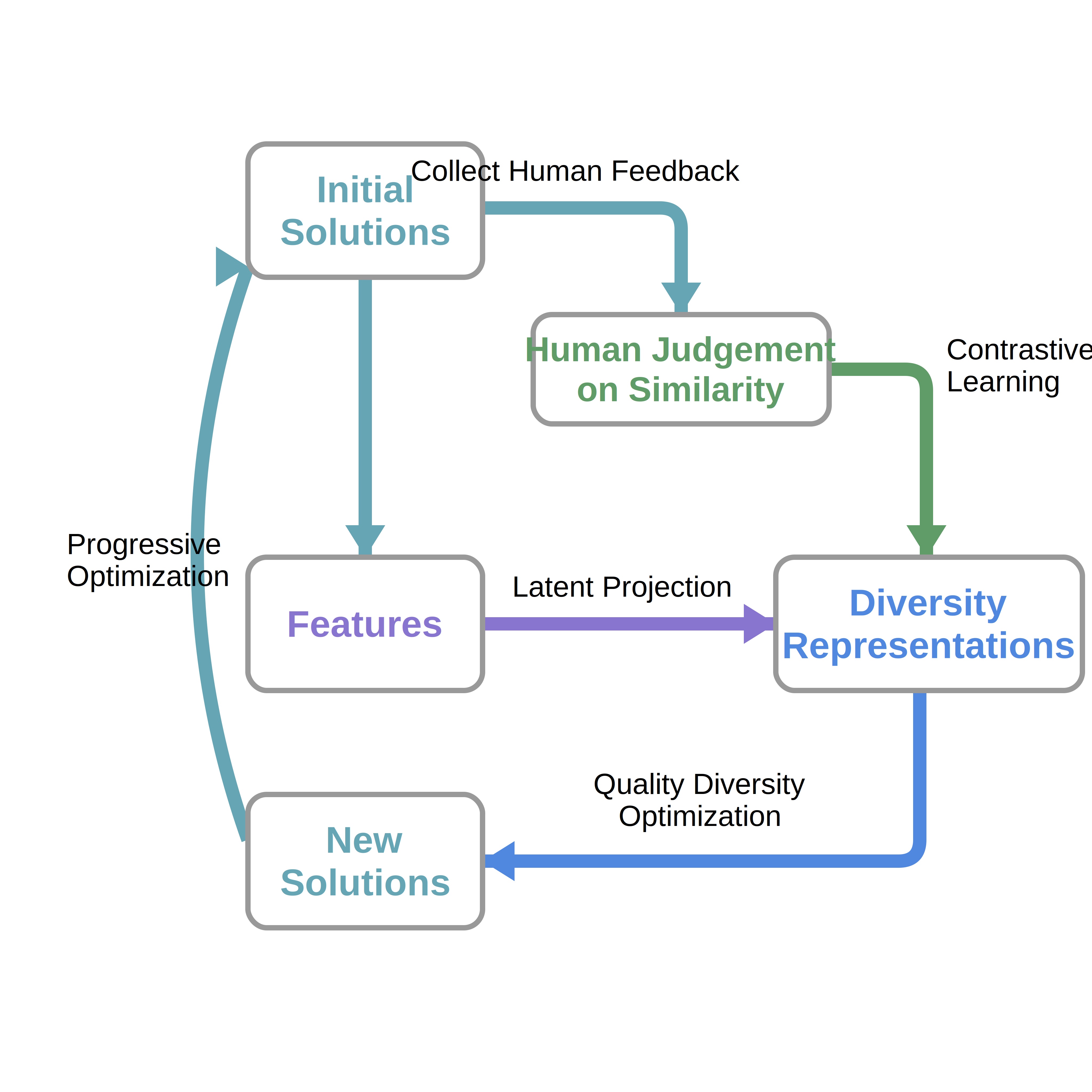}
    \\
    \hline

    \methodlabel{Ours}
    & \appimgcell{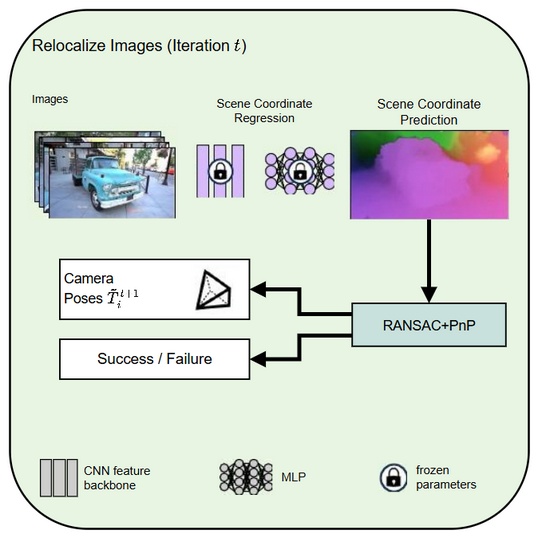}
    & \appimgcell{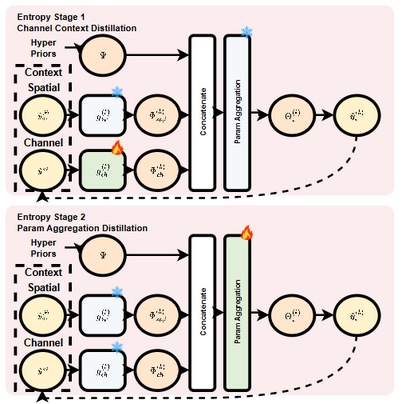}
    & \appimgcell{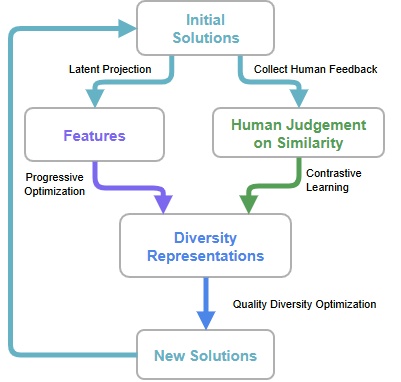}
    \\
    \hline
    \end{tabular}
    }

    \caption{
        \textbf{Additional qualitative comparison under the 1:1 target aspect ratio.}
        In the left example, Nano Banana Pro and GPT Image 2.0 introduce relationship errors when attempting relayout.
        In the middle and right examples, only our method performs clear relayout, while Nano Banana Pro, GPT Image 2.0, and AutoFigure-Edit mainly rely on stretching, compression, or direct reuse of the original image.
    }
    \label{fig:qualitative_comparison_extended1}
    \vspace{-3mm}
\end{figure*}

\begin{figure*}[t]
    \centering
    \small
    \setlength{\tabcolsep}{2pt}
    \renewcommand{\arraystretch}{1.05}

    \newcommand{\AppRowHeight}{0.11\textheight}    %
    \newcommand{\AppImageHeight}{0.10\textheight}  %
    \newcommand{\AppLabelWidth}{0.75cm}            %
    \newcommand{\AppCaseWidth}{3.2cm}              %

    \newcommand{\appimgcell}[1]{%
    \begin{minipage}[c][\AppRowHeight][c]{\linewidth}
    \centering
    \setlength{\fboxsep}{0pt}%
    \setlength{\fboxrule}{0.5pt}%
    \fcolorbox{black}{white}{%
    \includegraphics[
        width=0.97\linewidth,
        height=\AppImageHeight,
        keepaspectratio
    ]{#1}%
    }%
    \end{minipage}
    }

    \newcommand{\methodlabel}[1]{%
    \rotatebox[origin=c]{90}{\textbf{#1}}%
    }

    \resizebox{\textwidth}{!}{%
    \begin{tabular}{
        >{\centering\arraybackslash}m{\AppLabelWidth}
        |>{\centering\arraybackslash}m{\AppCaseWidth}
        |>{\centering\arraybackslash}m{\AppCaseWidth}
        |>{\centering\arraybackslash}m{\AppCaseWidth}
    }
    \hline

    \methodlabel{Ratio}
    & \textbf{2:3}
    & \textbf{2:3}
    & \textbf{2:3}
    \\
    \hline

    \methodlabel{Original flowchart}
    & \appimgcell{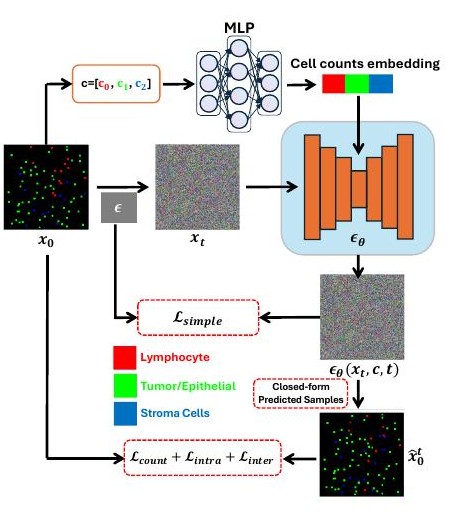}
    & \appimgcell{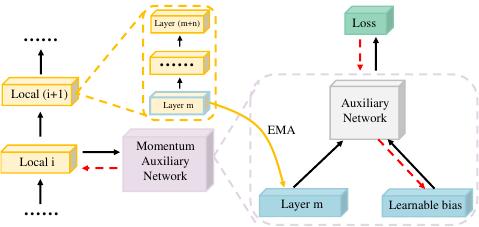}
    & \appimgcell{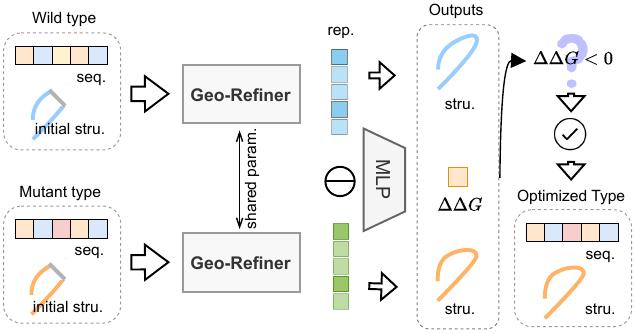}
    \\
    \hline

    \methodlabel{Nano Banana Pro}
    & \appimgcell{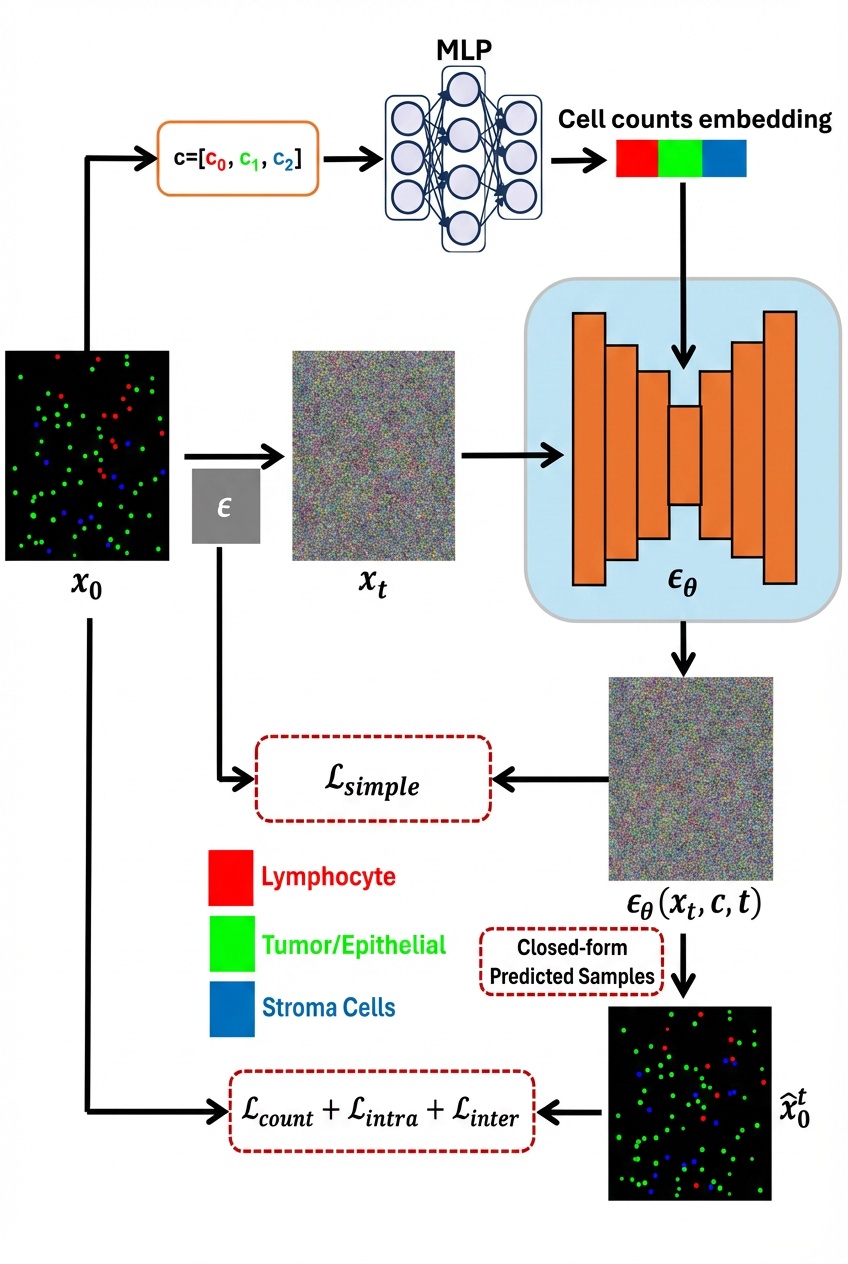}
    & \appimgcell{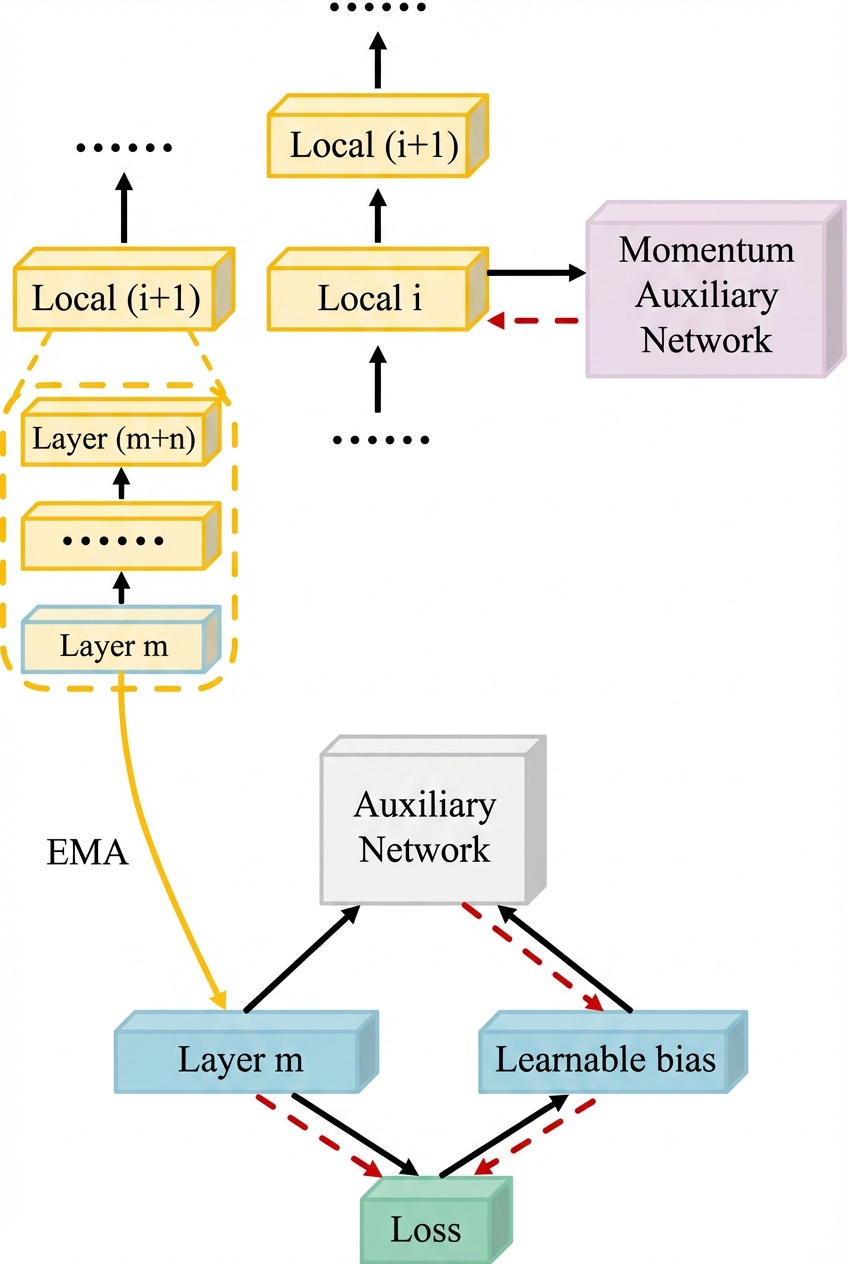}
    & \appimgcell{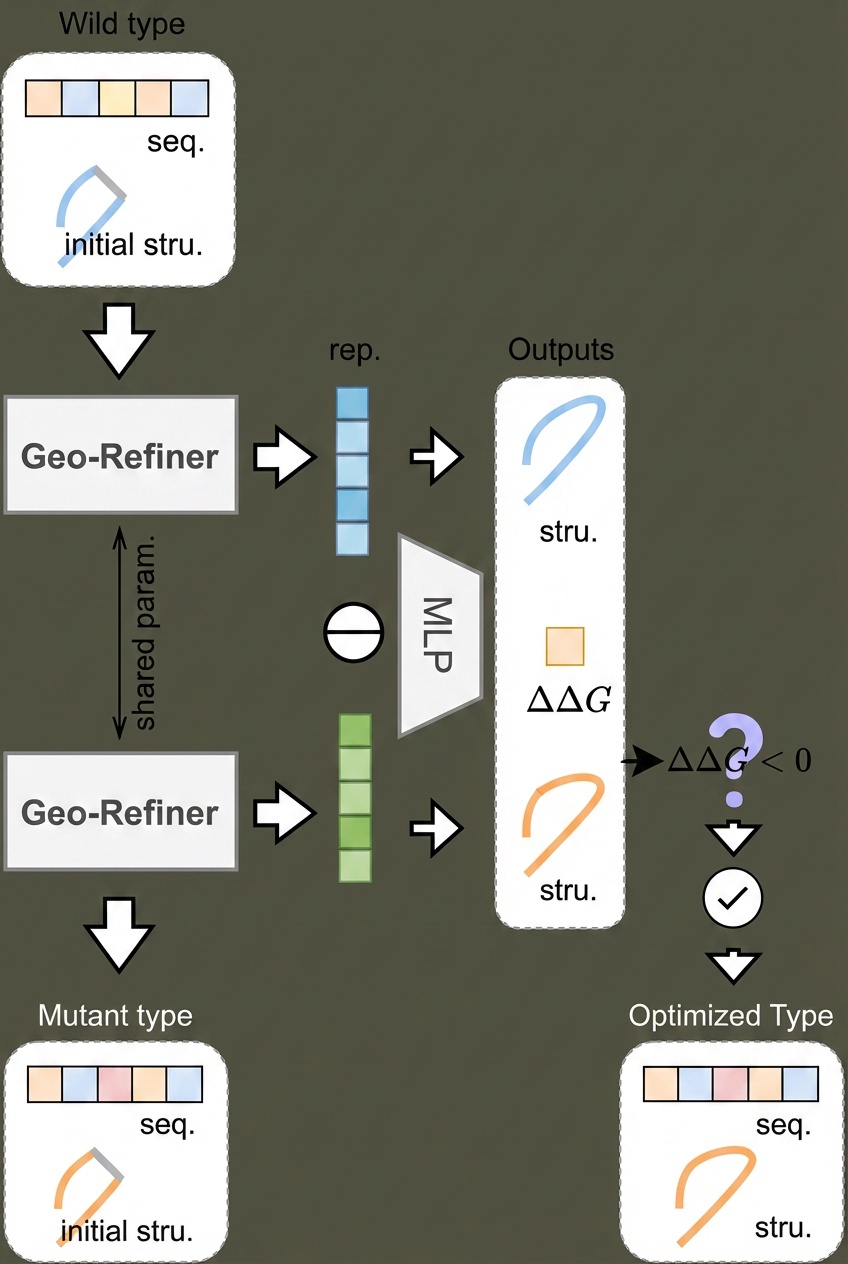}
    \\
    \hline

    \methodlabel{GPT Image 2.0}
    & \appimgcell{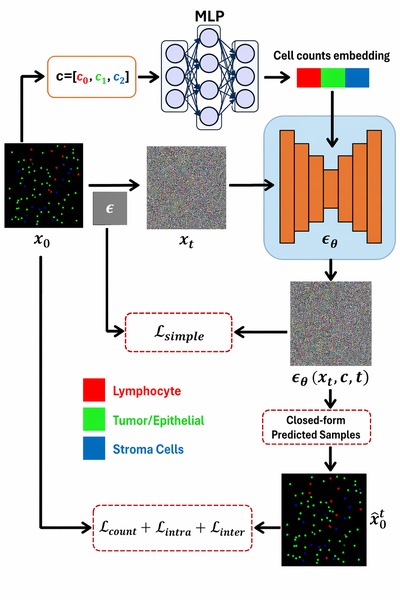}
    & \appimgcell{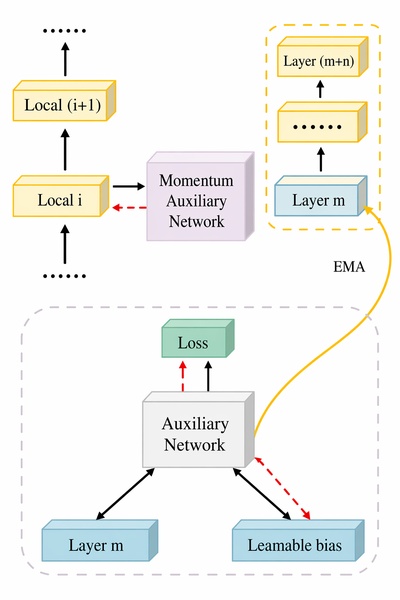}
    & \appimgcell{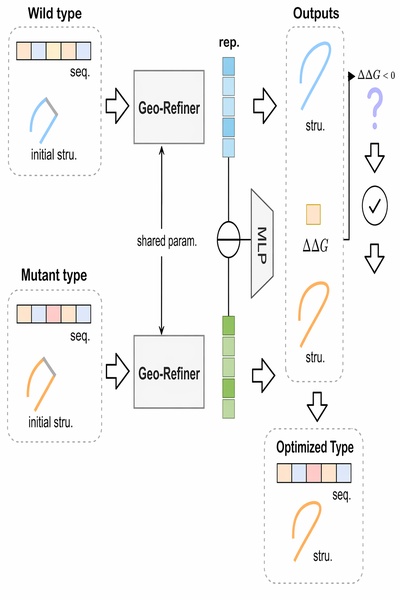}
    \\
    \hline

    \methodlabel{PaperBanana}
    & \appimgcell{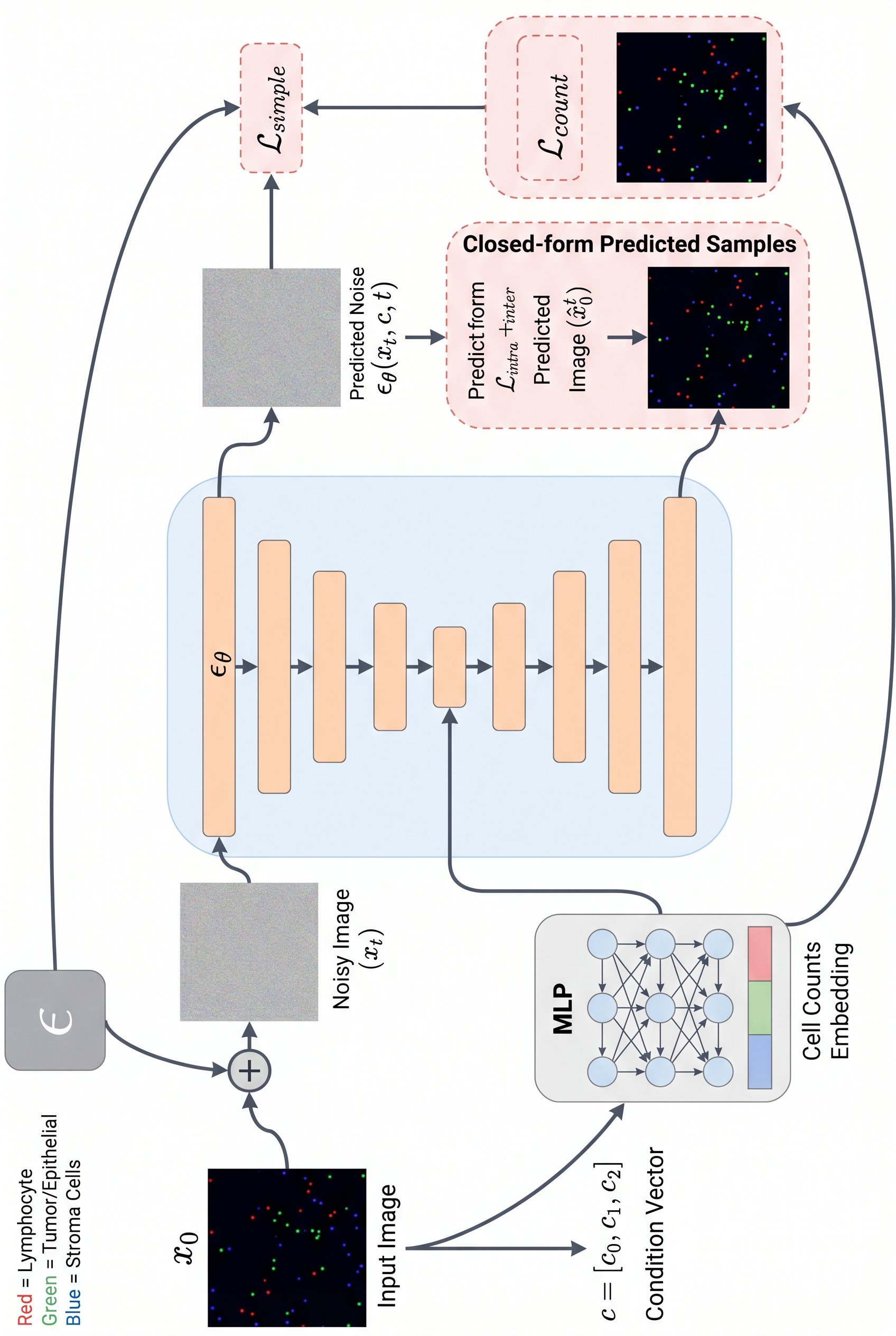}
    & \appimgcell{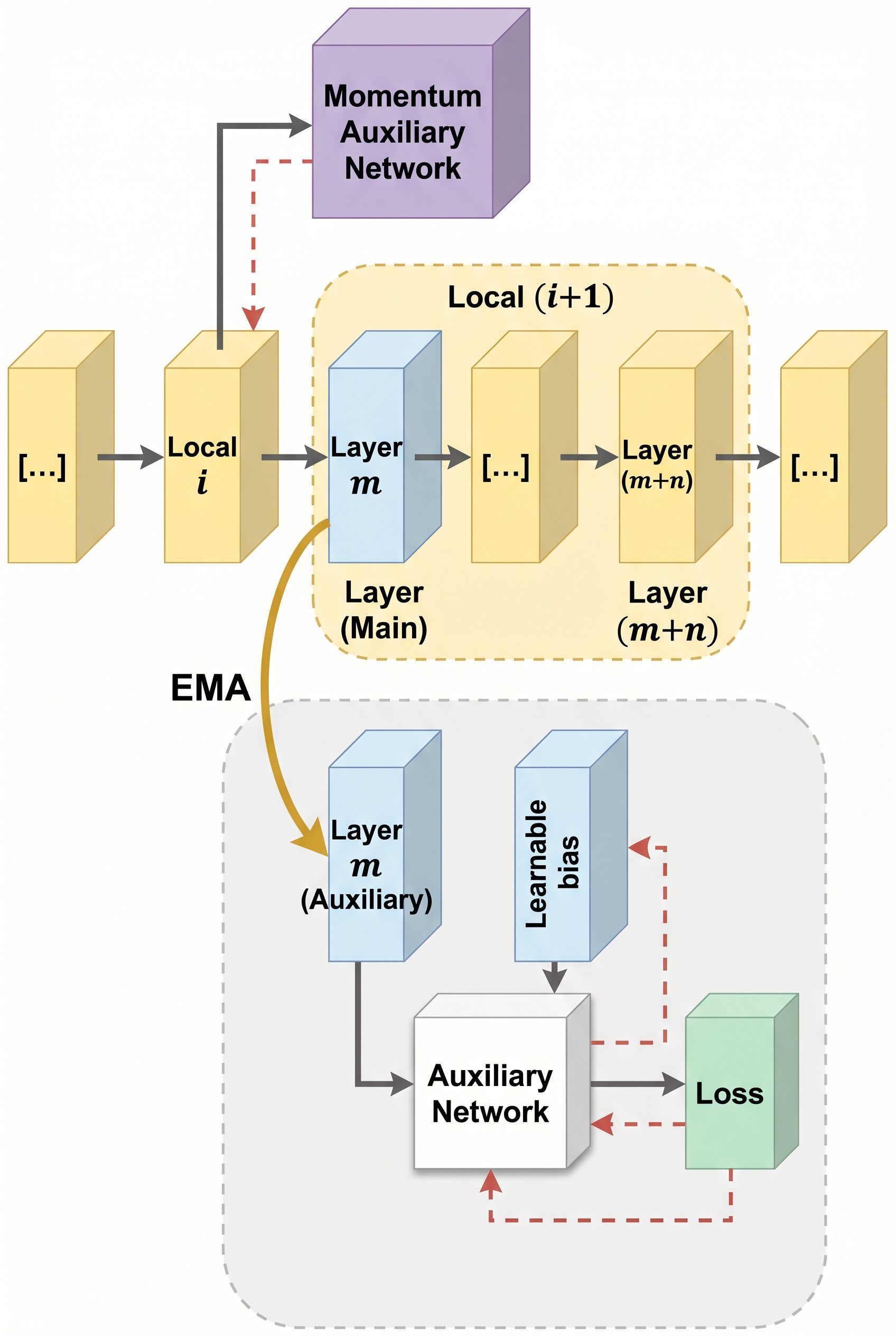}
    & \appimgcell{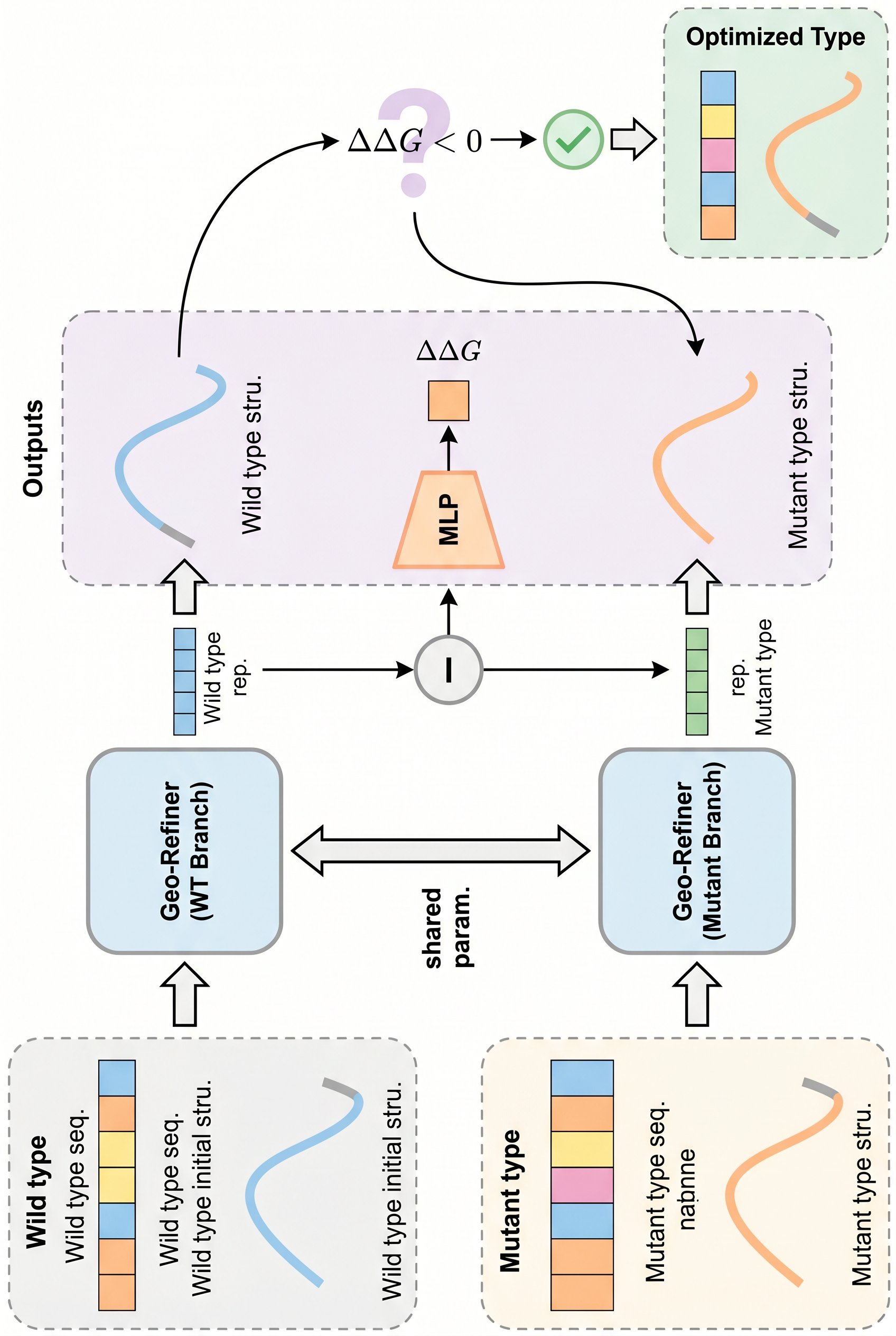}
    \\
    \hline

    \methodlabel{AutoFigure-Edit}
    & \appimgcell{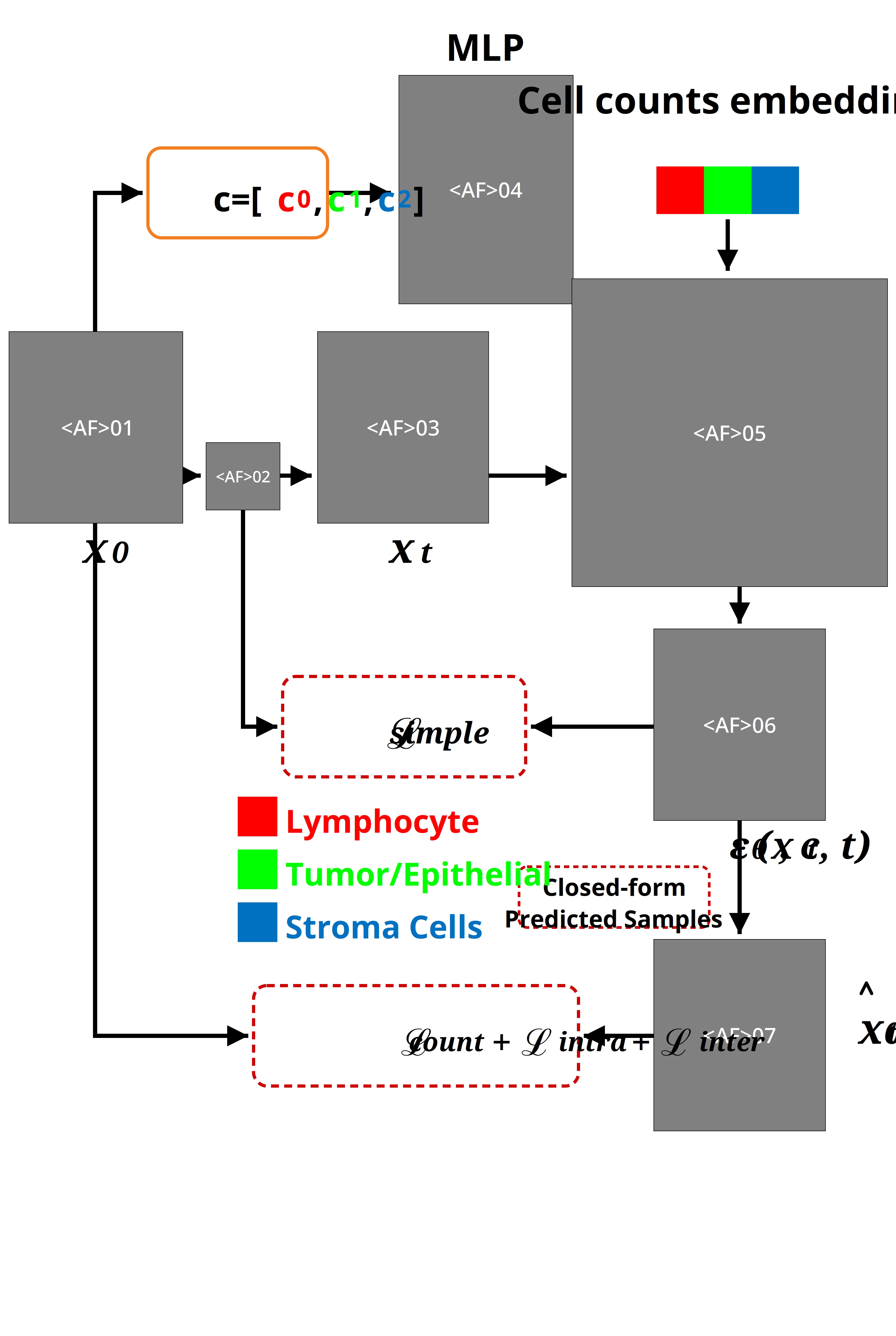}
    & \appimgcell{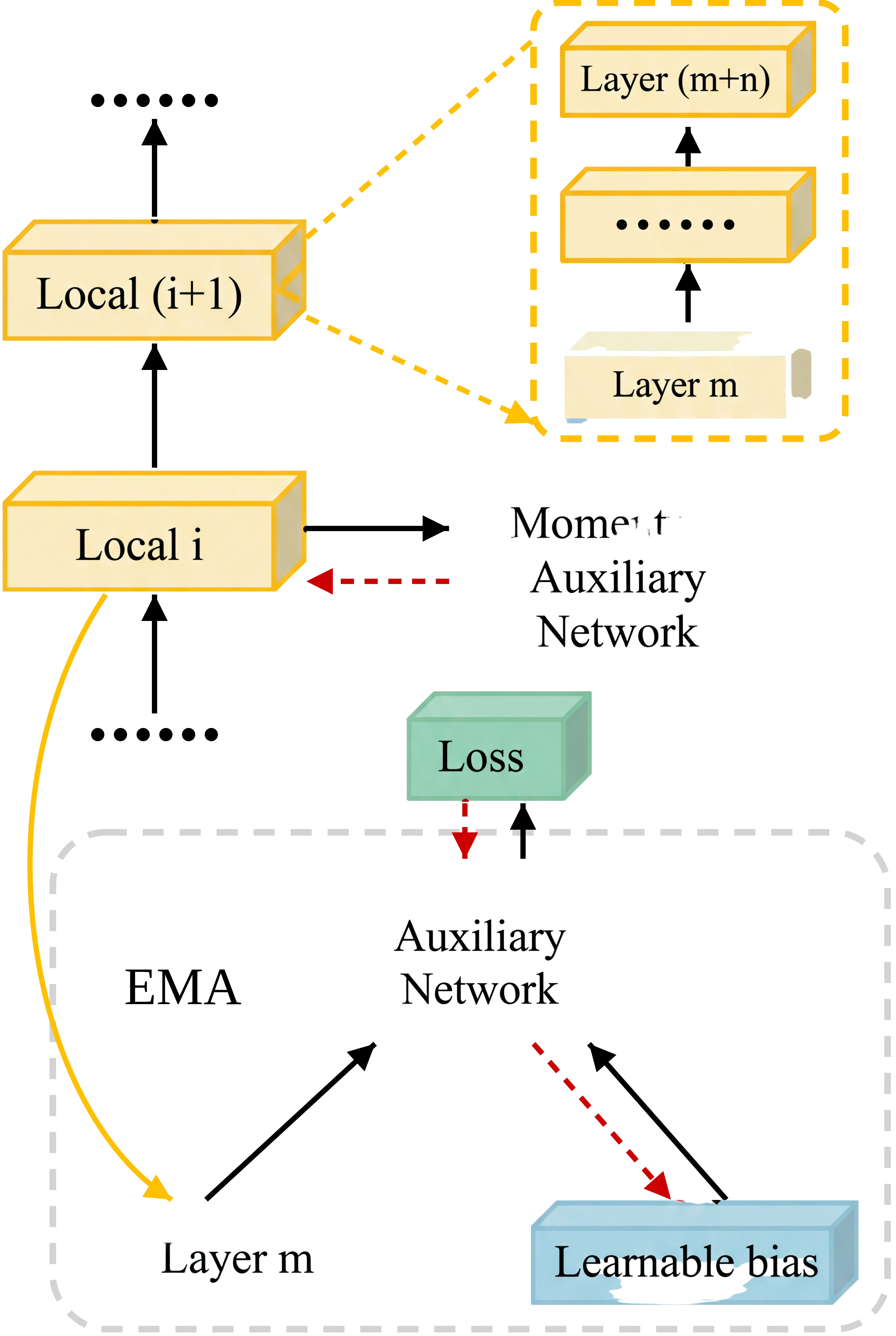}
    & \appimgcell{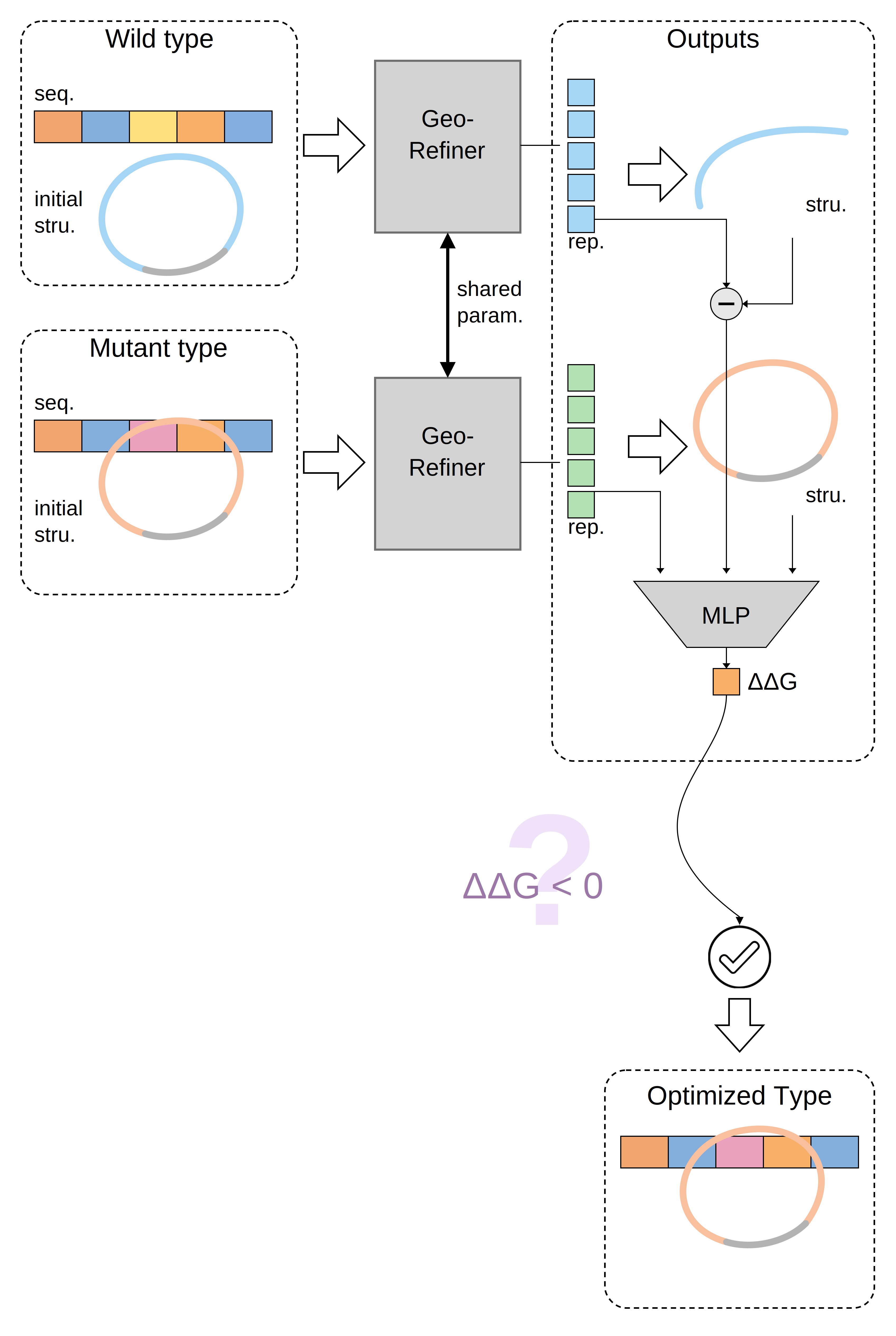}
    \\
    \hline

    \methodlabel{Ours}
    & \appimgcell{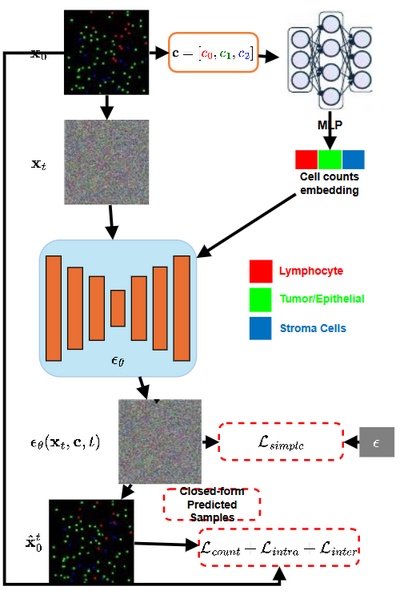}
    & \appimgcell{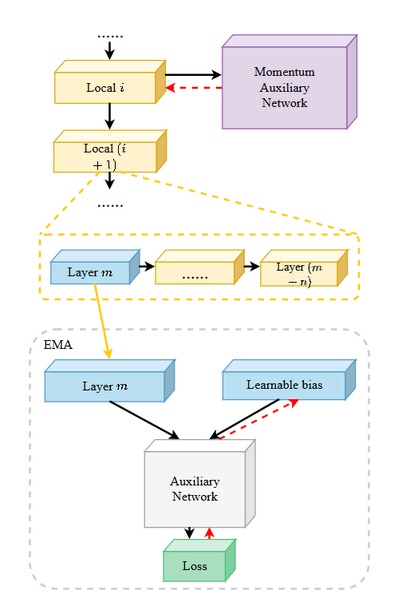}
    & \appimgcell{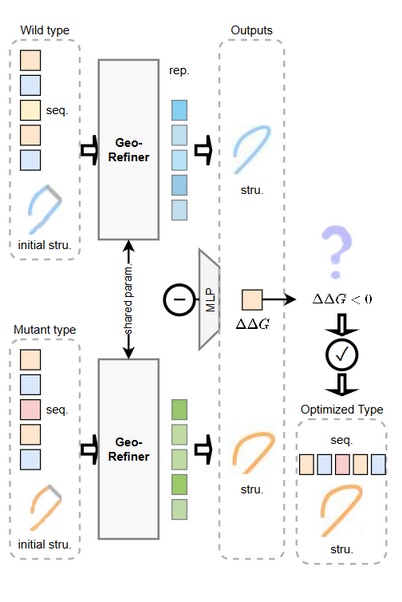}
    \\
    \hline
    \end{tabular}
    }

   \caption{
        \textbf{Additional qualitative comparison under the 2:3 target aspect ratio.}
        In the left example, Nano Banana Pro and GPT Image 2.0 mostly stretch the original image without meaningful relayout.
        In the middle example, all baselines introduce relationship errors after relayout.
        In the right example, GPT Image 2.0 and AutoFigure-Edit leave large unused space, while PaperBanana rotates a horizontal layout to fit the portrait canvas.
    }
    \label{fig:qualitative_comparison_extended2}
    \vspace{-3mm}
\end{figure*}

\begin{figure*}[t]
    \centering
    \small
    \setlength{\tabcolsep}{2pt}
    \renewcommand{\arraystretch}{1.05}

    \newcommand{\AppRowHeight}{0.11\textheight}    %
    \newcommand{\AppImageHeight}{0.10\textheight}  %
    \newcommand{\AppLabelWidth}{0.75cm}            %
    \newcommand{\AppCaseWidth}{3.2cm}              %

    \newcommand{\appimgcell}[1]{%
    \begin{minipage}[c][\AppRowHeight][c]{\linewidth}
    \centering
    \setlength{\fboxsep}{0pt}%
    \setlength{\fboxrule}{0.5pt}%
    \fcolorbox{black}{white}{%
    \includegraphics[
        width=0.97\linewidth,
        height=\AppImageHeight,
        keepaspectratio
    ]{#1}%
    }%
    \end{minipage}
    }

    \newcommand{\methodlabel}[1]{%
    \rotatebox[origin=c]{90}{\textbf{#1}}%
    }

    \resizebox{\textwidth}{!}{%
    \begin{tabular}{
        >{\centering\arraybackslash}m{\AppLabelWidth}
        |>{\centering\arraybackslash}m{\AppCaseWidth}
        |>{\centering\arraybackslash}m{\AppCaseWidth}
        |>{\centering\arraybackslash}m{\AppCaseWidth}
    }
    \hline

    \methodlabel{Ratio}
    & \textbf{3:2}
    & \textbf{3:2}
    & \textbf{3:2}
    \\
    \hline

    \methodlabel{Original flowchart}
    & \appimgcell{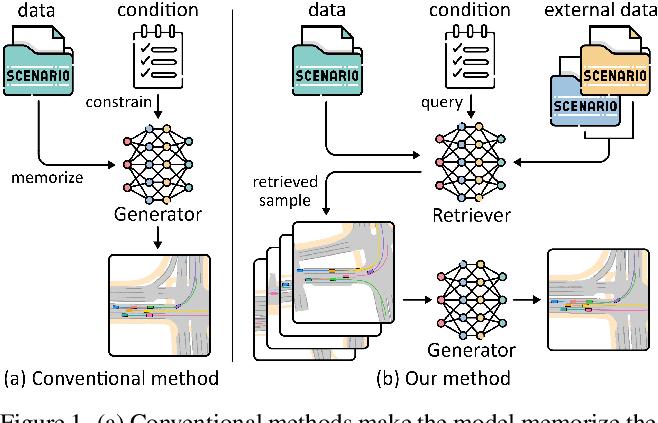}
    & \appimgcell{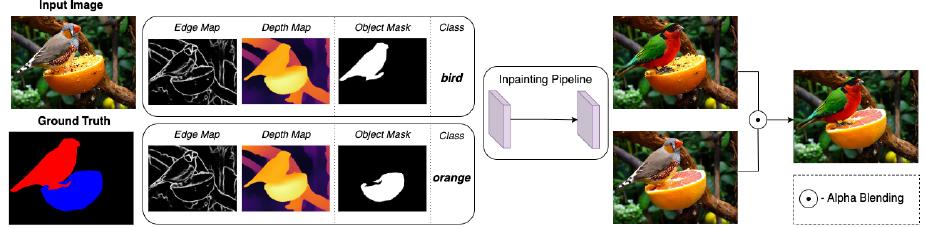}
    & \appimgcell{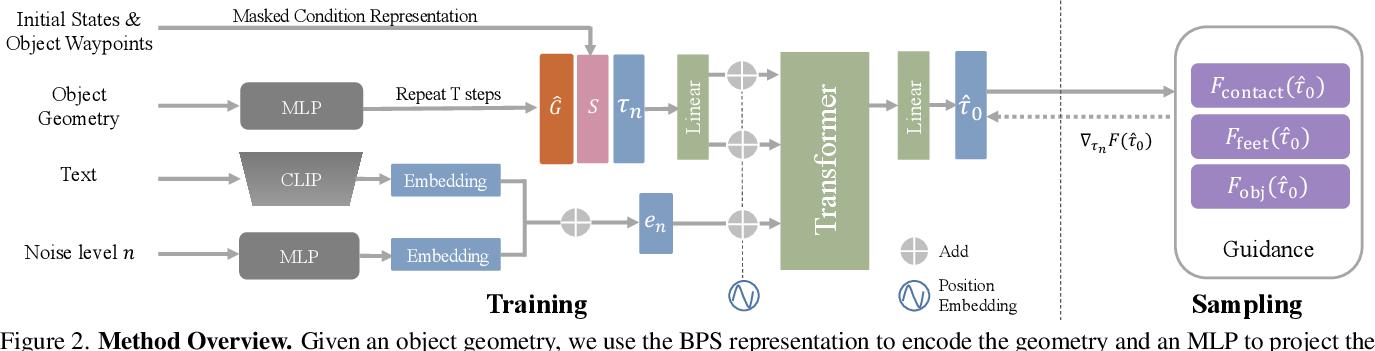}
    \\
    \hline

    \methodlabel{Nano Banana Pro}
    & \appimgcell{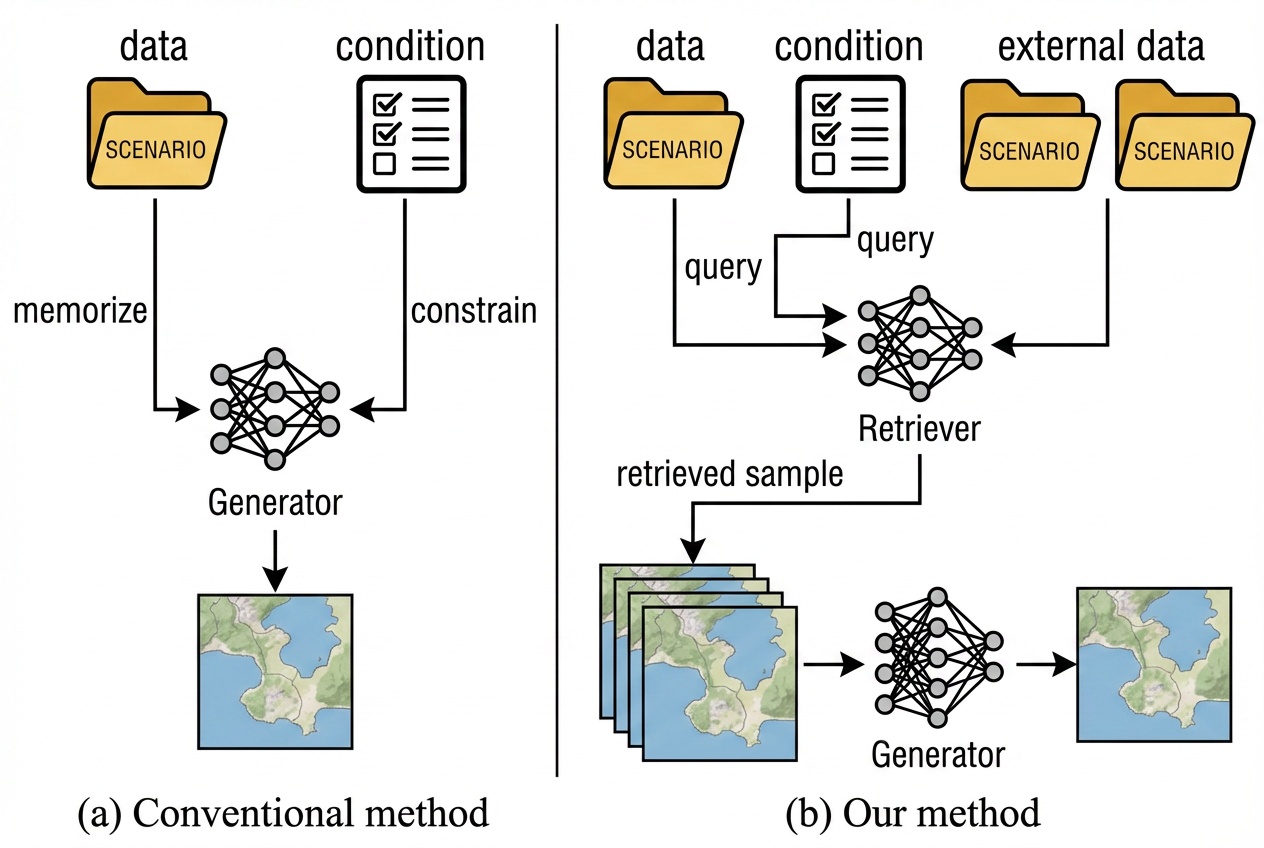}
    & \appimgcell{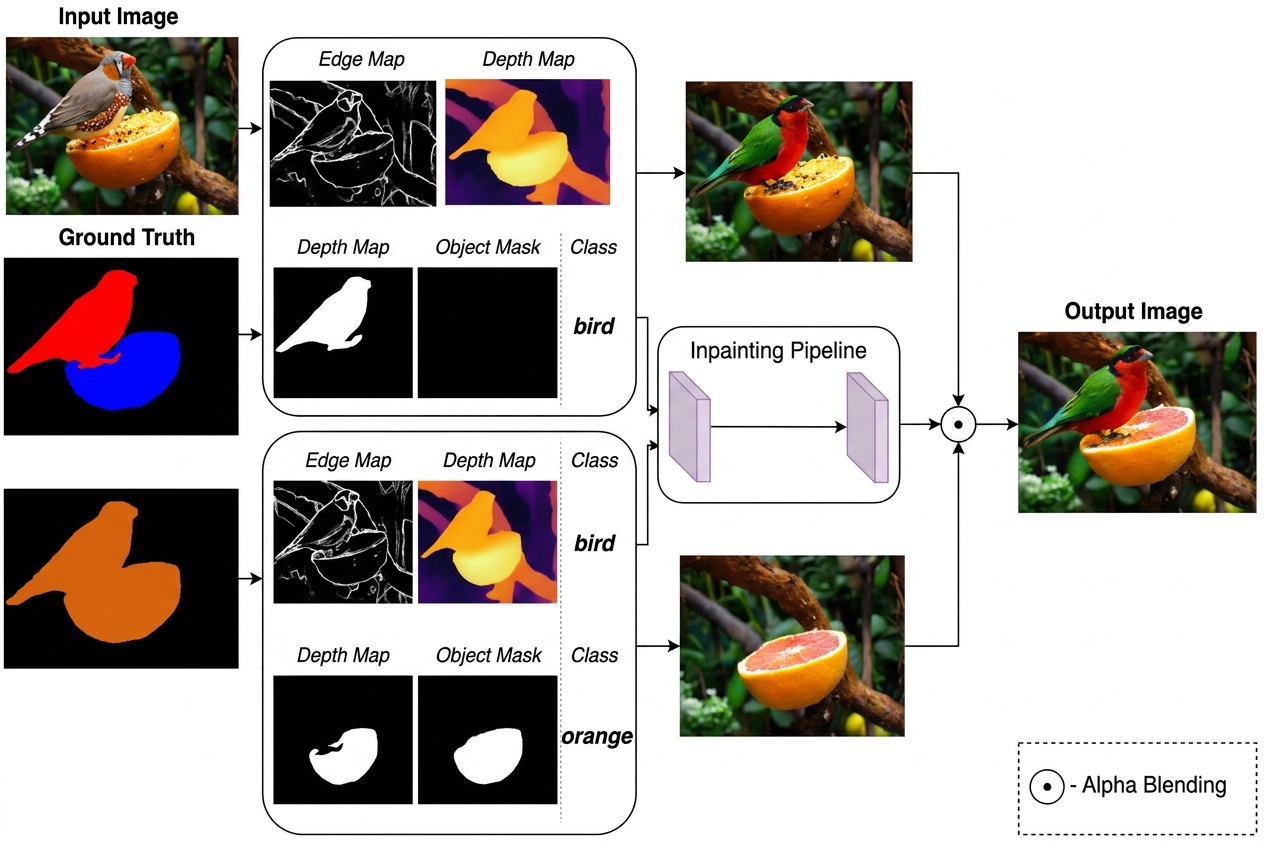}
    & \appimgcell{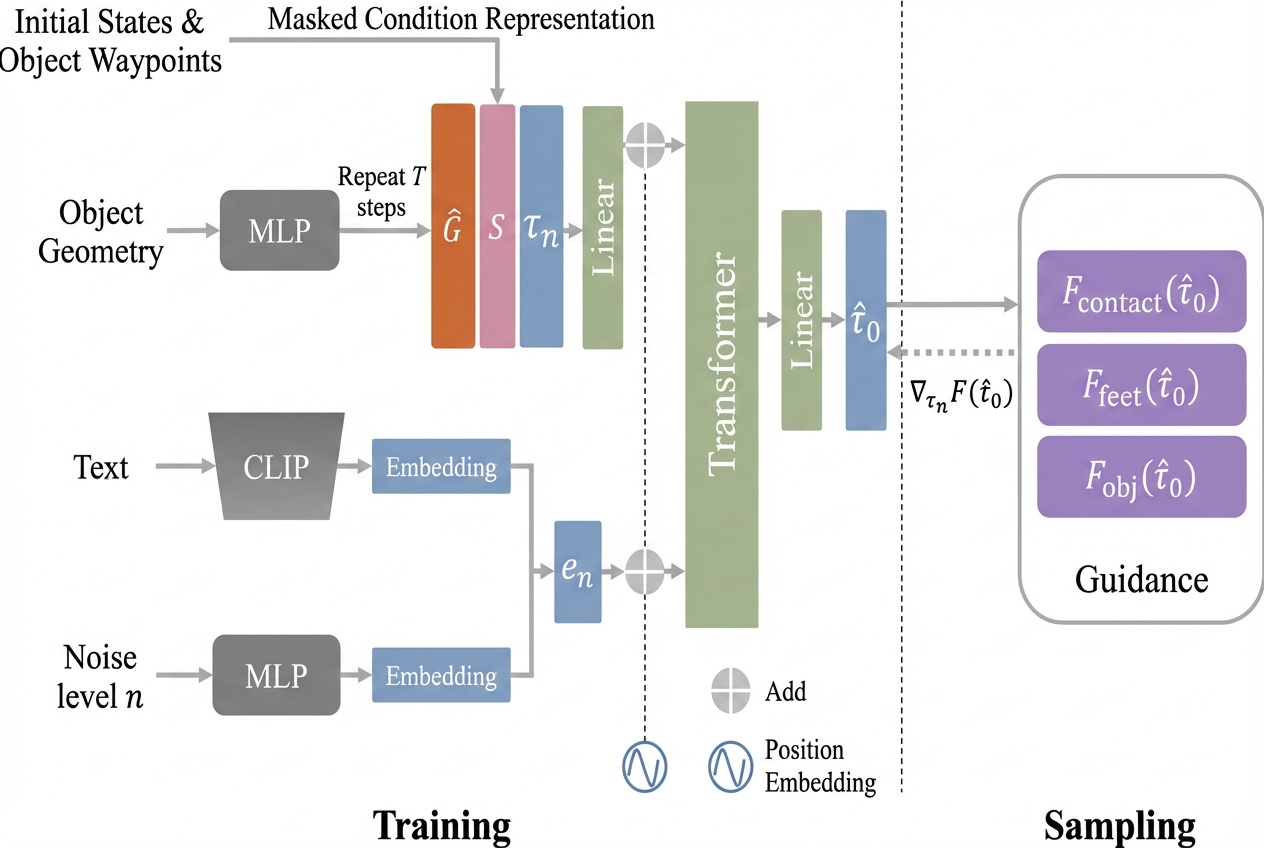}
    \\
    \hline

    \methodlabel{GPT Image 2.0}
    & \appimgcell{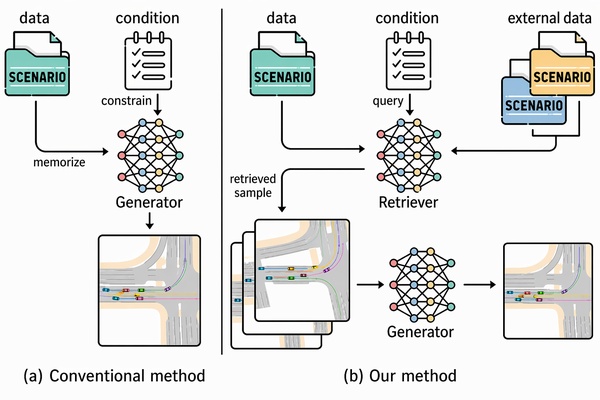}
    & \appimgcell{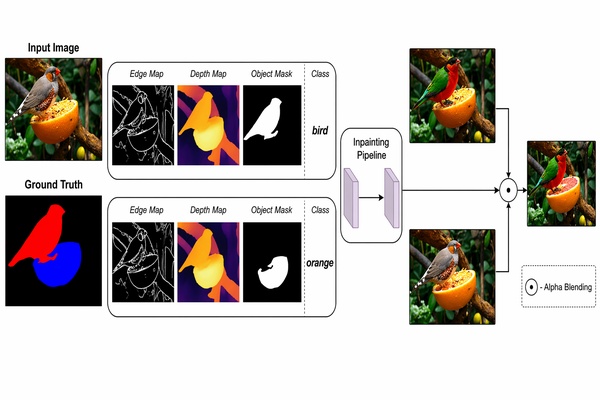}
    & \appimgcell{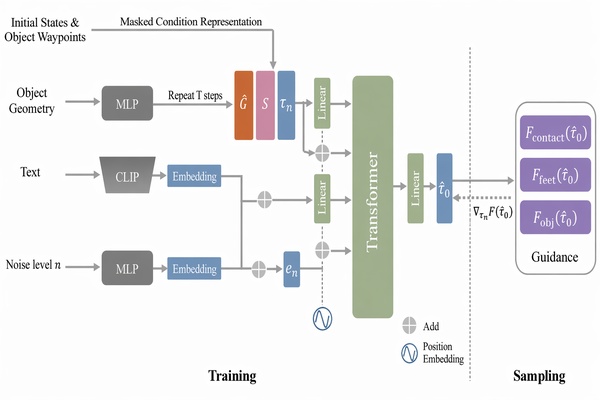}
    \\
    \hline

    \methodlabel{PaperBanana}
    & \appimgcell{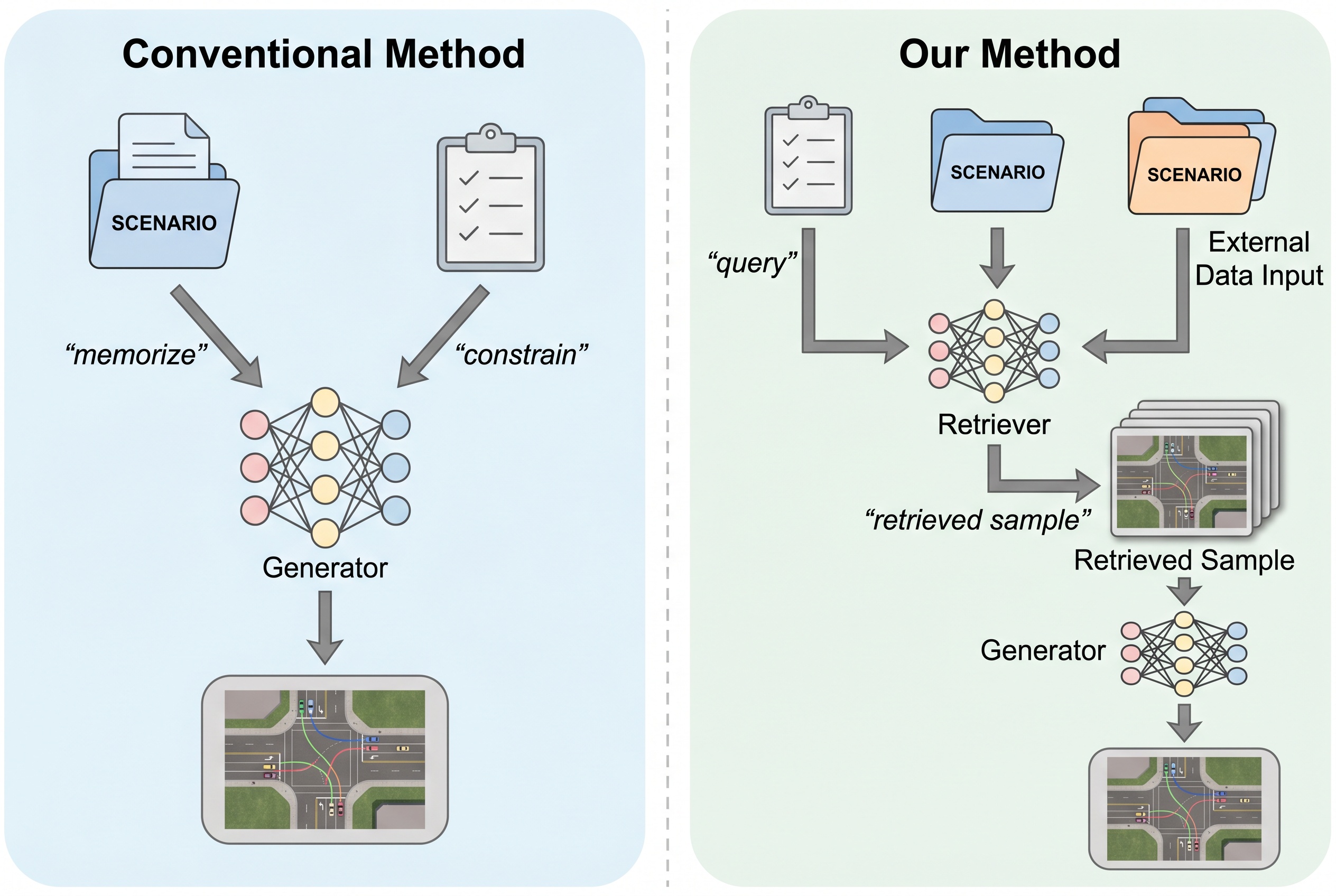}
    & \appimgcell{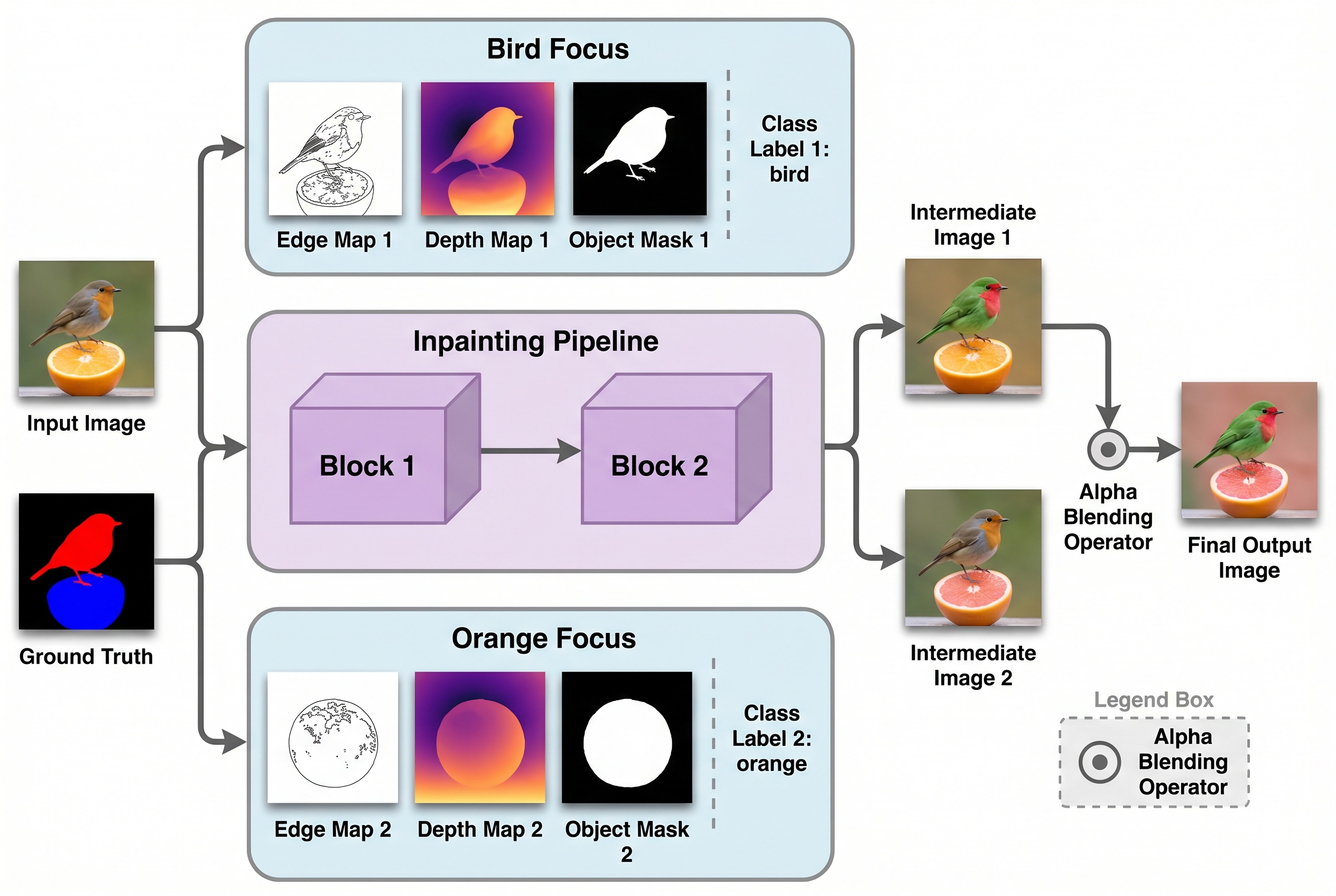}
    & \appimgcell{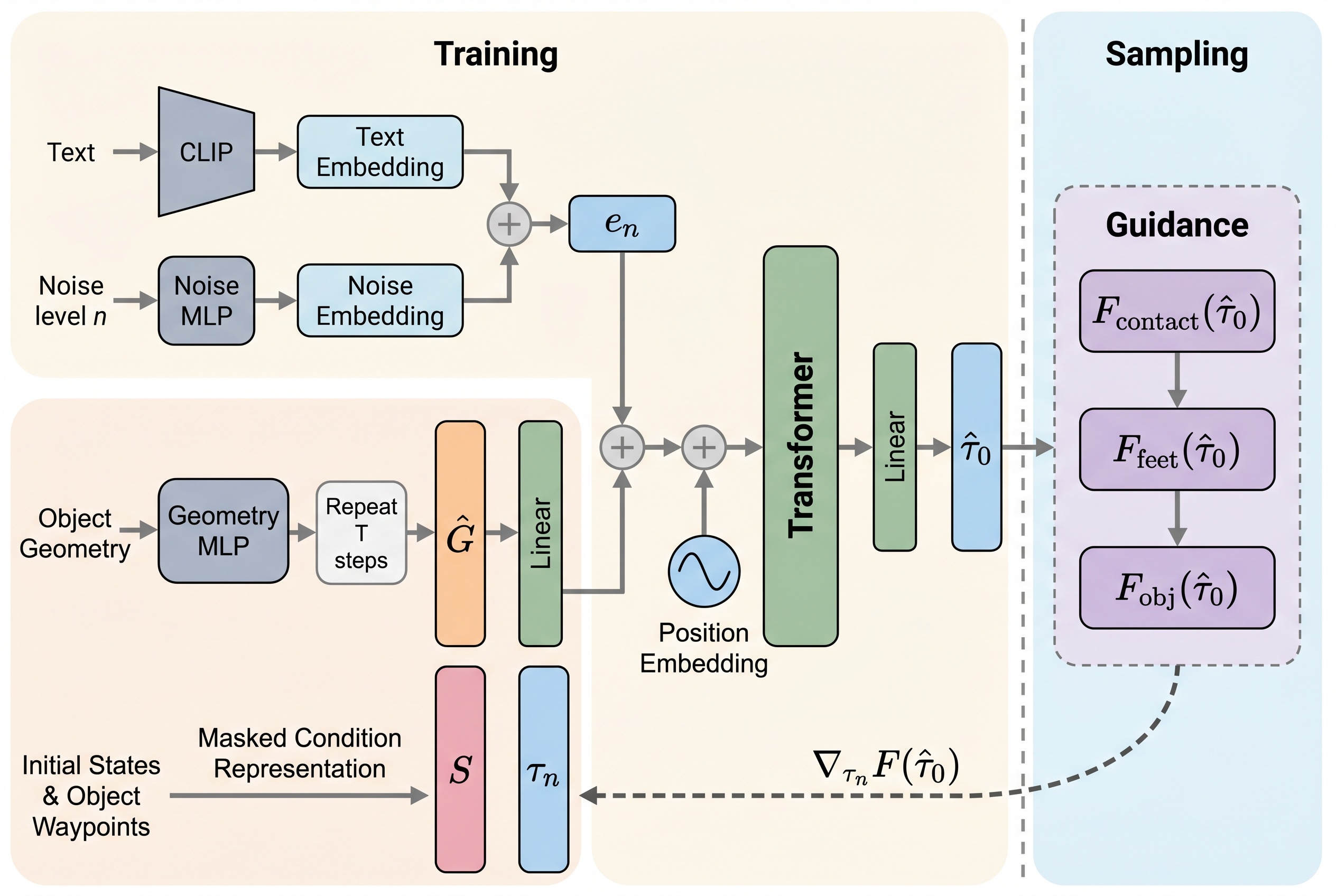}
    \\
    \hline

    \methodlabel{AutoFigure-Edit}
    & \appimgcell{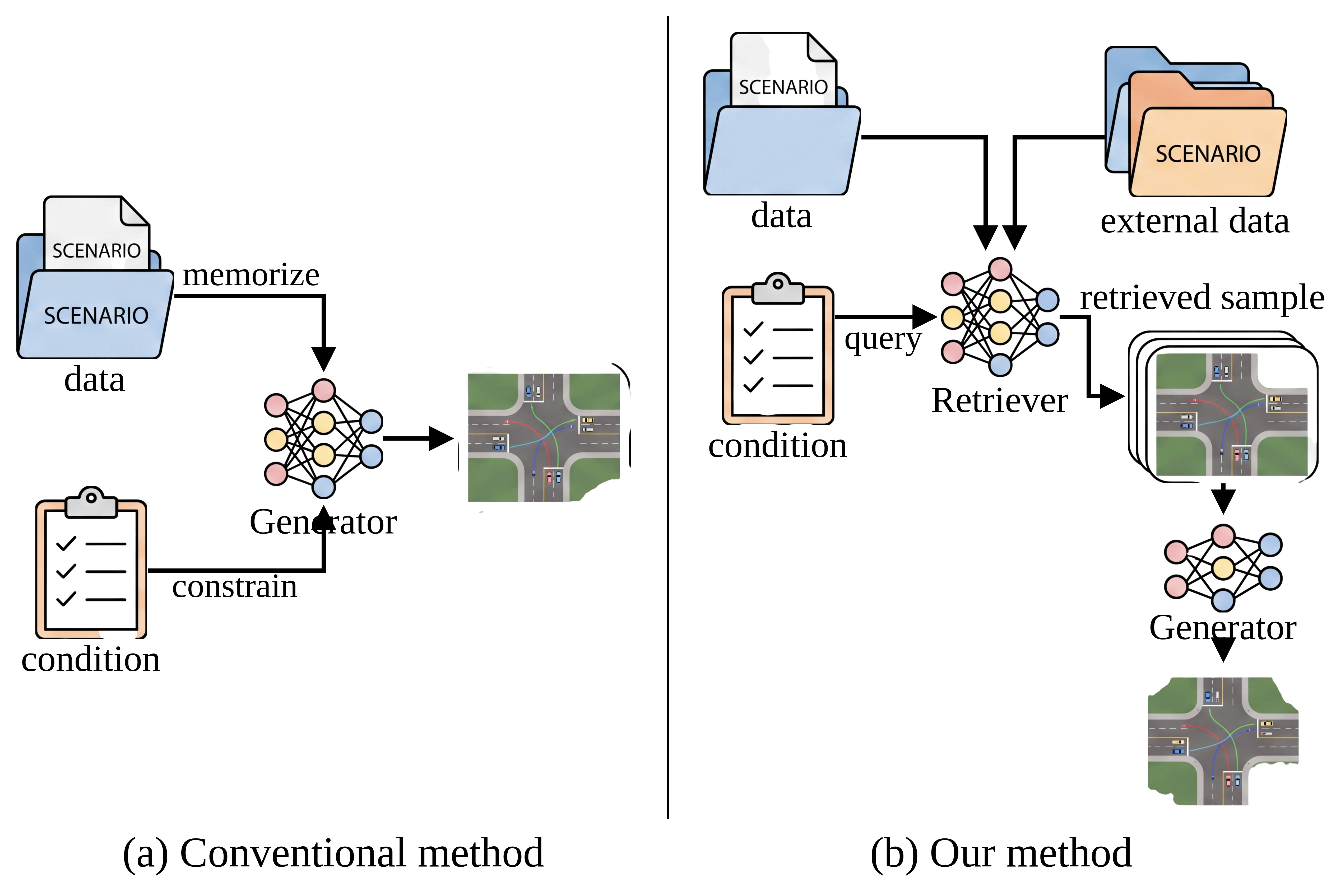}
    & \appimgcell{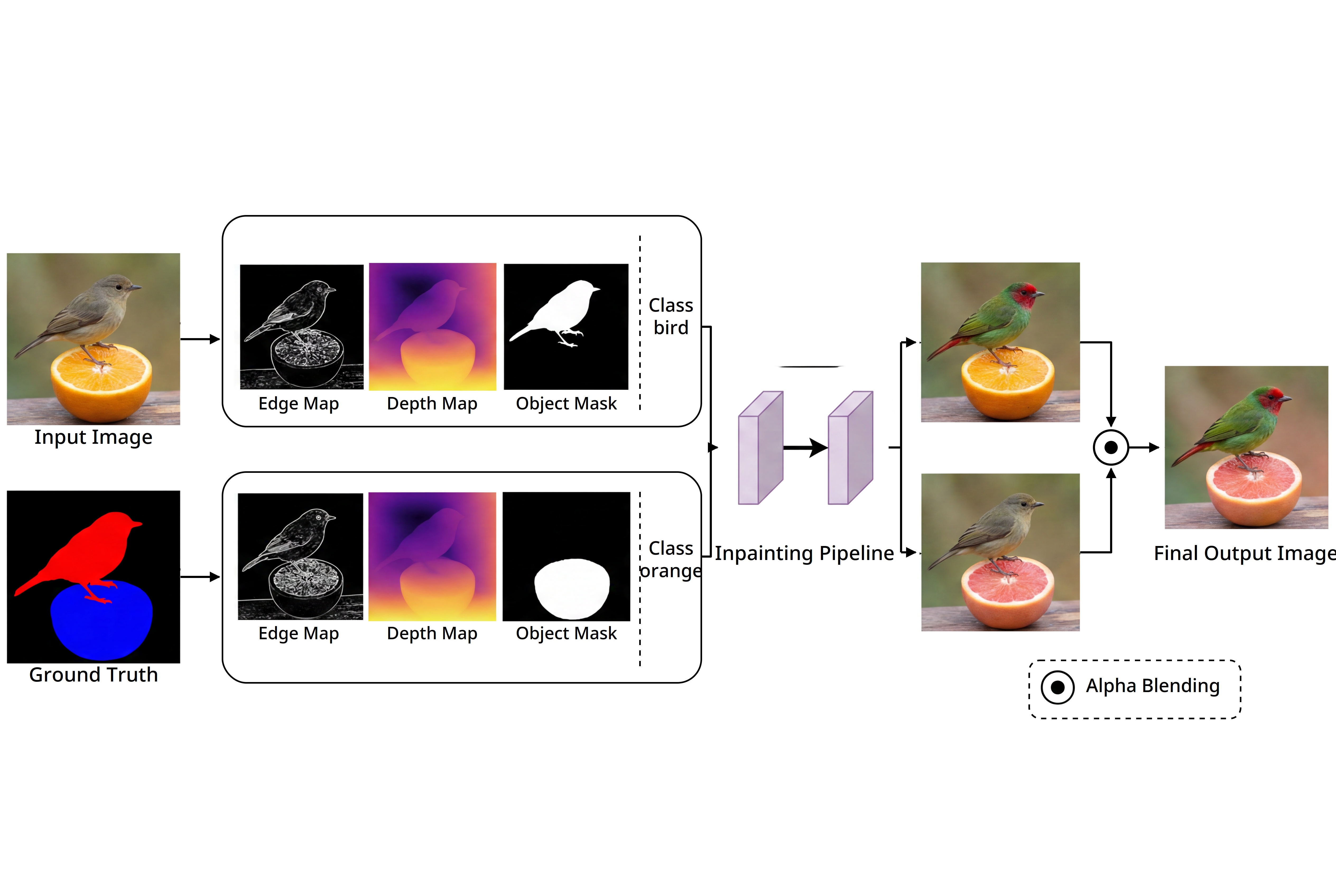}
    & \appimgcell{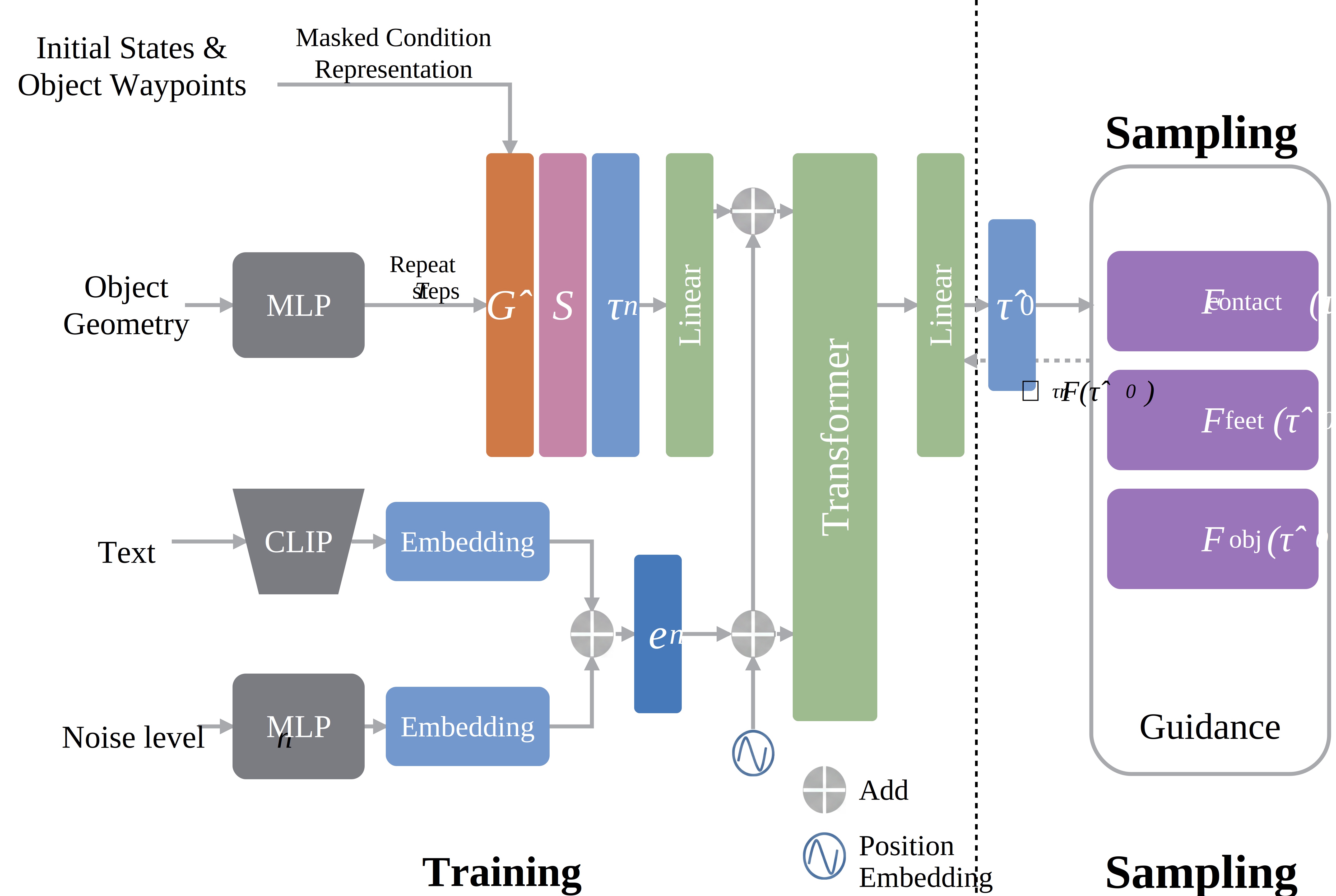}
    \\
    \hline

    \methodlabel{Ours}
    & \appimgcell{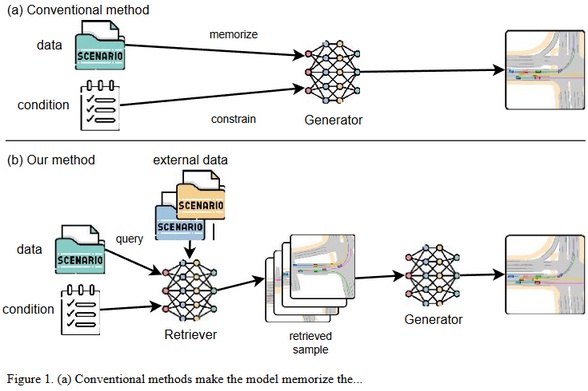}
    & \appimgcell{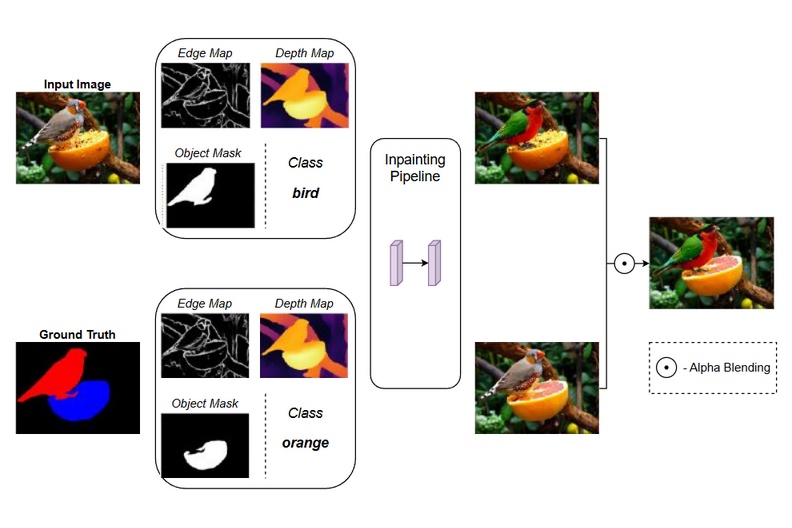}
    & \appimgcell{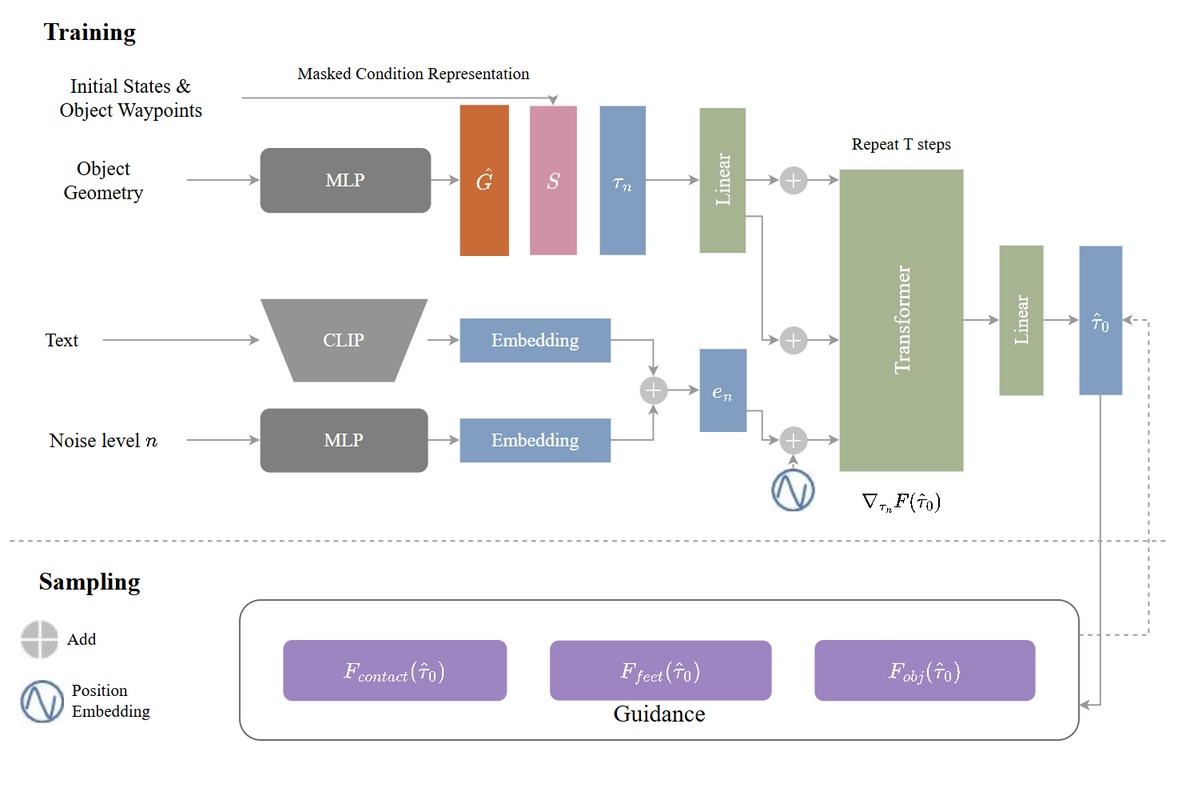}
    \\
    \hline
    \end{tabular}
    }

  \caption{
        \textbf{Additional qualitative comparison under the 3:2 target aspect ratio.}
        In the left example, Nano Banana Pro and PaperBanana introduce many hallucinated elements, while only our method performs clear relayout.
        In the middle example, Nano Banana Pro and PaperBanana also hallucinate structures, and AutoFigure-Edit and GPT Image 2.0 leave large unused space.
        In the right example, Nano Banana Pro, GPT Image 2.0, and AutoFigure-Edit mostly stretch the original image without meaningful relayout.
    }
    \label{fig:qualitative_comparison_extended3}
    \vspace{-3mm}
\end{figure*}

\begin{figure*}[t]
    \centering
    \small
    \setlength{\tabcolsep}{2pt}
    \renewcommand{\arraystretch}{1.05}

    \newcommand{\AppRowHeight}{0.11\textheight}    %
    \newcommand{\AppImageHeight}{0.10\textheight}  %
    \newcommand{\AppLabelWidth}{0.75cm}            %
    \newcommand{\AppCaseWidth}{3.2cm}              %

    \newcommand{\appimgcell}[1]{%
    \begin{minipage}[c][\AppRowHeight][c]{\linewidth}
    \centering
    \setlength{\fboxsep}{0pt}%
    \setlength{\fboxrule}{0.5pt}%
    \fcolorbox{black}{white}{%
    \includegraphics[
        width=0.97\linewidth,
        height=\AppImageHeight,
        keepaspectratio
    ]{#1}%
    }%
    \end{minipage}
    }

    \newcommand{\methodlabel}[1]{%
    \rotatebox[origin=c]{90}{\textbf{#1}}%
    }

    \resizebox{\textwidth}{!}{%
    \begin{tabular}{
        >{\centering\arraybackslash}m{\AppLabelWidth}
        |>{\centering\arraybackslash}m{\AppCaseWidth}
        |>{\centering\arraybackslash}m{\AppCaseWidth}
        |>{\centering\arraybackslash}m{\AppCaseWidth}
    }
    \hline

    \methodlabel{Ratio}
    & \textbf{9:16}
    & \textbf{9:16}
    & \textbf{9:16}
    \\
    \hline

    \methodlabel{Original flowchart}
    & \appimgcell{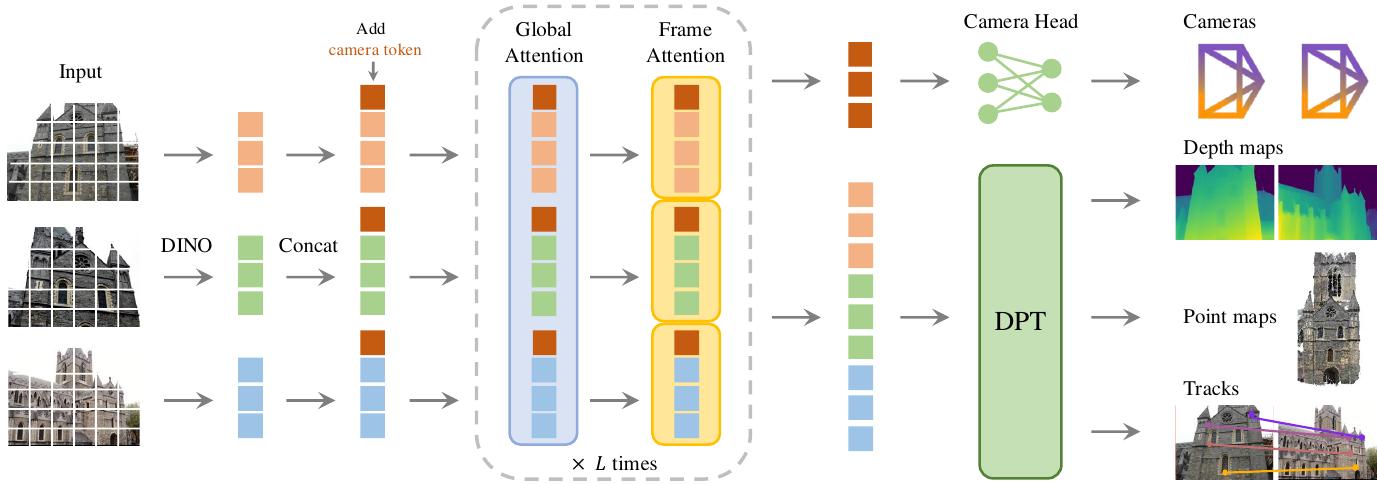}
    & \appimgcell{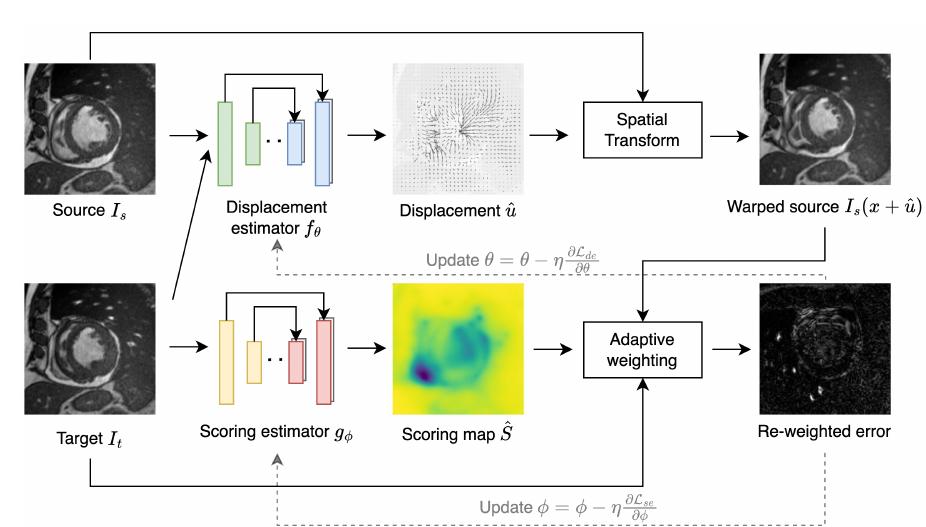}
    & \appimgcell{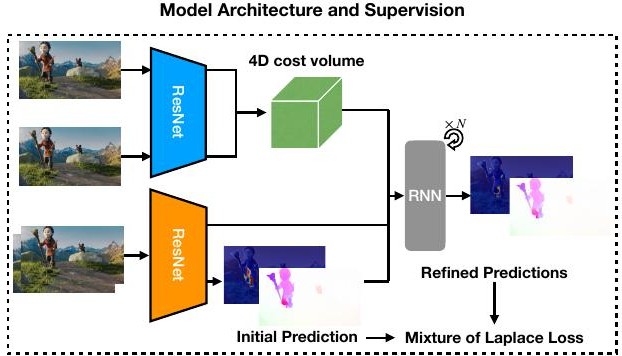}
    \\
    \hline

    \methodlabel{Nano Banana Pro}
    & \appimgcell{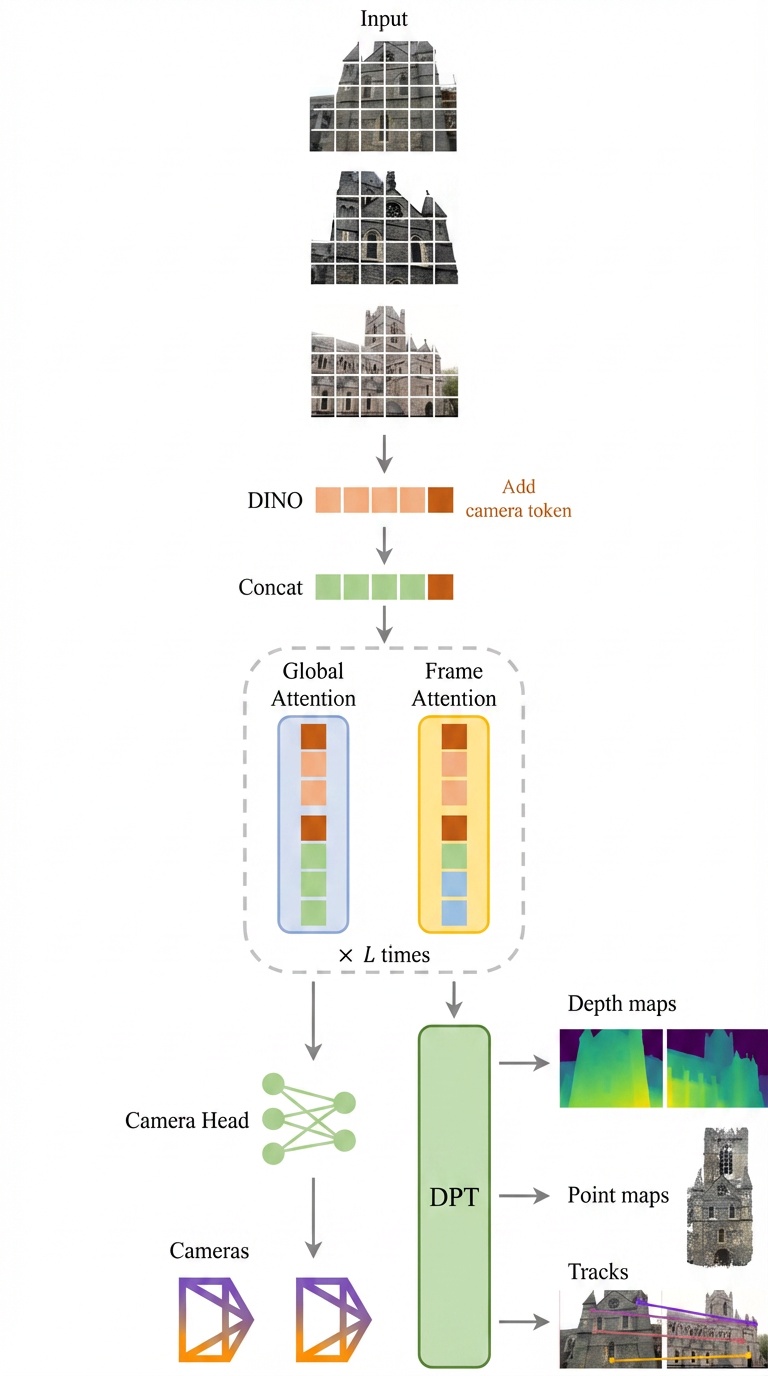}
    & \appimgcell{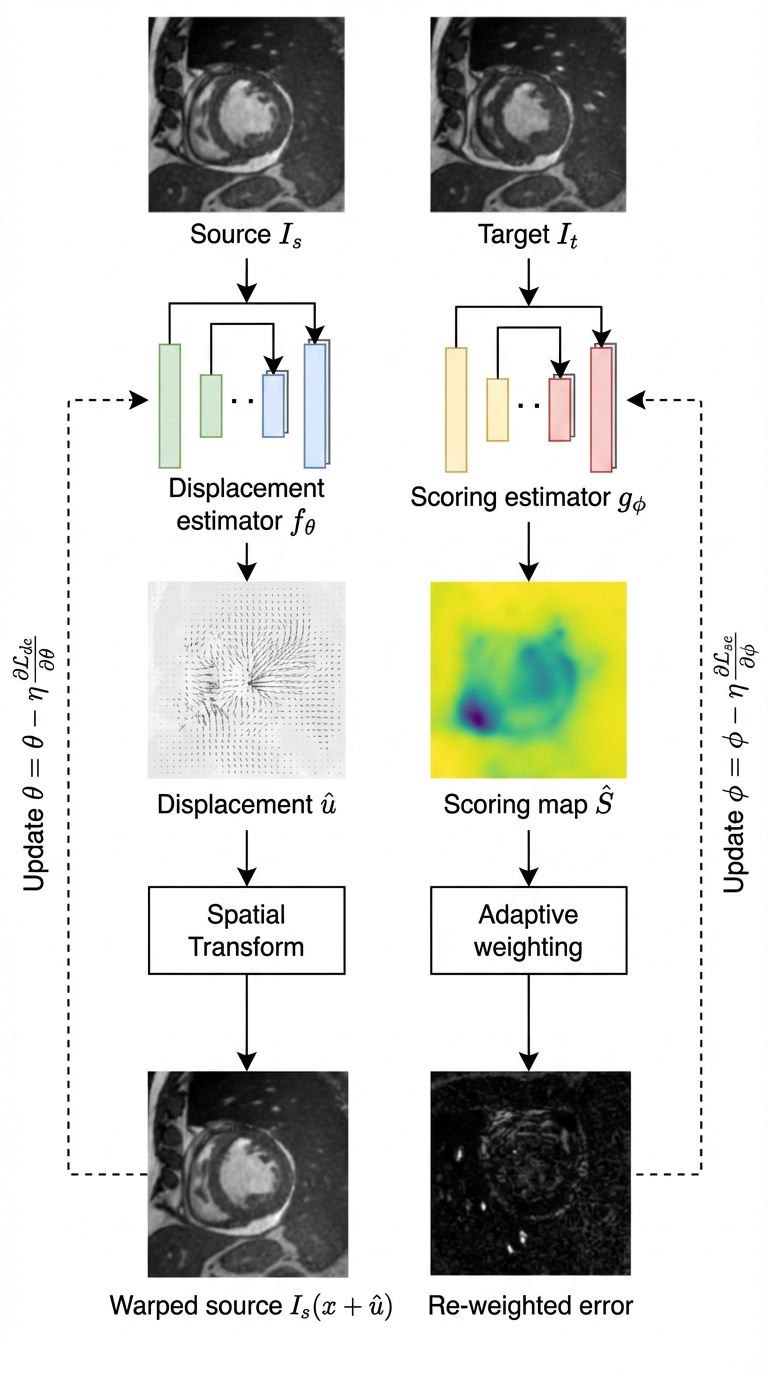}
    & \appimgcell{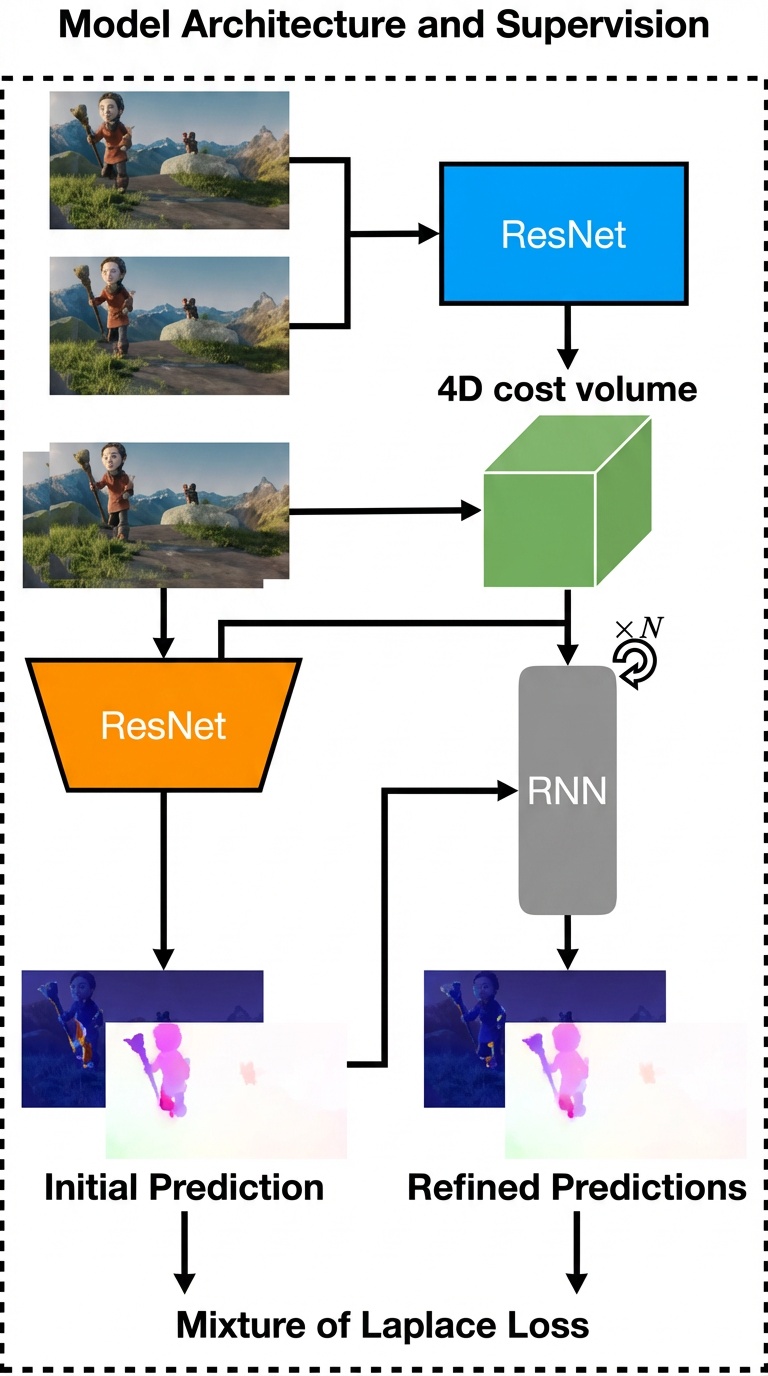}
    \\
    \hline

    \methodlabel{GPT Image 2.0}
    & \appimgcell{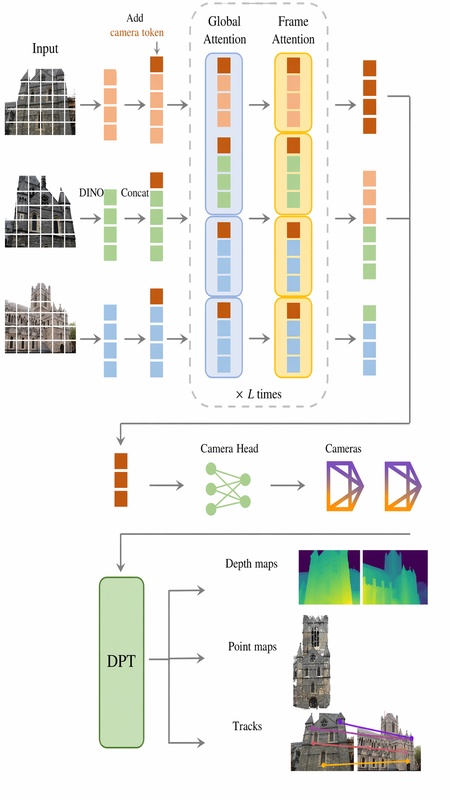}
    & \appimgcell{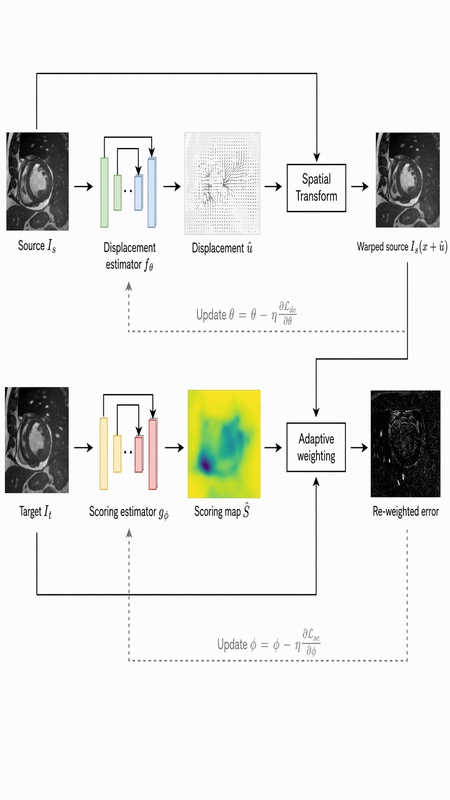}
    & \appimgcell{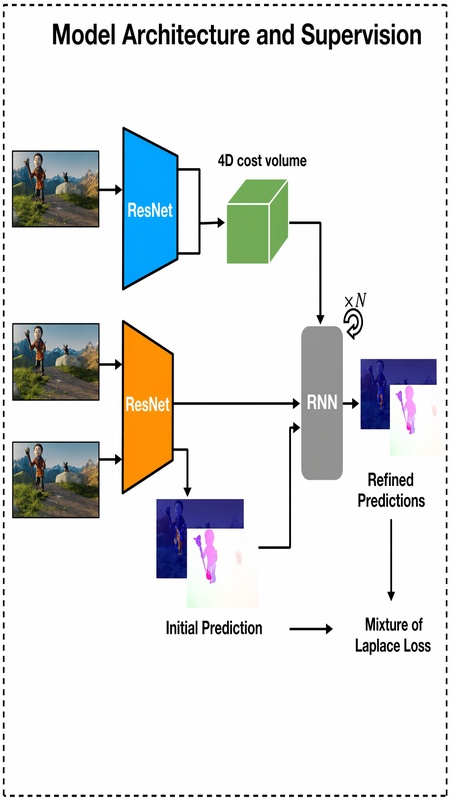}
    \\
    \hline

    \methodlabel{PaperBanana}
    & \appimgcell{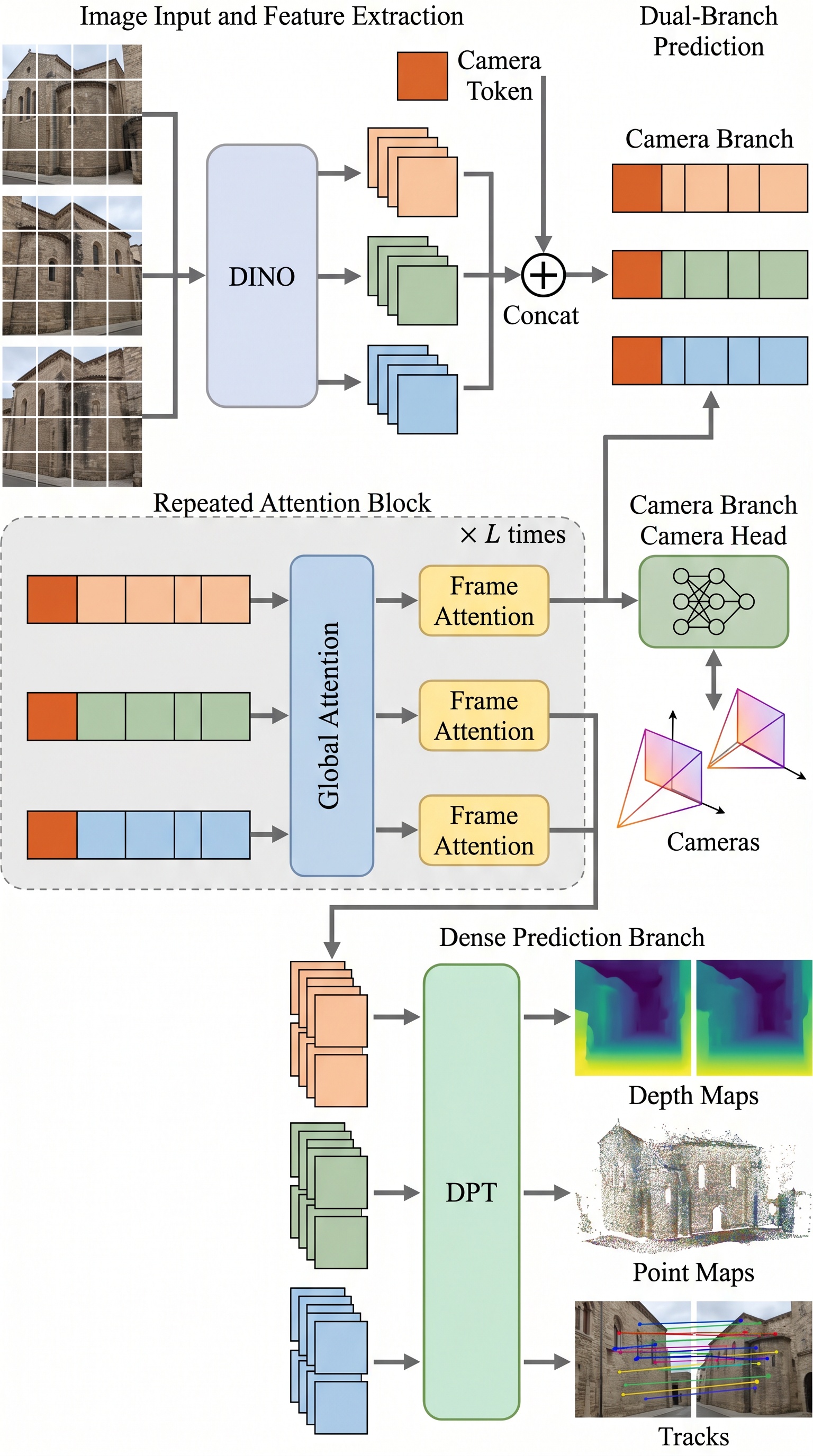}
    & \appimgcell{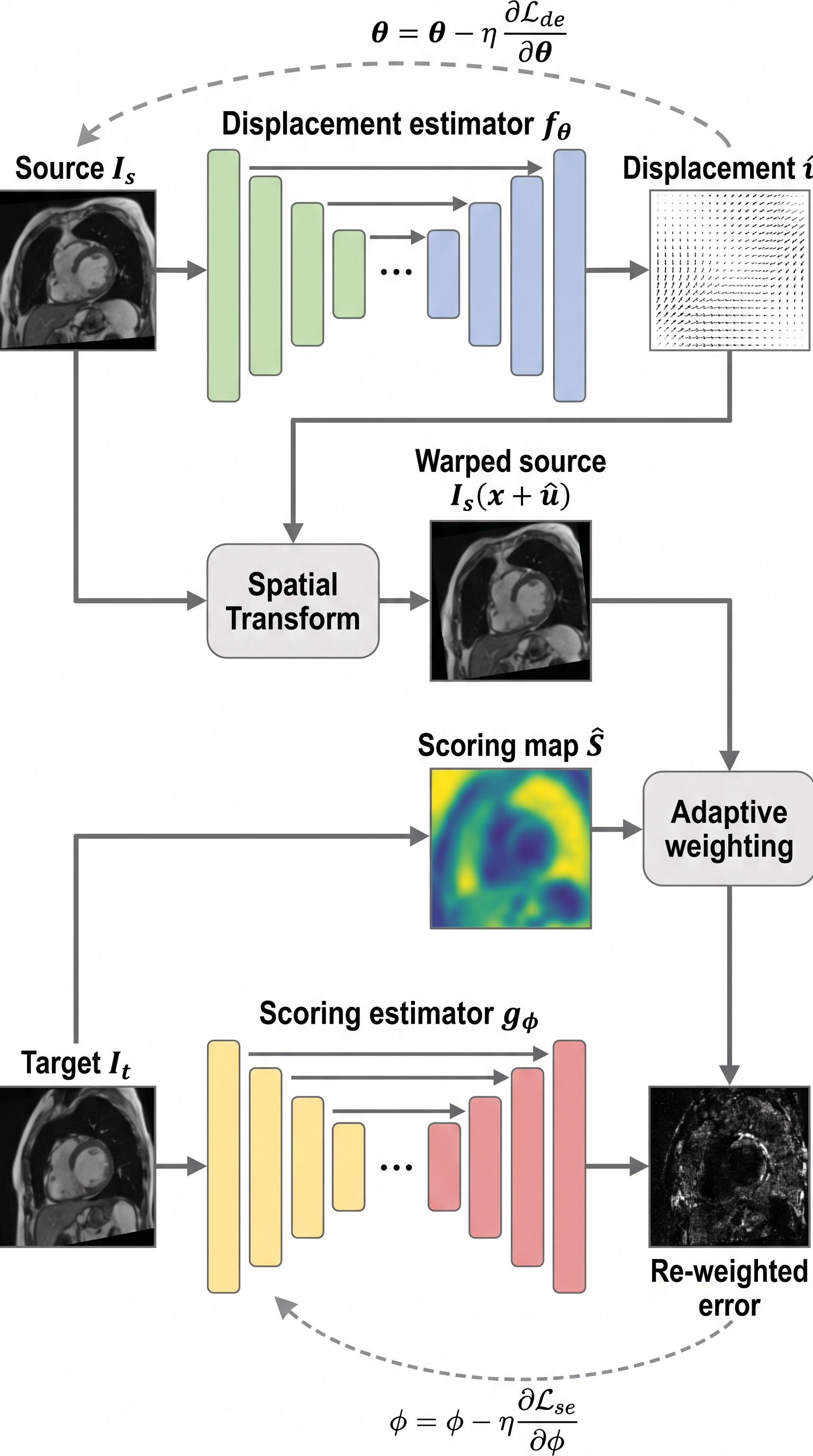}
    & \appimgcell{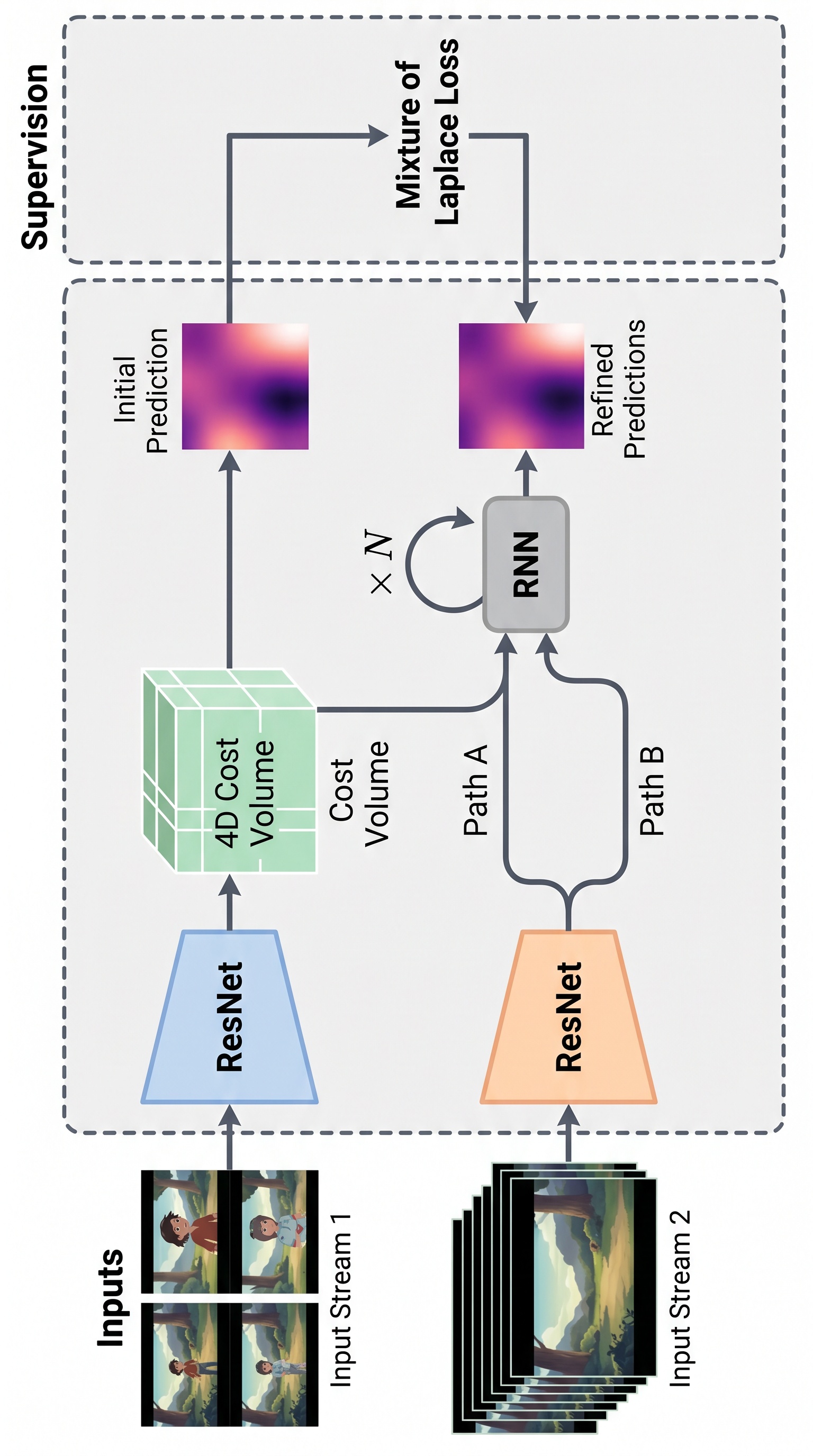}
    \\
    \hline

    \methodlabel{AutoFigure-Edit}
    & \appimgcell{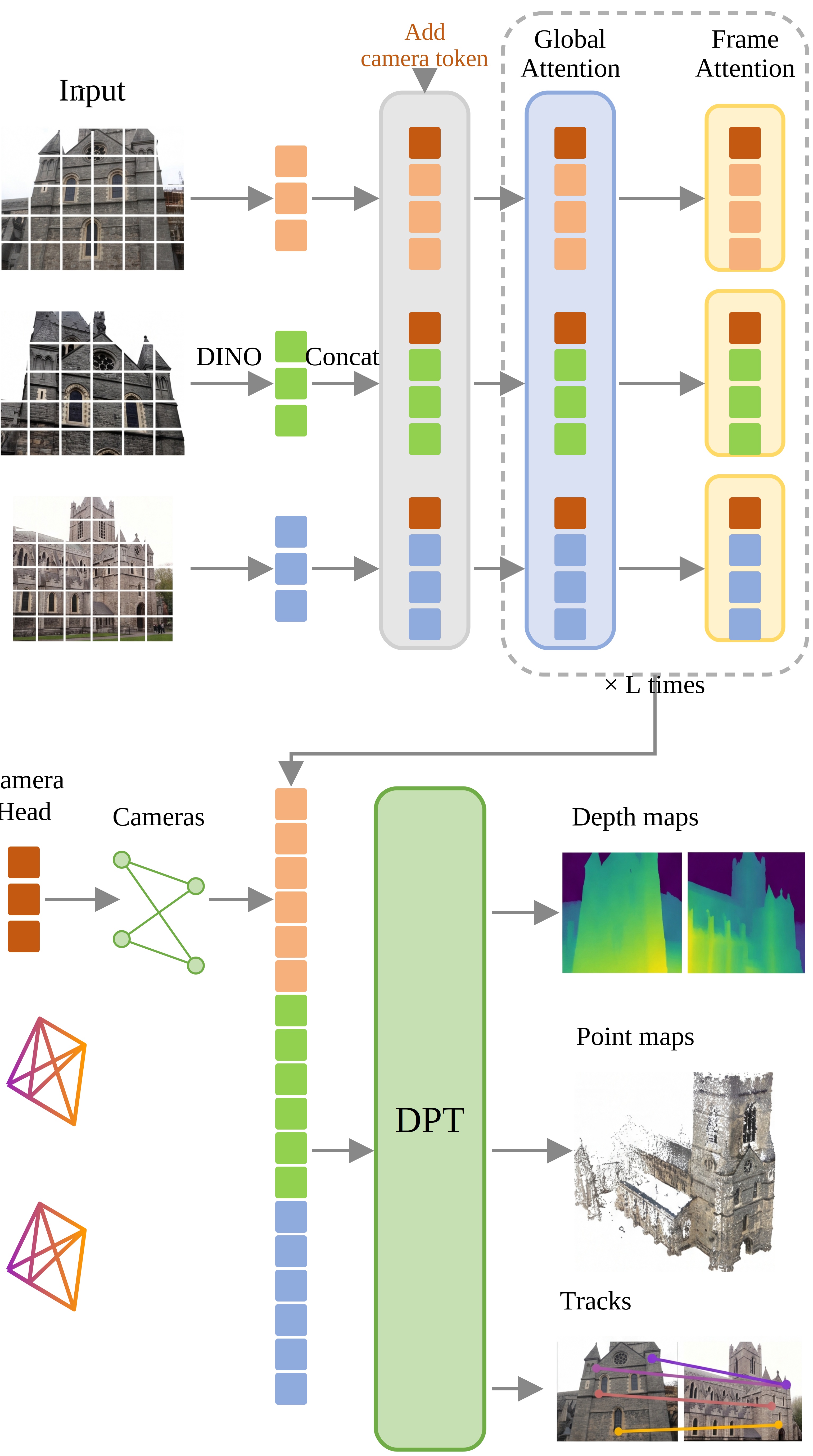}
    & \appimgcell{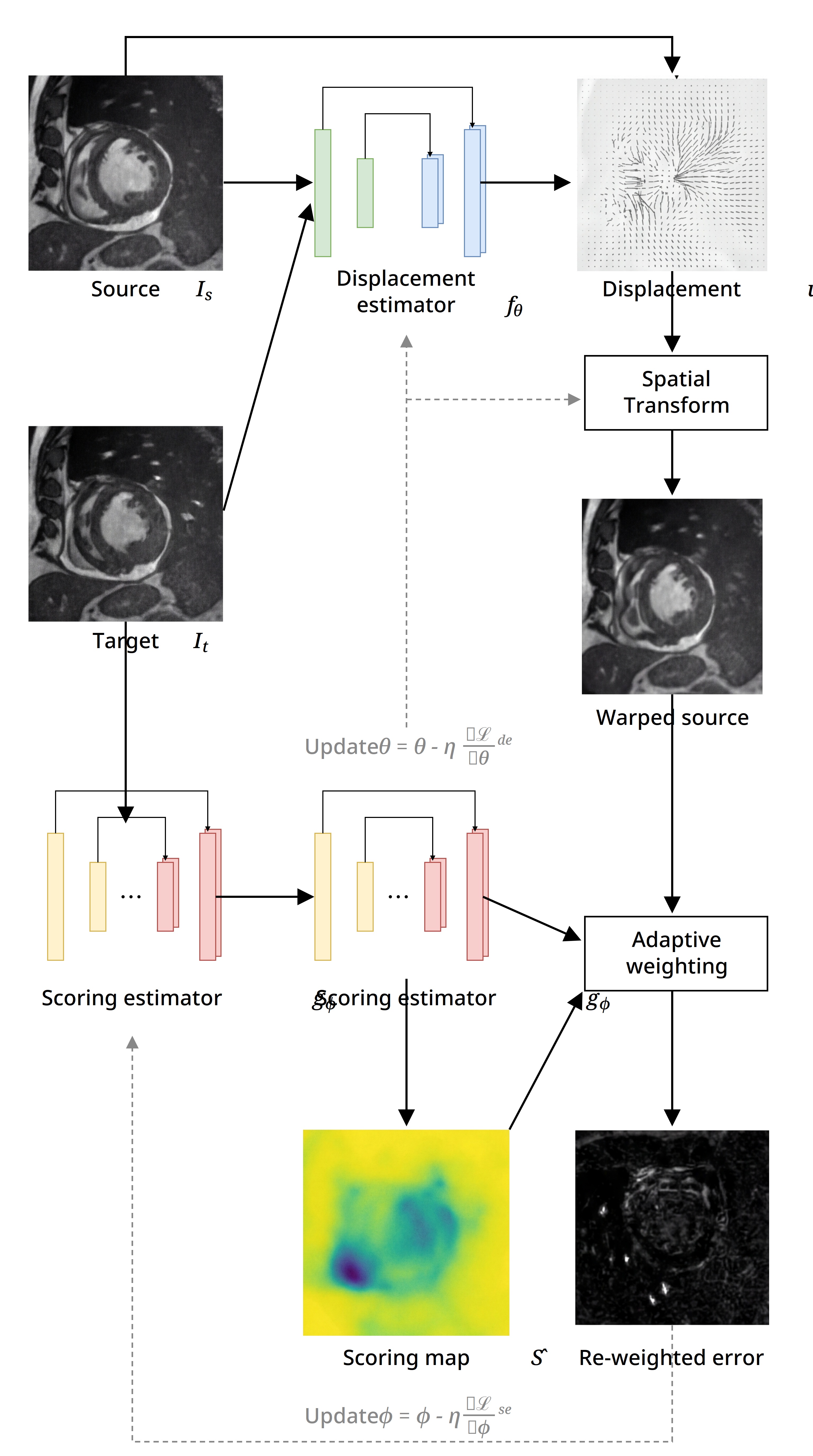}
    & \appimgcell{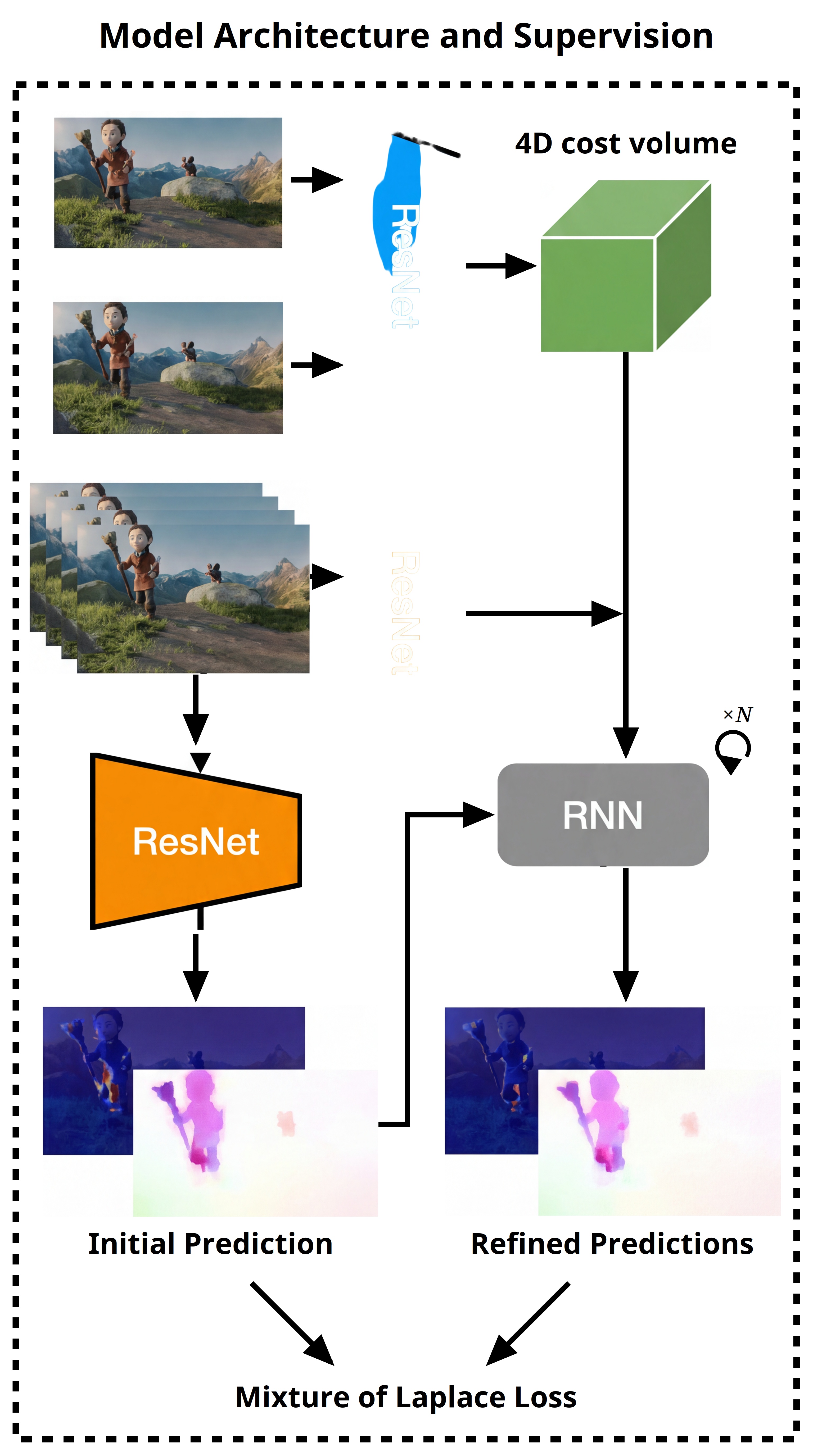}
    \\
    \hline

    \methodlabel{Ours}
    & \appimgcell{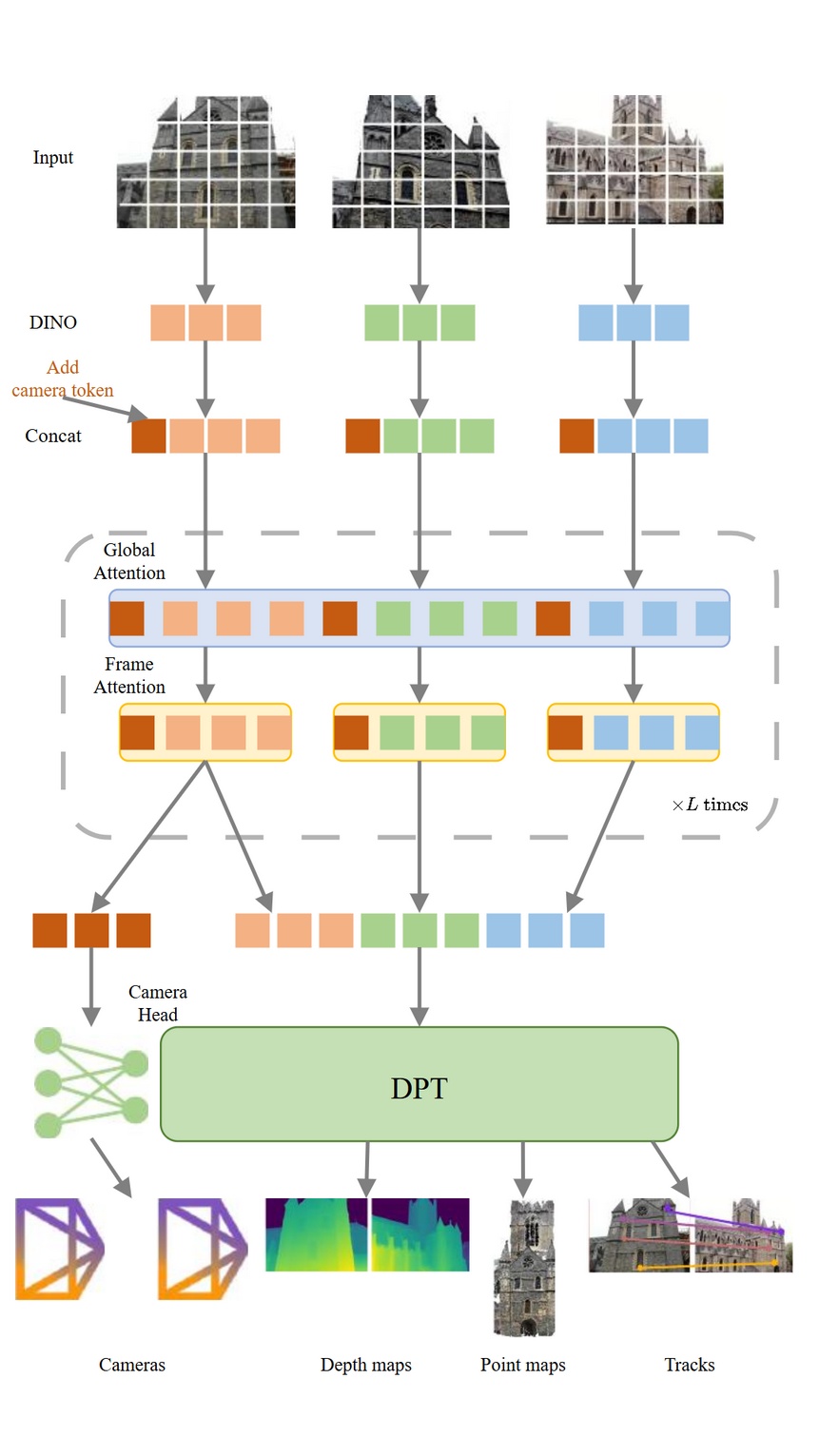}
    & \appimgcell{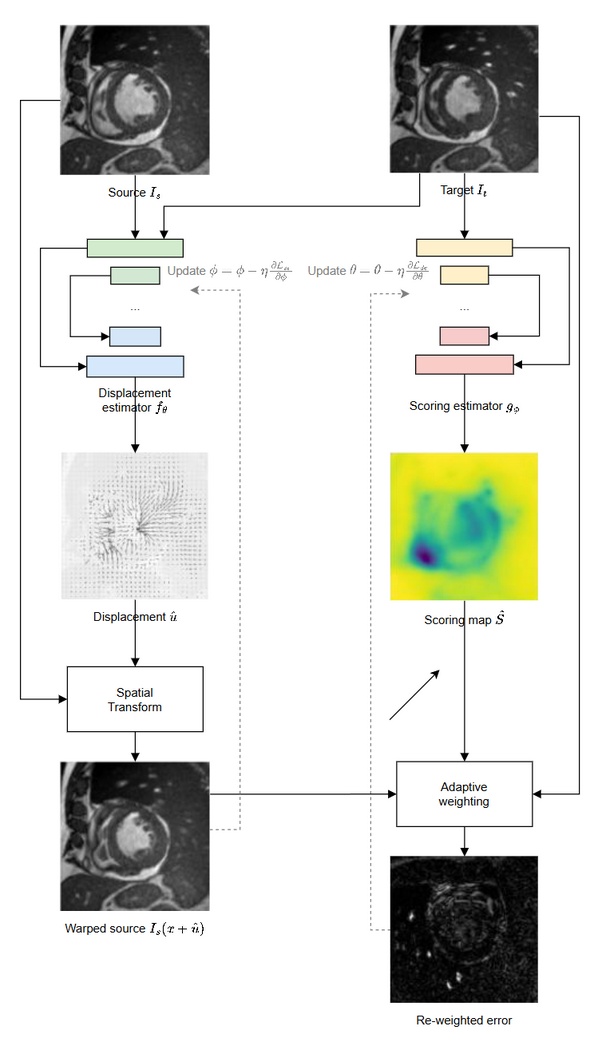}
    & \appimgcell{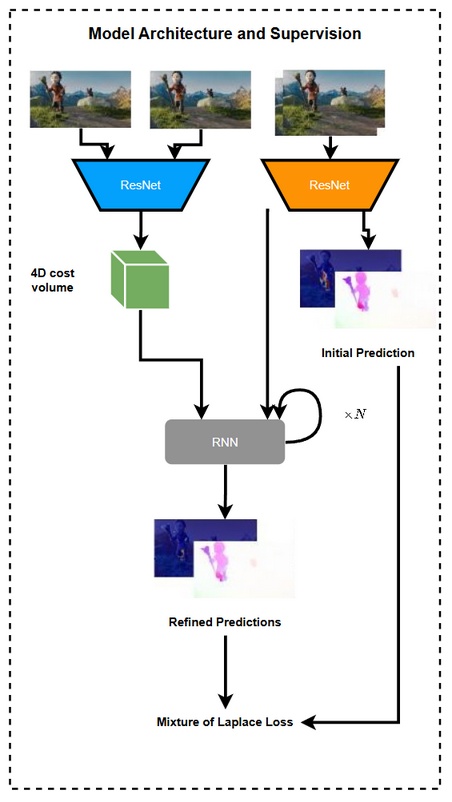}
    \\
    \hline
    \end{tabular}
    }

   \caption{
        \textbf{Additional qualitative comparison under the 9:16 target aspect ratio.}
        In the left example, all four baselines introduce relationship errors, while our method preserves the correct flow structure.
        In the middle example, Nano Banana Pro introduces relationship errors after relayout, GPT Image 2.0 mainly stretches the original image, and AutoFigure-Edit leaves large unused space.
        In the right example, Nano Banana Pro, GPT Image 2.0, and AutoFigure-Edit all produce relationship errors, while PaperBanana rotates a horizontal layout to fit the portrait canvas.
    }
    \label{fig:qualitative_comparison_extended4}
    \vspace{-3mm}
\end{figure*}

\begin{figure*}[t]
    \centering
    \small
    \setlength{\tabcolsep}{2pt}
    \renewcommand{\arraystretch}{1.05}

    \newcommand{\AppRowHeight}{0.11\textheight}    %
    \newcommand{\AppImageHeight}{0.10\textheight}  %
    \newcommand{\AppLabelWidth}{0.75cm}            %
    \newcommand{\AppCaseWidth}{3.2cm}              %

    \newcommand{\appimgcell}[1]{%
    \begin{minipage}[c][\AppRowHeight][c]{\linewidth}
    \centering
    \setlength{\fboxsep}{0pt}%
    \setlength{\fboxrule}{0.5pt}%
    \fcolorbox{black}{white}{%
    \includegraphics[
        width=0.97\linewidth,
        height=\AppImageHeight,
        keepaspectratio
    ]{#1}%
    }%
    \end{minipage}
    }

    \newcommand{\methodlabel}[1]{%
    \rotatebox[origin=c]{90}{\textbf{#1}}%
    }

    \resizebox{\textwidth}{!}{%
    \begin{tabular}{
        >{\centering\arraybackslash}m{\AppLabelWidth}
        |>{\centering\arraybackslash}m{\AppCaseWidth}
        |>{\centering\arraybackslash}m{\AppCaseWidth}
        |>{\centering\arraybackslash}m{\AppCaseWidth}
    }
    \hline

    \methodlabel{Ratio}
    & \textbf{16:9}
    & \textbf{16:9}
    & \textbf{16:9}
    \\
    \hline

    \methodlabel{Original flowchart}
    & \appimgcell{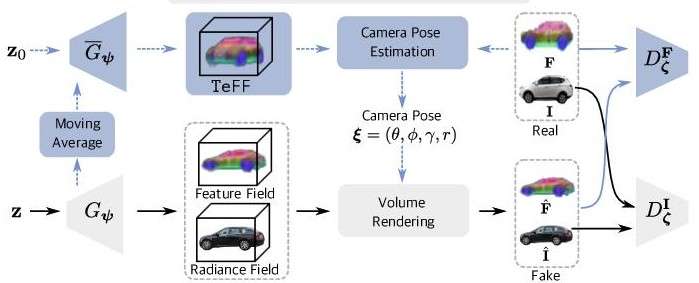}
    & \appimgcell{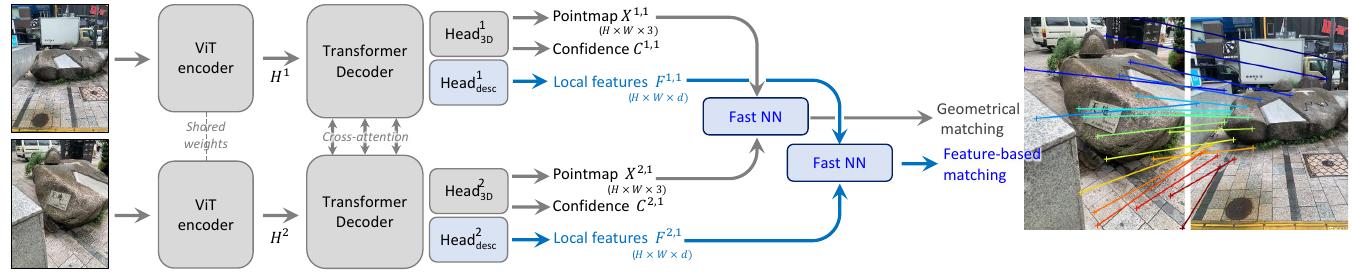}
    & \appimgcell{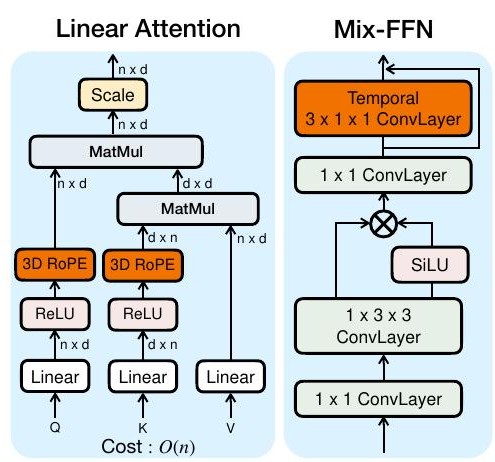}
    \\
    \hline

    \methodlabel{Nano Banana Pro}
    & \appimgcell{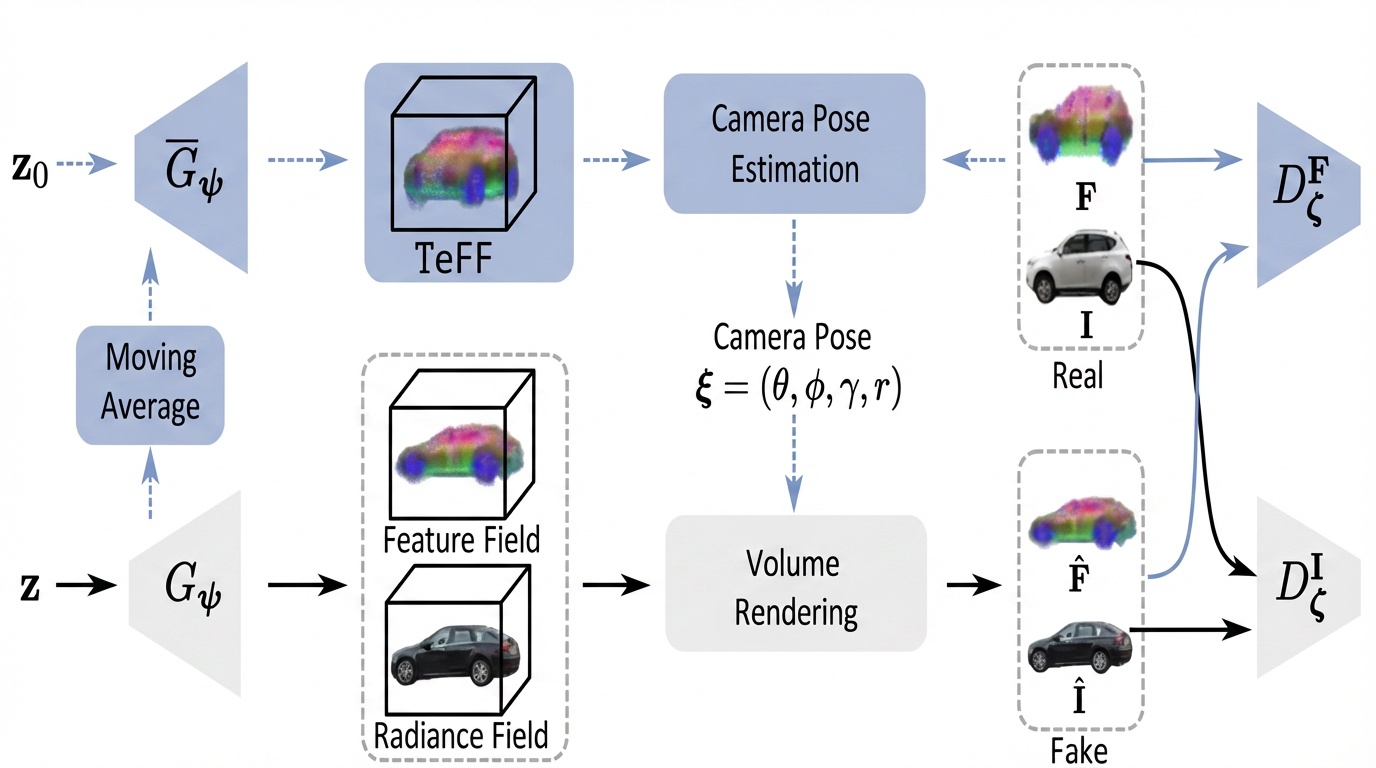}
    & \appimgcell{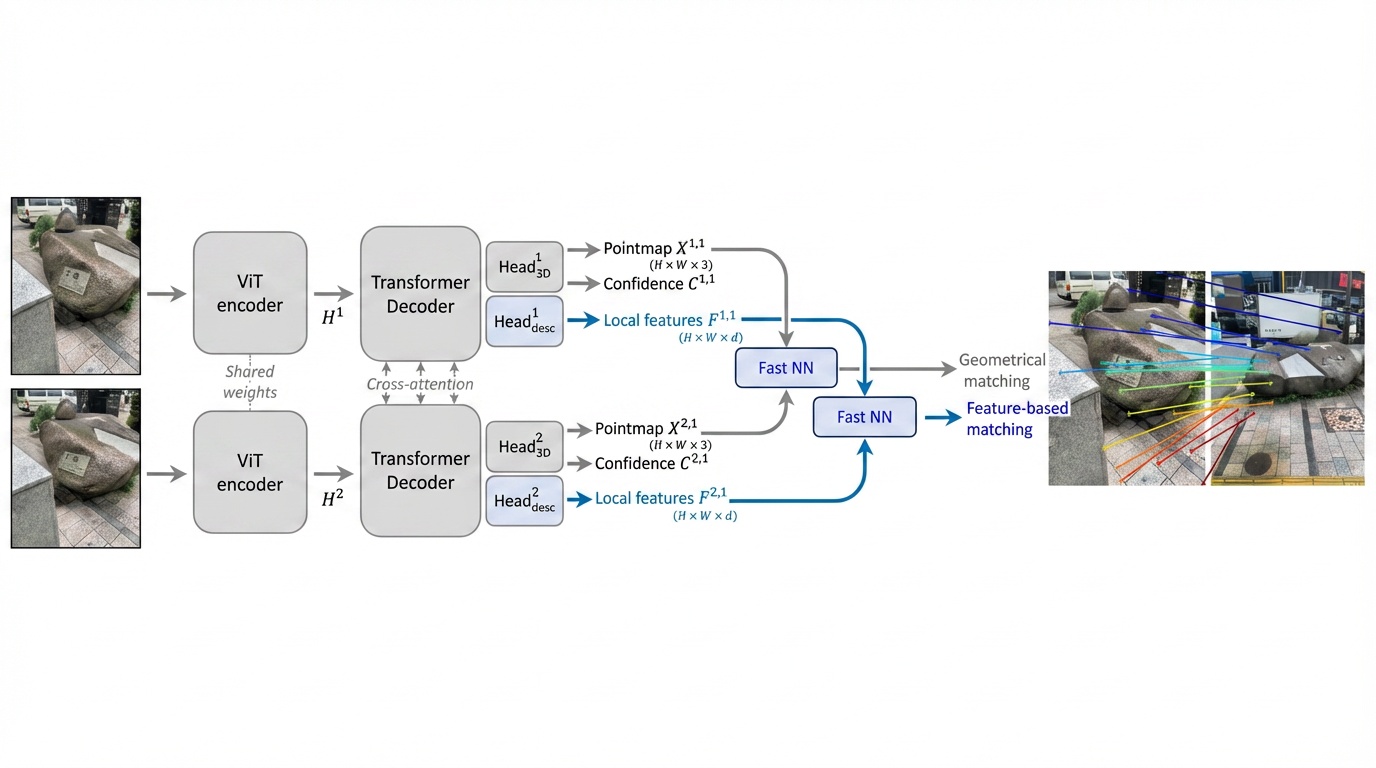}
    & \appimgcell{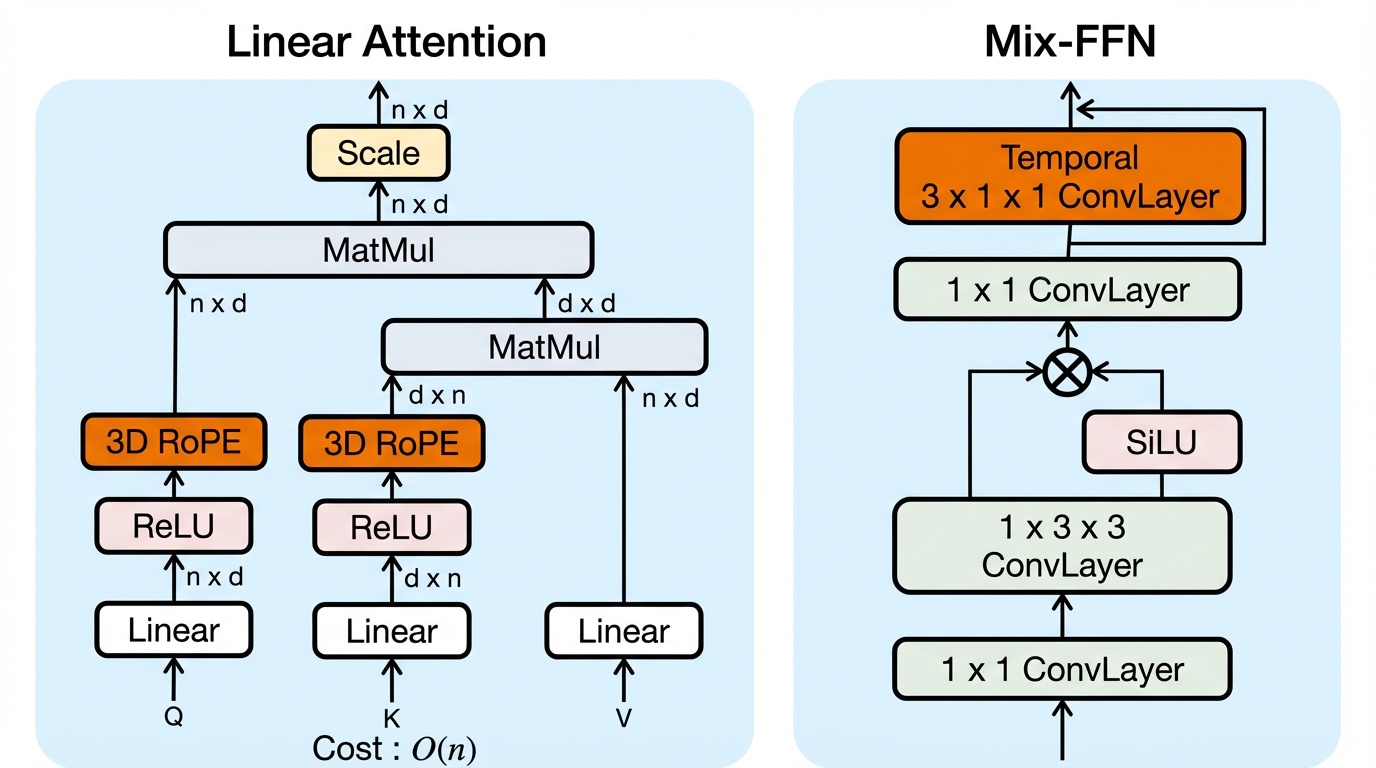}
    \\
    \hline

    \methodlabel{GPT Image 2.0}
    & \appimgcell{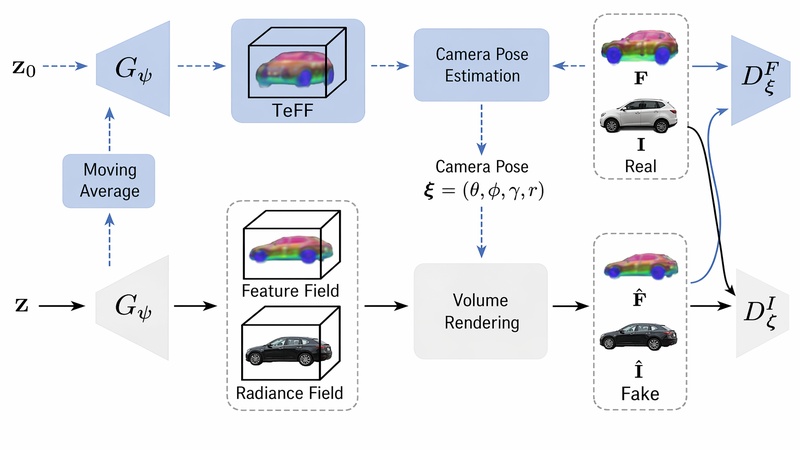}
    & \appimgcell{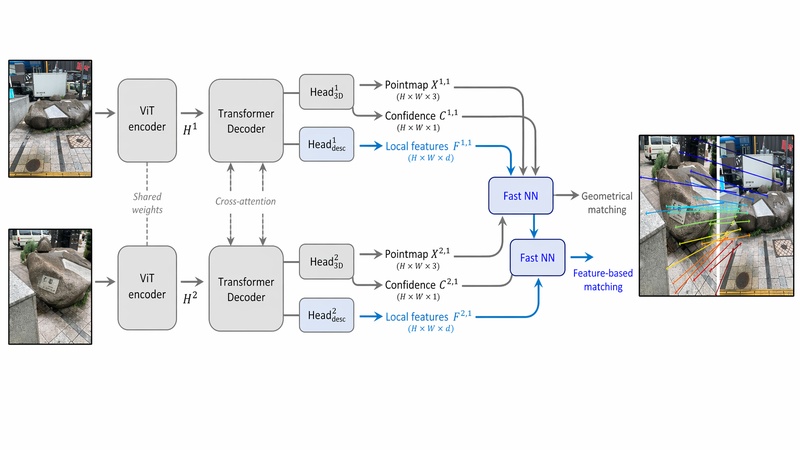}
    & \appimgcell{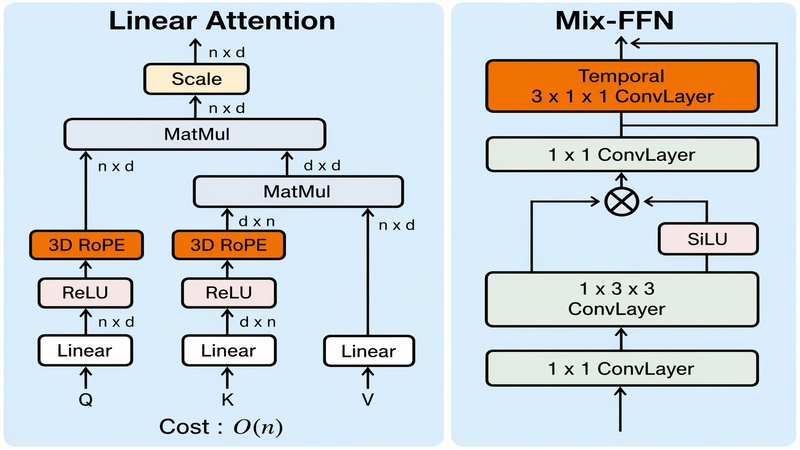}
    \\
    \hline

    \methodlabel{PaperBanana}
    & \appimgcell{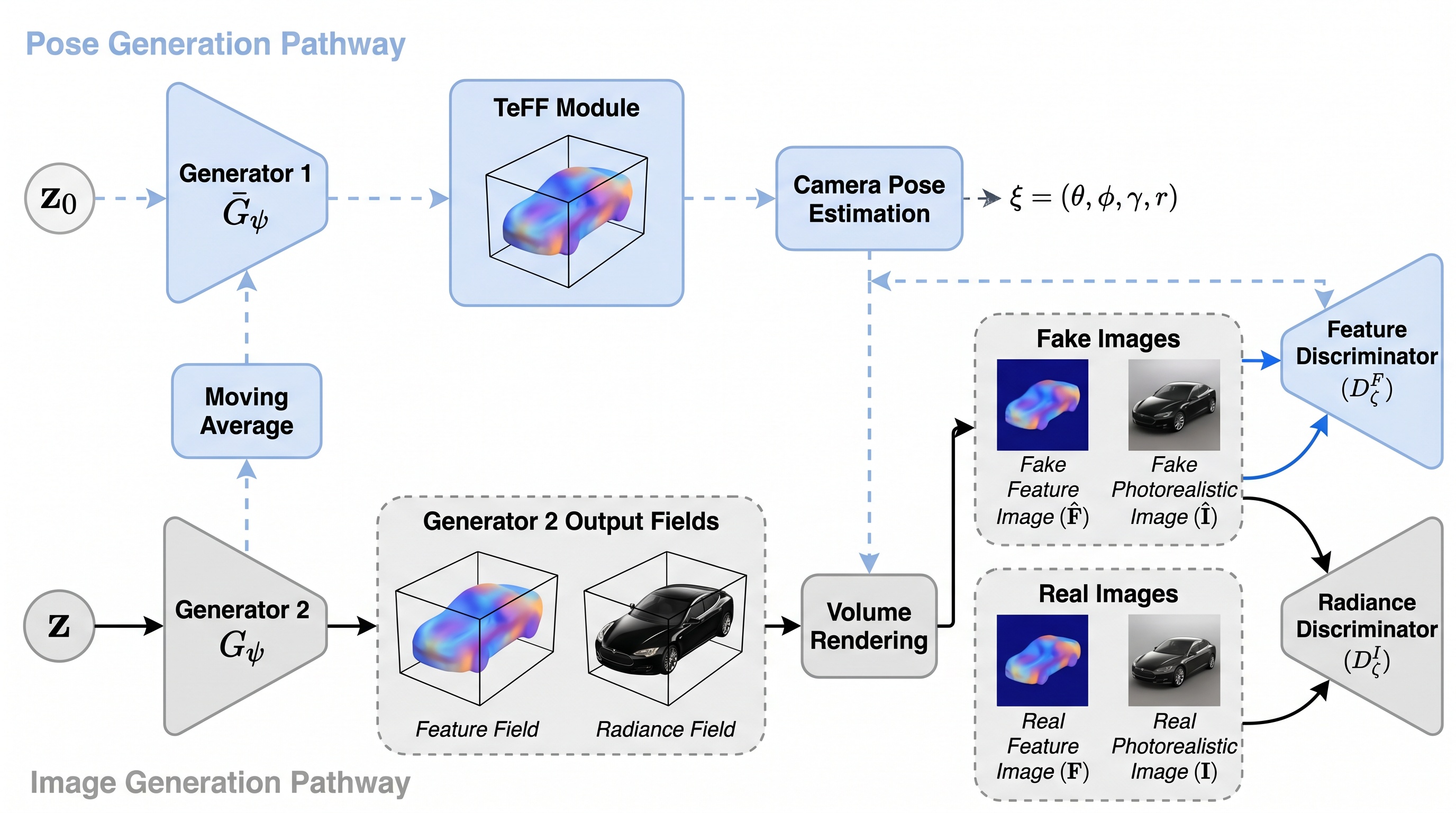}
    & \appimgcell{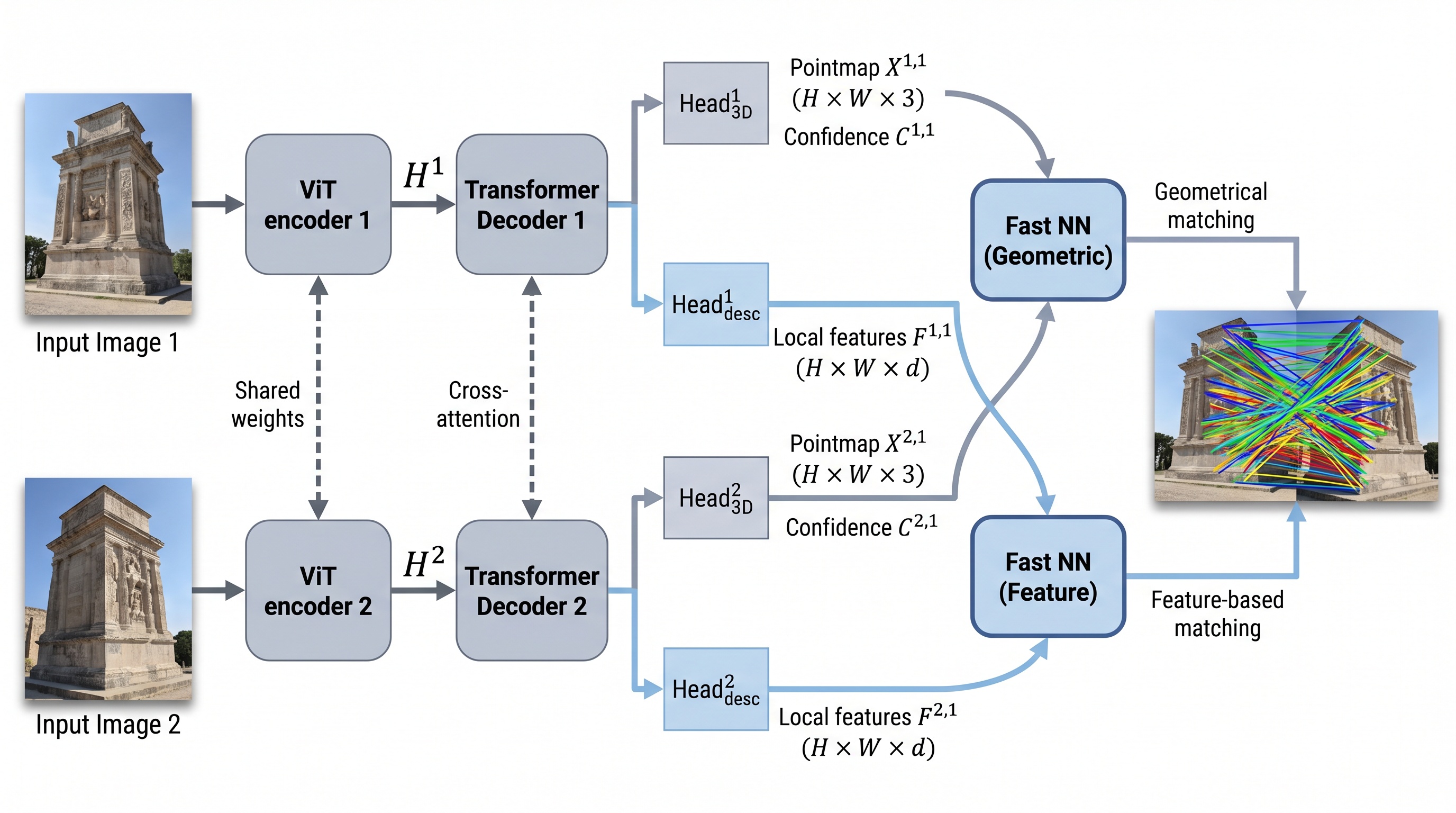}
    & \appimgcell{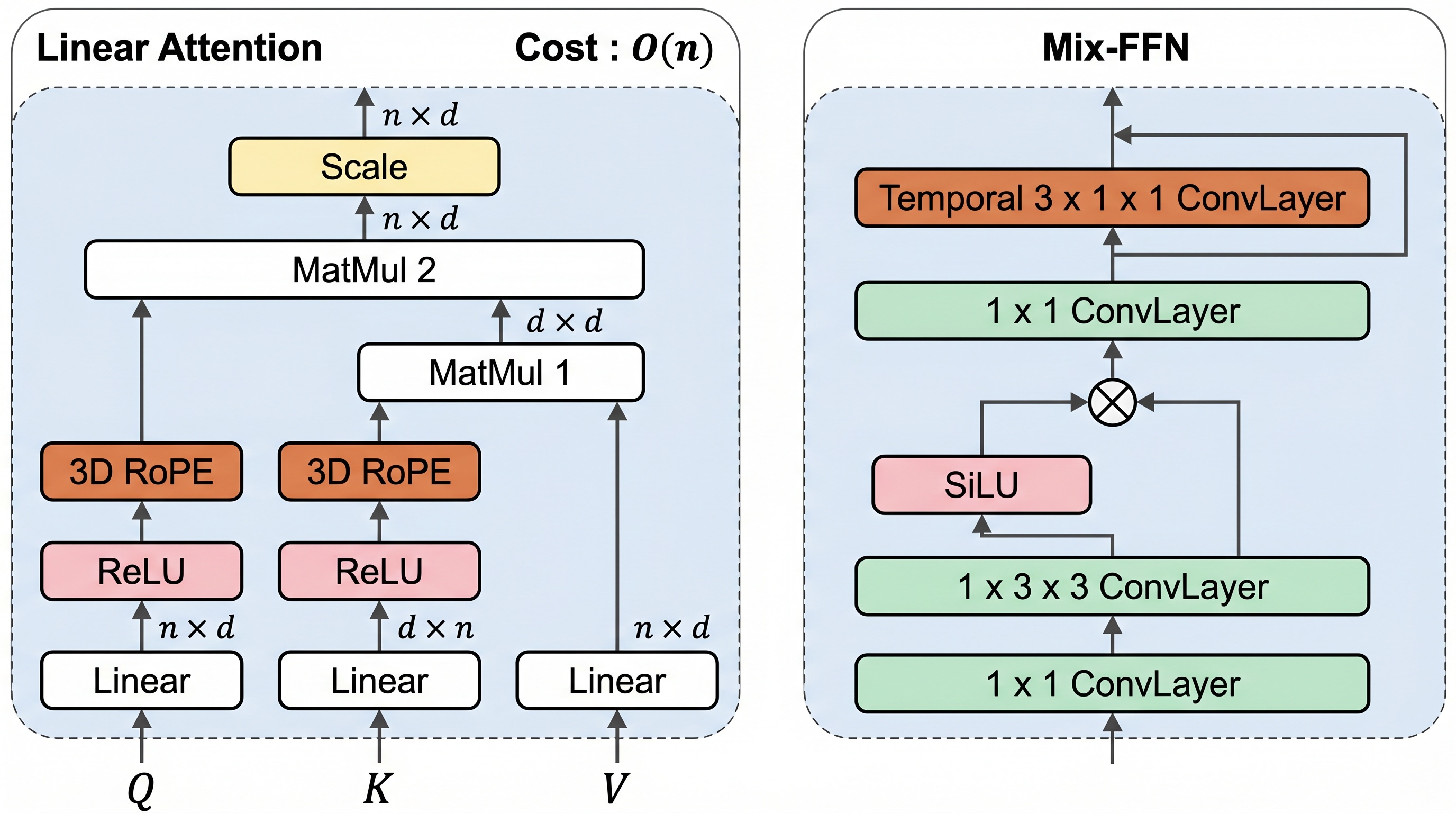}
    \\
    \hline

    \methodlabel{AutoFigure-Edit}
    & \appimgcell{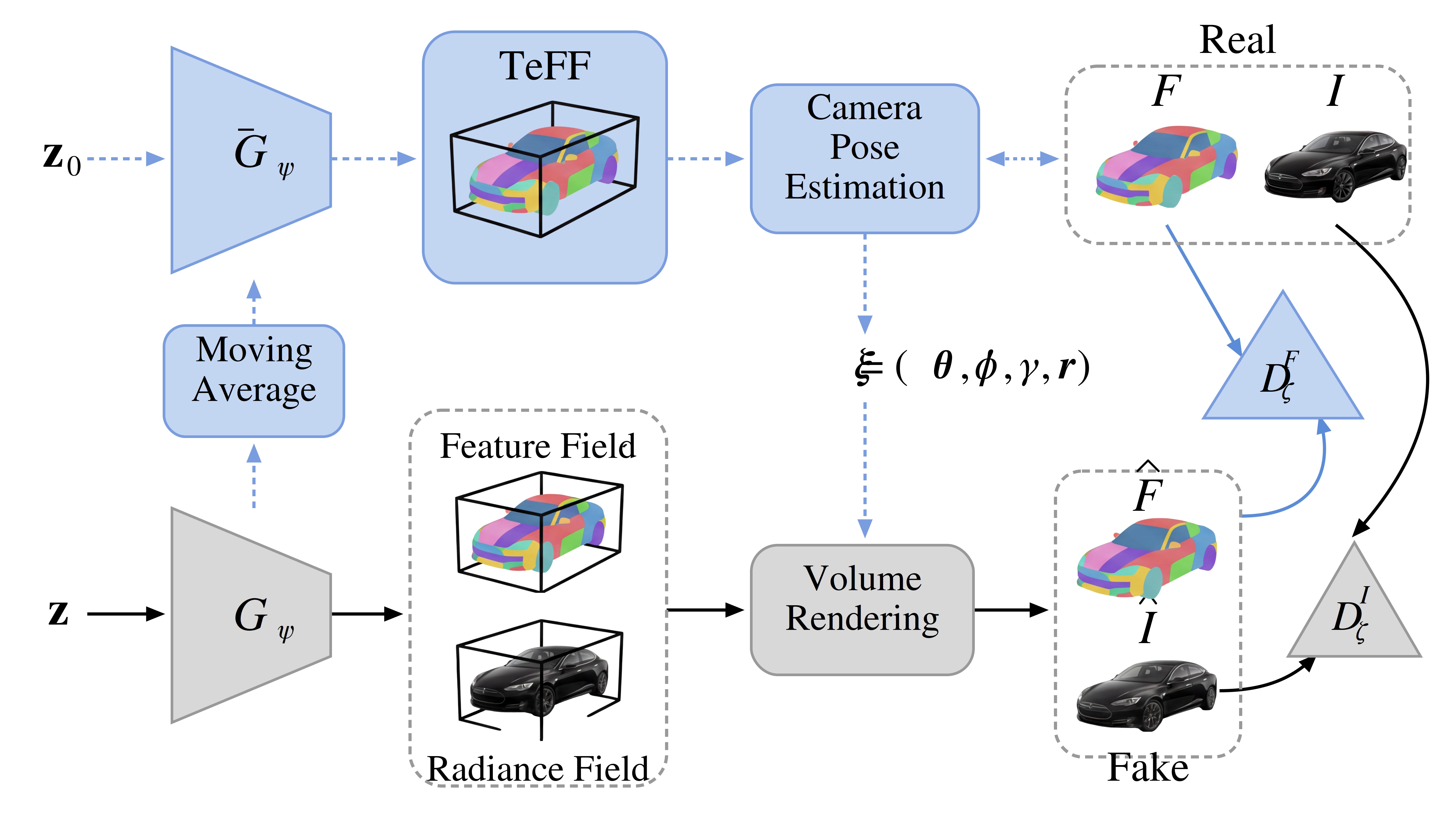}
    & \appimgcell{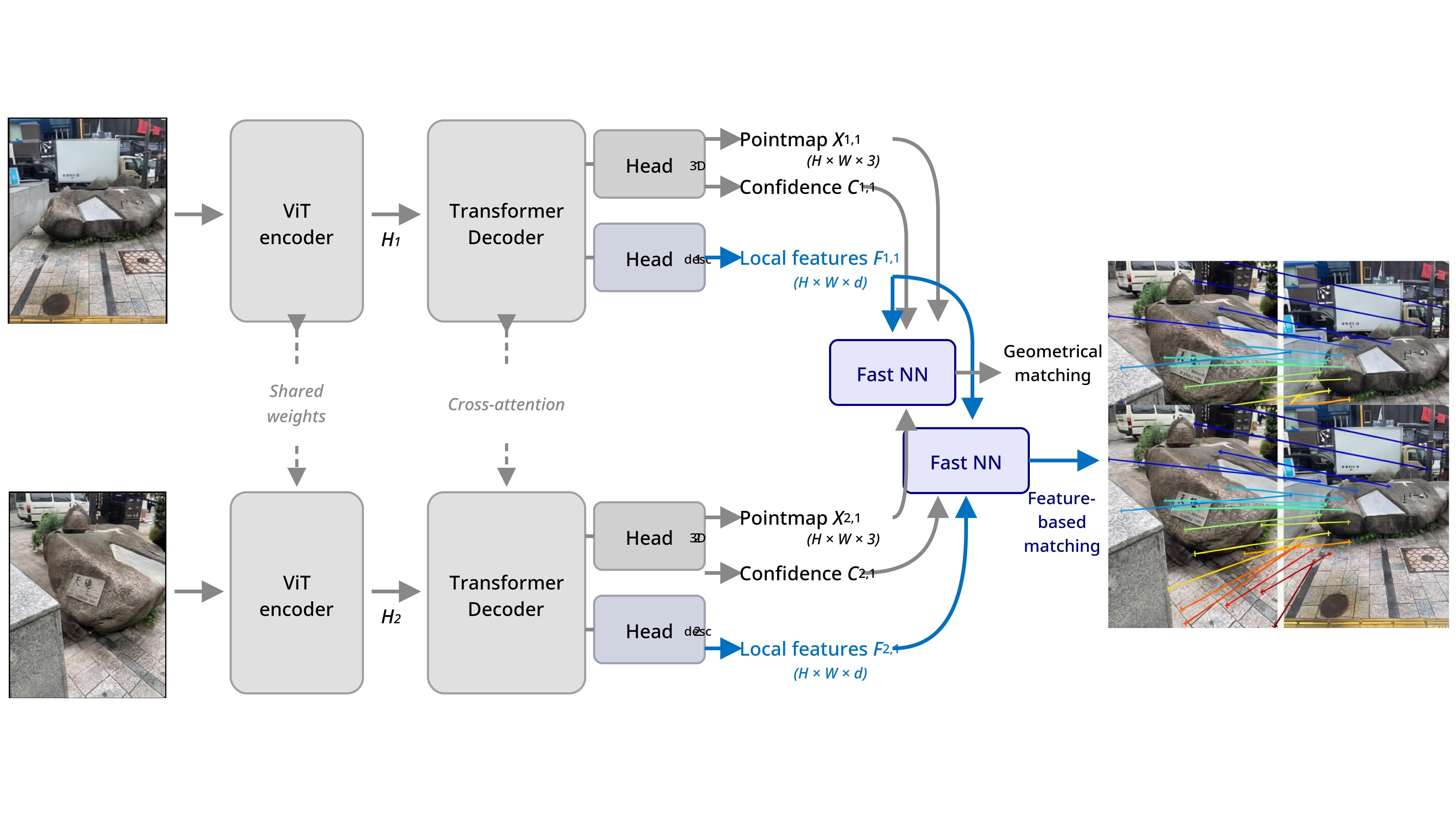}
    & \appimgcell{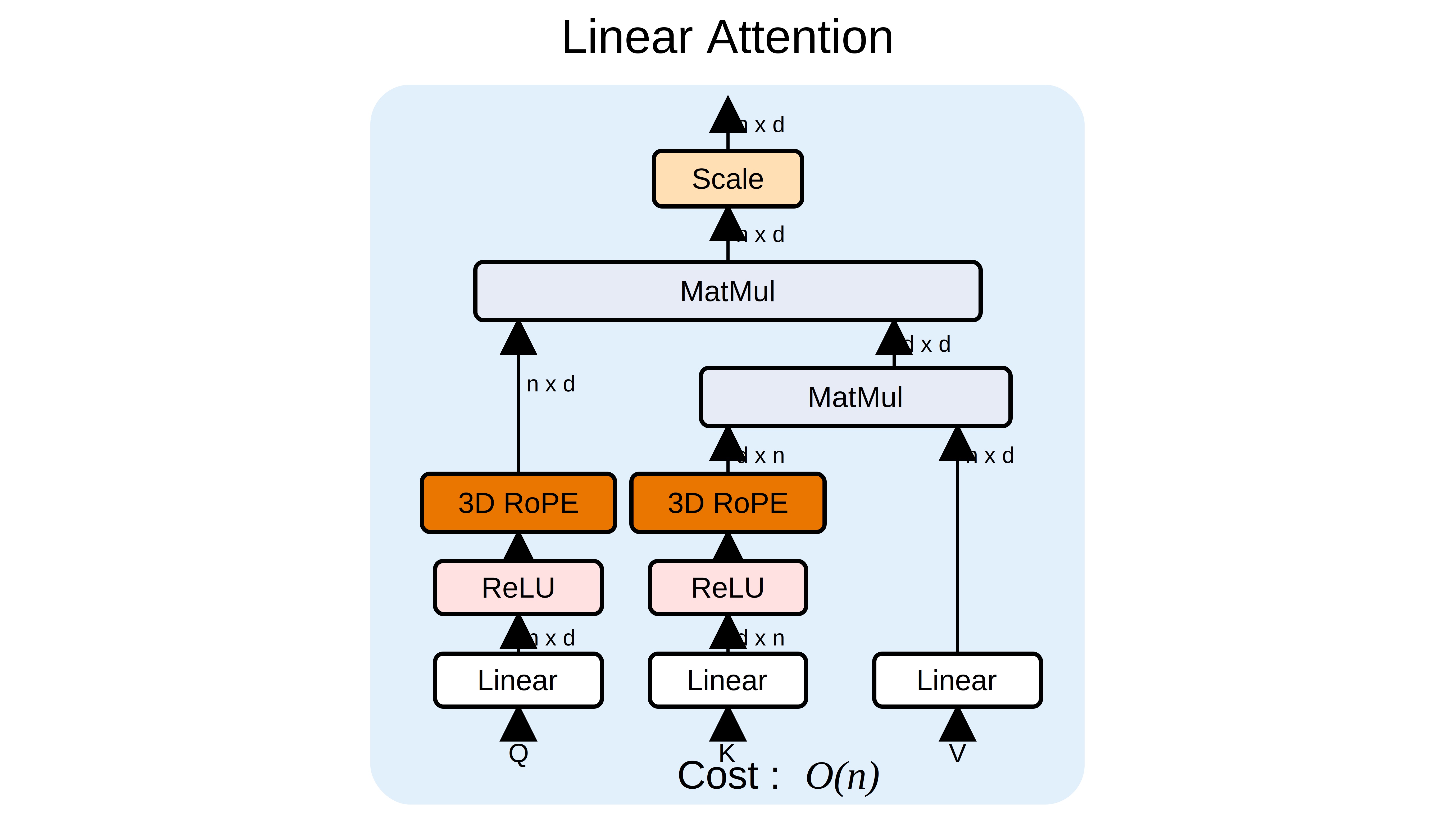}
    \\
    \hline

    \methodlabel{Ours}
    & \appimgcell{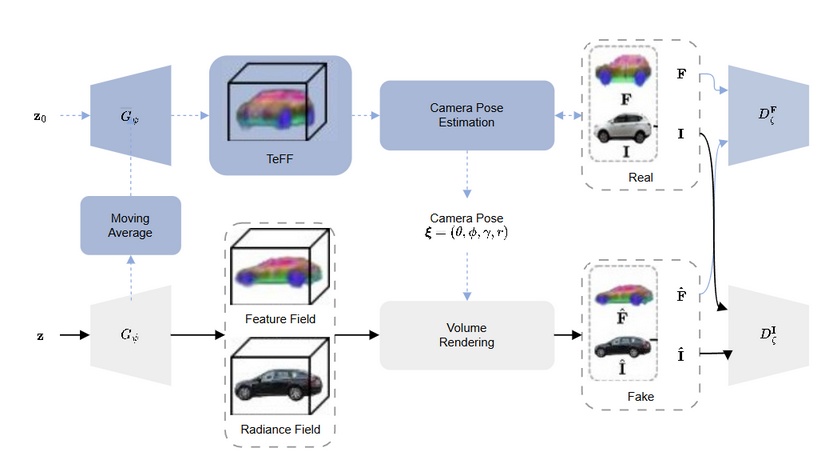}
    & \appimgcell{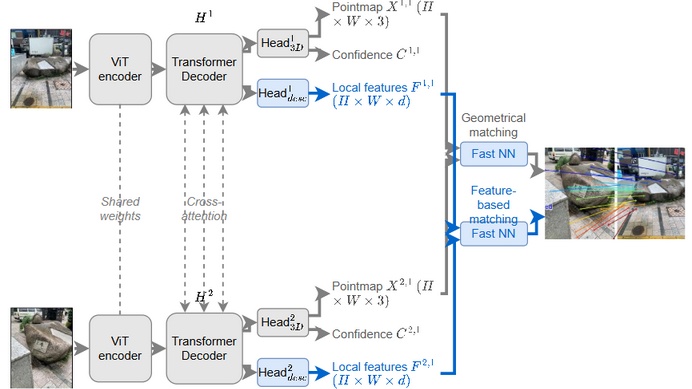}
    & \appimgcell{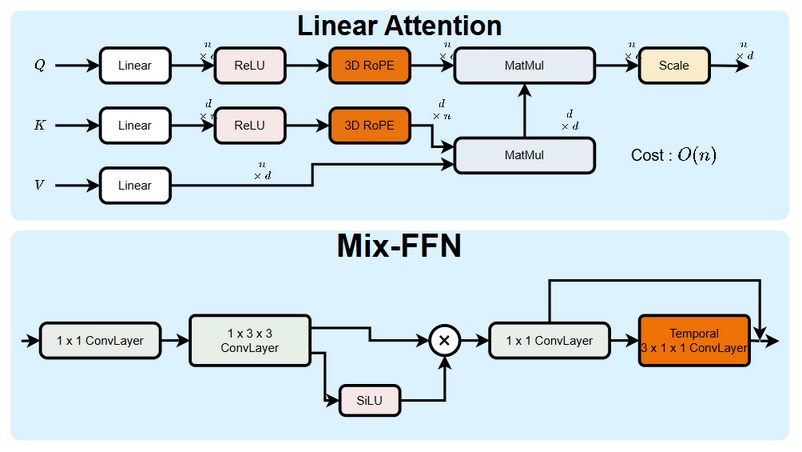}
    \\
    \hline
    \end{tabular}
    }

    \caption{
        \textbf{Additional qualitative comparison under the 16:9 target aspect ratio.}
        In the left example, Nano Banana Pro and GPT Image 2.0 directly stretch the original image, causing the embedded image blocks to deform, while PaperBanana and AutoFigure-Edit introduce hallucinated elements.
        In the middle example, Nano Banana Pro, GPT Image 2.0, and AutoFigure-Edit leave large blank regions, while PaperBanana produces many hallucinated structures.
        In the right example, Nano Banana Pro, GPT Image 2.0, and PaperBanana largely keep the original layout without meaningful relayout, and AutoFigure-Edit leaves large unused space. Only our method performs clear relayout to better adapt the layout to the target canvas.
    }
    \label{fig:qualitative_comparison_extended5}
    \vspace{-3mm}
\end{figure*}

This appendix section extends the main comparison results in Sec.~\ref{sec:comparisons} with additional qualitative results under five target aspect ratios: 1:1, 2:3, 3:2, 9:16, and 16:9. Each aspect ratio is presented in a separate figure, with the extended comparisons shown in Fig.~\ref{fig:qualitative_comparison_extended1}, Fig.~\ref{fig:qualitative_comparison_extended2}, Fig.~\ref{fig:qualitative_comparison_extended3}, Fig.~\ref{fig:qualitative_comparison_extended4}, and Fig.~\ref{fig:qualitative_comparison_extended5}. These examples provide more baseline comparisons and further illustrate common failures such as hallucinated elements, stretched or compressed outputs, relationship errors, and broken text/math rendering. Our method consistently better preserves the flowchart structure while adapting to the target canvas in an editable format.

\clearpage
\newpage
\section{Pipeline Agent Prompts}
\label{app:pipeline-prompts}

This section reproduces the implementation prompts and rule text used by the pipeline agents, corresponding to the Parse (Section~\ref{sec:parse}), Style (Section~\ref{sec:style}), and Layout (Section~\ref{sec:layout}) stages described in the main paper. Minor editorial adjustments were made for readability and presentation consistency.

\subsection{Parse Agent}
\label{app:parse-agent}
\begin{lstlisting}[style=promptappendix,caption={Parse agent main prompt.},label={lst:parse-agent-main}]
**Role:**
You are a tool that converts a flowchart image into mxGraphModel XML readable and editable in draw.io.

**Inputs:**
- Target Diagram: [the provided flowchart image]
- Canvas size: width {width}, height {height}
- previous_XML: {previous_xml}
- previous_suggestions: {feedback}
- bbox_json: {bbox_json}

`bbox_json` is a high-confidence geometric prior. Each object may include: `type` (rect / image / polygon), `x`, `y`, `width`, `height`, and `polyCoords` (for polygon only; normalized coordinates relative to the object's bounding box). Use bbox_json as the primary basis for node shapes and `<mxGeometry>` values. If `previous_suggestions` is non-empty, apply those corrections first before addressing other issues.

**Task:**
Identify the diagram structure, text, mathematical notation, relationships, groups, and image placeholders, then output complete draw.io mxGraphModel XML.

**Evaluation Criteria:**
Step 0 — Coordinate System
Treat the image as a 2D plane with origin (0, 0) at the top-left and ({width}, {height}) at the bottom-right. For each object: x = xmin, y = ymin, width = xmax - xmin, height = ymax - ymin. All styles must be written in the `style="..."` attribute of `<mxCell>`; `<mxStyle>` child nodes are not permitted.

Step 1 — Object Detection
Identify all visually distinct objects: blocks, edges, images, curlyBrackets, containers, and group boundaries.

- Blocks: enclosed by a border or distinct background, containing text or images, with a clear geometric shape.
- Edges: thin lines, arrows, polylines, or curves connecting two objects.
- CurlyBrackets: brace shapes enclosing a group of objects.

For each object, estimate a bounding box [xmin, ymin, xmax, ymax] that fully encloses the shape boundary (not just the text inside).

Step 2 — Block Shape Determination
Determine each block's shape using bbox_json:

- polygon (highest priority): Use `shape=mxgraph.basic.polygon` with `polyCoords` taken directly from bbox_json. Use bbox_json's `x, y, width, height` for `<mxGeometry>`. Do not degrade to a rectangle or any other shape.
- rect: Determine the geometric shape from visual appearance — rectangle, rounded rectangle, ellipse, rhombus, etc. Use bbox_json's geometry for `<mxGeometry>`.
- image: Gray "IMAGE" blocks, blocks whose style contains image=, or image-placeholder blocks are stand-ins for source visual content. Image objects should be represented as `shape=image` with the placeholder URL `image=https://placehold.co/WxH?text="IMAGE"` (W and H are the node's integer width and height).

If multiple candidate bounding boxes overlap heavily with nearly identical dimensions and regular offsets, merge them into a single object (layered stack) rather than creating separate nodes.

Step 3 — Grouping
Create a Group or Container when multiple blocks form a coherent whole:

- Shared connection target: If an edge points to a shared area of multiple blocks rather than a single block, create a transparent Group enclosing those blocks (`connectable="1"`), and connect the edge to the Group.
- Visible enclosure: If blocks share a common outer border, background region, dashed frame, or colored area, create a Container whose bounding box fully encloses all inner blocks.
- Regular arrangement with shared cues: If blocks are arranged in a stable pattern (column, row, grid) and share a common title, background, curlyBracket, or edge target, create a Group.
- CurlyBracket group: If a curlyBracket encloses a set of blocks, treat those blocks as a Group. The curlyBracket's `direction` is determined solely by its opening direction: opening down → `south`; up → `north`; left → `west`; right → `east`.
- Semantic group: If blocks clearly constitute the same module, phase, subsystem, or input/output unit, create a transparent Group even without a visible outer border.

In all cases, inner blocks' `parent` must point to the Group, and the Group's bounding box must fully enclose all members.

Step 4 — Text Extraction and Attribution
Extract all recognizable text and bind each segment to exactly one of the following categories based on both position and semantics:

- Block interior text: located inside the block and belongs to its content or title.
- Edge label: near an edge, describing the connection, condition, or transition.
- block_external_label: outside a block but semantically names that single block. Create an outer group with the block and this text cell as children.
- group_external_title: at the title position of a group/container, naming the whole module or phase. Attach as a direct child of the group.
- annotation: independent supplementary text not attached to any block, group, or edge.

Do not leave identified text without a corresponding XML representation. For text outside its parent's bounding box, use relative coordinates (negative values for above/left, values exceeding parent dimensions for below/right).

Step 5 — Mathematical Text
If text contains mathematical symbols, formulas, superscripts, Greek letters, or scientific notation, treat it as mathematical text:
- Convert to valid LaTeX wrapped in `\(...\)` for inline math or `$$...$$` for standalone formulas.
- The node's style must include `html=1;math=1`.
- Use `&#xa;` for line breaks; do not use `<br>`.
- Escape special characters: `<` → `&lt;`, `>` → `&gt;`, `&` → `&amp;`.
- If any mathematical text exists in the diagram, set `math="1"` on the outermost `<mxGraphModel>`.

Step 6 — Relationship Detection (Edges)
Every edge must have both a `source` and a `target`. For each connection:

- Determine which object the arrow departs from and which it points to. A small gap between the arrow and a boundary is acceptable if the direction clearly indicates a connection.
- If the arrow points to a shared area rather than a single block, connect to the corresponding Group or Container, not to an inner child.
- An edge connects to the object at the same hierarchy level as its endpoint; do not descend to a deeper child unless the arrow clearly enters the child's bounding box.
- Path styles: straight line (direct), polyline (`edgeStyle=orthogonalEdgeStyle` with waypoints in `<Array as="points">`), or curve (`curved=1`).
- Connection sides: use `exitX`, `exitY`, `entryX`, `entryY`. Defaults (no `direction`): top = (0.5, 0), bottom = (0.5, 1), left = (0, 0.5), right = (1, 0.5).
- Edge labels go directly into the edge's `value`.

**Output Format:**
Return the complete draw.io XML starting with `<mxGraphModel` and ending with `</mxGraphModel>`. Include all nodes, edges (with waypoints and connection points), and text. Set `pageWidth="{width}"` and `pageHeight="{height}"`. Do not output Markdown code fences, `<mxfile>` wrappers, or standalone `<style>` tags.

```xml
<mxGraphModel dx="1000" dy="1000" grid="1" gridSize="10" guides="1" tooltips="1" connect="1" arrows="1" fold="1" page="1" pageScale="1" pageWidth="{width}" pageHeight="{height}" math="1" shadow="0">
  <root>
    <mxCell id="0"/>
    <mxCell id="1" parent="0"/>
    <!-- nodes and edges -->
  </root>
</mxGraphModel>
```
\end{lstlisting}

\subsection{Parse Critic}
\label{app:parse-critic}
\begin{lstlisting}[style=promptappendix,caption={Parse critic prompt.},label={lst:parse-agent-critic}]
**Role:**
You are a parse validation and correction suggestion agent.

**Inputs:**
You will receive:
1. A mxGraphModel XML
2. A reasons list from a programmatic checker
3. A flowchart image exported from that XML
4. The original flowchart image
5. reasons: {reasons}
6. bbox_json: {bbox_json}

**Task:**
Inspect the XML and rendered image against the original flowchart image, then return whether the parse is acceptable and, if not, provide specific correction suggestions.

**Evaluation Criteria:**
Checks to perform:
1. Verify each block's x, y, width, height against the original image.
2. Check that edge source/target matches the original (which object each arrow connects).
3. Check edge labels for correctness and overlapping.
4. Check that no edges are missing source or target.
5. Check that each edge endpoint connects to the correct object level (Block vs. Group/Container): an endpoint inside a Container but not entering any child's bounding box should connect to the Container, not a child block.
6. Check curlyBracket opening directions against the original image.
7. For every [R2] entry in the reasons list, produce a specific correction.

Pass determination:
- Return `Pass: true` if the reasons list is empty and all checks pass (minor non-readability issues are acceptable).
- Return `Pass: false` for any of: [R2] entries present; obvious block size/position deviations; missing, overlapping, or wrong edge labels; edge source/target inconsistent with the original.

Suggestion writing rules:
- For block corrections: state which block's x, y, width, or height deviates and by how much, or what the corrected coordinates should be.
- For group corrections: state which blocks should be grouped, whether the group should be transparent, which blocks' parent should point to the group, and which edges' source/target should be re-connected to the group.
- For edge corrections: state the correct source and target IDs; if an edge passes through a block it should not, specify a detour path with waypoint coordinates and which side of the block to enter/exit. Prioritize fixing the wrong object-level connection before adjusting the path.
- For [R2] out-of-canvas corrections: state the block ID, its approximate current x/y/width/height, which direction it overflows, and how many pixels to shift (or shrink) with the corrected approximate coordinates. Enlarging the canvas is not permitted.
- For curlyBracket direction: direction is determined only by the opening direction (opening down → south, up → north, left → west, right → east).

**Output Format:**
Return JSON only. No Markdown fences or other text.

```json
{
  "Pass": true,
  "reasons": [],
  "suggestion": ""
}
```

If `Pass` is true, `suggestion` must be an empty string. If `Pass` is false, `suggestion` must be non-empty with specific, actionable fixes for each identified issue.
\end{lstlisting}

\subsection{Style Agent}
\label{app:style-agent}
\begin{lstlisting}[style=promptappendix,caption={Style agent main prompt.},label={lst:style-agent-main}]
**Role:**
You are a draw.io XML style transfer agent.

**Inputs:**
You will receive:
1. The original XML
2. The Style-Reference image
3. The Current render
4. block_color_hints — per-block color sampling hints, each containing: id, geometry, region_in_reference, reference_region_dominant_hex, and the current style's fillColor / strokeColor / fontColor / gradientColor. These are auxiliary hints only; primary judgment must come from visual comparison of the images.
5. The suggestion from the previous critic round — apply every suggestion faithfully.

**Task:**
Without altering structure, text content, or geometry, modify styles in the XML so that the Current render's visual appearance more closely matches the Style-Reference.

**Evaluation Criteria:**
Colors
- Use the visually corresponding region in the Style-Reference as the primary basis. Correct fillColor / strokeColor / fontColor / gradientColor to match.
- You may refer to `reference_region_dominant_hex` in block_color_hints as a starting point, but always verify against the image.
- If the reference region appears to be primarily text or lines rather than a filled area, do not force that color into fillColor — correct the corresponding fontColor or strokeColor instead.
- Modify gradientColor only when the Style-Reference clearly shows a gradient effect. Do not introduce gradients not visible in the reference.
- Only suggest a color change when the difference is clearly visible; do not adjust colors that are already close.

Connector and arrow styles
- Check that connectors which should be plain lines are not mistakenly styled as arrows, and that arrows appear where the reference shows them.
- You may adjust: endArrow, startArrow, strokeWidth, dashed, curved, edgeStyle, elbow, strokeColor.
- Modify styles only — do not alter connector direction logic, exit side, entry side, or routing path.

Text and shape styles
- Correct fontColor to match the reference. This is the highest-priority text change.
- Check whether the reference uses bold or regular weight and set fontStyle accordingly. Compare carefully before suggesting a change.
- If fontSize is too large and causes text to overflow the block border, reduce it. Increasing fontSize is strictly prohibited.
- Change fontFamily only when the difference is clearly visible (e.g., the reference is clearly serif while the current is sans-serif).
- If text content in the current XML differs from the reference, you may correct it; otherwise never modify text content.

**Constraints:**
- Do not modify mxCell id, value, source/target.
- Do not modify mxGeometry x, y, width, height.
- Do not modify canvas size (pageWidth / pageHeight).
- Do not add or delete nodes or edges.
- Do not modify connector direction, exit side, entry side, or routing path.
- Gray "IMAGE" blocks, blocks whose style contains image=, or image-placeholder blocks are stand-ins for source visual content. Do not modify styles on image-placeholder blocks.

**Output Format:**
Return only a valid mxGraphModel XML. Do not add any explanation, prefix, suffix, Markdown fence, JSON, heading, or commentary. The very first character of your reply must be `<` and the reply must contain only XML.
\end{lstlisting}

\subsection{Style Critic}
\label{app:style-critic}
\begin{lstlisting}[style=promptappendix,caption={Style critic prompt.},label={lst:style-agent-critic}]
**Role:**
You are a strict visual style critic for draw.io renders.

**Inputs:**
You will receive:
1. Current (the latest render)
2. Style-Reference (the target)
3. block_color_hints — per-block: id, geometry, reference_region_dominant_hex, and the current style's fillColor / strokeColor / fontColor / gradientColor. Auxiliary hints only; primary judgment must come from visual comparison.

**Task:**
Compare the visual differences between Current and Style-Reference. Check style only — do not check whether text content has been altered or whether geometry has been re-arranged.

**Evaluation Criteria:**
Colors
- Check whether fillColor / strokeColor / fontColor / gradientColor are appropriate for each block.
- Only suggest a change when the difference is clearly visible in the images. Do not flag minor color shifts that are barely perceptible.
- You may cite reference_region_dominant_hex to support your suggestion.
- Specify whether to change fillColor, strokeColor, fontColor, or gradientColor — and in which direction (lighter, darker, more saturated, different hue, etc.).

Connector and arrow styles
- Check whether plain line segments have been mistakenly changed to arrows, or whether arrows are missing where the reference shows them.
- Check endArrow, startArrow, strokeWidth, dashed, curved, edgeStyle, elbow, strokeColor.
- Style only — do not request changes to connector direction, routing path, exit side, or entry side.

Text and shape styles
- Check font size, font family, font weight, and text color against the reference.
- If fontSize is too large and causes text to overflow the block border, suggest reducing it. Increasing fontSize is strictly prohibited.
- If the reference appears bolder, say to increase font weight; if lighter, say to remove bold. Compare carefully before suggesting.
- If the reference font is clearly different in style (e.g., serif vs. sans-serif), suggest changing fontFamily to match.

**Constraints:**
- Return JSON only and follow the requested schema exactly.
- Gray "IMAGE" blocks, blocks whose style contains image=, or image-placeholder blocks are stand-ins for source visual content. Do not request style changes to image-placeholder blocks.

Pass determination:
- Return `Pass: true` if the Current render already closely matches the Style-Reference in style and only negligible differences remain.
- Return `Pass: false` if clearly visible style differences exist in any of the three dimensions above.

**Output Format:**
Return JSON only. No Markdown fences or other text.

```json
{
  "Pass": true,
  "reasons": [],
  "suggestion": ""
}
```

`reasons` is always `[]`. If `Pass` is true, `suggestion` must be `""`. If `Pass` is false, `suggestion` must be a multi-line string with each line starting with `"- "`. Each line must specifically identify: which object (by name, text content, or position), which style attribute to change, and in which direction. For example:
- "- Set the top-left container's fillColor to more closely match the reference region dominant color."
- "- Reduce the right-side connector's strokeWidth to match the thinner black line in the reference."
- "- Change the bottom-right title's fontStyle from bold to regular weight, as the reference does not appear that bold."
\end{lstlisting}

\subsection{Layout Agent}
\label{app:layout-agent}
\begin{lstlisting}[style=promptappendix,caption={Layout agent main prompt.},label={lst:layout-agent-main}]
**Role:**
You are a flowchart layout agent.

**Inputs:**
You will receive a complete mxGraphModel XML (openable in draw.io). Canvas size is given by `pageWidth` / `pageHeight`, or overridden by the user. Blocks are `mxCell vertex="1"` nodes; edges are `mxCell edge="1"` nodes.

**Task:**
Given a mxGraphModel XML, compute new positions (x, y) and sizes (w, h) for every block on the specified canvas and output the re-laid-out mxGraphModel XML. Use only block IDs, geometry (w, h as size ratio reference), and edge source/target for layout decisions.

**Constraints:**
Sizing and Proportions
- Blocks that are larger in the original must receive proportionally larger space in the new layout (preserve relative area ratios as closely as possible).
- Gray "IMAGE" blocks, blocks whose style contains image=, or image-placeholder blocks are stand-ins for source visual content. Preserve the original aspect ratio for image-placeholder blocks.

Placement and Flow
- Blocks connected by an edge must be placed adjacent so that their connecting arrows do not overlap other elements.
- Use the full canvas — avoid excessive whitespace at any edge; distribute blocks to fill open areas.
- Only avoid overlaps that are newly introduced. If two blocks overlapped in the original, they may remain overlapping; maintain a minimum gap of at least the shorter side of the smaller block between all other pairs.
- Coordinate system: origin (0, 0) at top-left; x increases rightward, y increases downward.
- All input block IDs must appear in the output unchanged.
- Directional centering: if a block is primarily approached from left/right, its row must be centered vertically; if from top/bottom, its column must be centered horizontally.
- Overall centering: the bounding-box center of all blocks must align approximately with the canvas center.

Container and Group Arrangement
- If a large container primarily holds multiple child objects rather than acting as a single node, its width/height orientation may be changed to suit the new layout (e.g., a vertical container may become horizontal). Choose the orientation that best fits the overall flow direction, neatens internal elements, reduces long connectors, and improves canvas utilization.
- If a container changes position, orientation, or dimensions, all closely attached objects (intermediate nodes, decorative connectors, primary entry/exit lines) must be repositioned accordingly. Keep connectors short, direct, and aligned with the main flow.
- The internal positions, order, column count, and row count of child blocks within a container may be freely re-arranged, as long as they remain within the container. Prefer the arrangement that is most organized, space-efficient, and aligned with the primary reading direction.
- If a block's primary semantic role is a dataset, method collection, or result collection rather than a single indivisible entity, left/right or top/bottom order of internal elements may be swapped, and a single column may be changed to a single row (or vice versa), to improve overall flow clarity and canvas utilization.
- If a curlyBracket encloses a group of blocks, treat those blocks as a local group. CurlyBracket direction is determined only by its opening direction: opening down → `south`; up → `north`; left → `west`; right → `east`.

Whitespace and Density
- Actively reduce local large gaps: if related blocks still have excessive whitespace between them, pull them closer, scale them up, or redistribute rows/columns to fill it. Do not merely center the outer bounding box while leaving local groups internally sparse.
- If the canvas is large, do not simply spread small blocks further apart — also consider enlarging blocks to keep local areas full.
- Form compact, even, visually continuous clusters for connected components or co-module blocks.

Canvas-Ratio Layout Reference
- Let r = canvas width / height.
- r >= 2.0 (strongly landscape): primary direction left-to-right; prefer horizontal arrangement with multiple rows.
- 1.25 <= r < 2.0 (mildly landscape): primarily left-to-right; 1-2 row horizontal mix.
- 0.8 <= r < 1.25 (near square): prefer grid/layered arrangement (e.g., 2x2, 3+2).
- 0.5 <= r < 0.8 (mildly portrait): primarily top-to-bottom; 1-2 column vertical mix.
- r < 0.5 (strongly portrait): primary direction top-to-bottom; prefer vertical arrangement with multiple columns.

Topological Layering
- Build a DAG from edges. Compute layer(v) = 1 + max(layer(u)) over all predecessors u; nodes with no predecessors have layer 0. Nodes with smaller layer values must appear earlier along the primary flow direction.

Arrow Routing
- Arrows connect to the left/right midpoint for horizontal flow, top/bottom midpoint for vertical flow.
- If a block has multiple incoming or outgoing edges, spread attachment points evenly along that side (10-20% of side length per step), centered around the midpoint.
- Prefer a single diagonal straight line from source to target; use minimal waypoints only to avoid passing through other blocks.
- All arrow control points must remain within the canvas bounds.

**Output Format:**
Return only the re-laid-out mxGraphModel XML — no explanations, no Markdown. Retain all original mxCells (style, value, parent, edge, etc.). Only update `mxGeometry` for `vertex="1"` cells. Set `pageWidth` / `pageHeight` to the user-specified canvas size. Do not alter any edge's source or target.
\end{lstlisting}

\subsection{Deterministic Layout Validation Rules}
\label{app:layout-rules}
\begin{lstlisting}[style=promptappendix,caption={Deterministic layout validation rules.},label={lst:layout-deterministic-rules}]
**Deterministic Layout Validation Rules:**
The layout stage first applies deterministic checks before the VLM critic. These checks generate a reasons list that the critic receives as input.
- R1 — Basic format validity: the mxGraphModel must define pageWidth and pageHeight, contain at least one vertex block, and each block must have an id and geometry fields x, y, width, and height.
- R2 — Canvas containment: every non-decoration block must have valid geometry and remain inside the target canvas.
- R3 — Same-level overlap avoidance: blocks at the same hierarchy level must not newly overlap, except for overlaps that already existed in the input XML.
- R4 — Canvas usage and centering: the layout must use enough horizontal and vertical span, avoid excessive margins on any side, and remain approximately centered on the canvas.
- R5 — Relative size preservation: block area ratios should remain close to the input layout, and image-placeholder blocks must preserve their original aspect ratio.
- R6 — Relationship preservation: edges must reference existing source and target blocks, and the output must not add or remove source-target relationships relative to the input XML.
- R7 — Edge control-point containment: edge waypoint/control points must remain inside the target canvas.
\end{lstlisting}

\subsection{Layout Critic}
\label{app:layout-critic}
\begin{lstlisting}[style=promptappendix,caption={Layout critic prompt.},label={lst:layout-agent-validator}]
**Role:**
You are a layout validation and suggestion agent.

**Inputs:**
You will receive:
1. A mxGraphModel XML
2. A reasons list generated by a programmatic checker
3. A flowchart image exported from that XML

**Task:**
Inspect the image for obvious visual problems and give a list of precise, actionable suggestions referencing the reasons list and XML. Do not regenerate reasons.

**Evaluation Criteria:**
Pass determination:
- Return `Pass: true` if the reasons list is empty and the overall layout, arrows, and reading flow are reasonable (minor improvable issues are acceptable).
- Return `Pass: false` if the image shows obvious problems: edges passing through blocks, tangled edges in the core area, chaotic flow direction, groups split apart, severely unbalanced node distribution, or poor text readability.

Visual problems to check:
- Edges passing through or hugging block borders
- Edges crossing in the central area or loop edges routed unnecessarily far out
- Unreasonable relationship arrangement (adjacent steps placed far apart, direct connections routed via long detours)
- An edge from a bracket or group border that first reverses direction before joining the main flow
- CurlyBrackets no longer clearly enclosing their intended group, or wrong opening direction
- Related nodes placed too far apart; obvious unbalanced distribution (left-heavy, top-crowded, corner-concentrated)
- Group titles clipped, covered, or overlapped by arrows
- Inner blocks of a group clearly off-center within the group
- Block text truncated or touching borders
- Large unused whitespace within a group while another side is overcrowded
- Unnecessarily long diagonal lines that could be shortened by repositioning nodes
- Tall-thin or flat-wide containers whose aspect ratio is unsuitable for the current canvas and flow direction

Suggestion writing rules:
- Address each issue from the reasons list and each visible image problem with a concrete fix.
- Specify exactly which nodes should move closer, swap left/right or top/bottom order, or align into the same row or column.
- If the main problem is readability rather than a hard deterministic error, describe it explicitly; do not return `Pass: true` only because the reasons list is empty.
- If a group title touches the border or is clipped, suggest adding top or left padding to that group, or moving the top row of nodes downward to reserve safe title space.
- If a large block or group has internal blocks and titles that are not centered, name the group, state whether the internal content is shifted left, right, up, or down, and suggest translating the local sublayout back toward the group center. If needed, suggest a more symmetric row or column arrangement without breaking the local flow direction.
- If an arrow crosses a group title or title-reserved area, identify the arrow and group title, then suggest routing it below the title region or around the outside of the group. If moving connected nodes would reduce the crossing, state that too.
- If a block extends outside the canvas, suggest shrinking that block or shrinking blocks in the same group. Do not suggest changing the canvas size.
- If nodes inside a group break the original local flow, identify which nodes should return to an upstream-to-downstream order, either top-to-bottom or left-to-right.
- If a poorly placed node creates a long diagonal edge, suggest moving that node closer to its main source or target, prioritizing shorter cross-group lines and fewer backtracking segments.
- If the relationship arrangement is unreasonable, identify which nodes are too far apart or arranged in an unnatural flow direction. Suggest moving them closer, aligning them in the same row or column, or reordering them to better match the main flow. Treat an edge path that is much longer than necessary as an improvable issue even if it does not cross a block.
- If an edge starts from a bracket, annotation, or group-border object and first reverses direction, slides along the source boundary, or takes an unnatural first segment, identify the edge, source object, and target. Suggest exiting from the side of the source closest to the target and shortening the first segment.
- If the diagram contains a curlyBracket, determine its direction only from the opening direction, not from the tip, protrusion, stem, or outer side. Verify that the enclosed objects lie on the side faced by the opening: opening down = `south`, opening up = `north`, opening left = `west`, opening right = `east`. If the bracket encloses the correct group but has the wrong direction value, state the corrected value using only opening-direction language.
- If an edge passes through a block, identify which edge segment should become a polyline or slight curve, state whether it should detour above, below, left, or right, and explain why that direction is more reasonable.
- If there is a loop edge, state which outside side it should route around to avoid crossing the main reading area.
- If related nodes are too far apart or a group is loose, identify which nodes should be compacted into a local group.
- If a local region has large empty gaps, identify which related blocks are too sparse and which whitespace can be filled. Suggest pulling blocks closer, enlarging blocks, or re-arranging them into a tighter row, column, or small grid. The goal is a compact, even, continuous local module without breaking the main flow. Container dimensions and internal object placement may both be adjusted to fill gaps.
- If horizontal or vertical span is insufficient, state whether the layout underuses horizontal space, vertical space, or both. Suggest moderately enlarging related blocks or spreading same-row, same-column, or same-group blocks in that direction while preserving the original reading flow and group relationships.
- If a long block or group frame is actually a container rather than a direction-fixed node, you may suggest changing its orientation or aspect ratio, such as vertical to horizontal or overly flat to taller. Also state how its internal blocks should be rearranged; do not change only the outer frame. For vertical crowding, consider making a tall narrow container more horizontal; for horizontal crowding, consider making an overly wide container more vertical.

**Output Format:**
Return JSON only. No Markdown fences or other text.

```json
{
  "Pass": true,
  "reasons": [],
  "suggestion": []
}
```

If `Pass` is true, `suggestion` must be an empty list. If `Pass` is false, `suggestion` must be a non-empty list of strings, with one specific, actionable fix per identified problem.
\end{lstlisting}

\section{Benchmark Evaluation Prompts}
\label{app:benchmark-prompts}

This section reproduces the prompts used by the four VLM-as-a-Judge evaluators described in the Experimental Setup of the main paper, covering Relationship Preservation (Appendix~\ref{app:bench-relationship}), Hallucination-free Rate (Appendix~\ref{app:bench-hallucination}), Layout Quality (Appendix~\ref{app:bench-layout}), and Style Similarity (Appendix~\ref{app:bench-style}). Minor editorial adjustments were made for readability and presentation consistency.

\subsection{Relationship Benchmark}
\label{app:bench-relationship}
\begin{lstlisting}[style=promptappendix,caption={Relationship benchmark evaluation prompt.},label={lst:bench-relationship}]
**Role:**
You are a flowchart relationship comparator.

**Inputs:**
You will be given:
1. The reference diagram (original)
2. The candidate diagram (to be checked)

**Task:**
Evaluate only whether the relationship / structural connections in the candidate are consistent with the reference. Do NOT judge a mismatch based on any of the following differences alone:
- Different layout positions
- Different sizes
- Different colors
- Different alignment
- Different visual style
- Differences in arrow/line style — if a connection exists where one should be, that counts as consistent regardless of style
- The same set of objects drawn more spread out or more compact

**Evaluation Criteria:**

Step 1 — Establish Object Correspondence
Before checking relationships, first determine which blocks, groups, and modules in the candidate correspond to which objects in the reference, and understand the overall logical flow the diagram is meant to express. Only after establishing this correspondence should you check whether relationships are consistent.

Step 2 — Judge Conservatively
Only flag a relationship as inconsistent when the discrepancy is both "clearly visible and certain" AND "logically inconsistent with the reference."
- If a line is too thin, resolution is too low, a region looks locally blurry, or it's unclear whether an arrow is connected, default to treating it as consistent — do not over-interpret.
- If there are extra or missing arrows but the overall connection logic and the meaning the diagram is trying to express remain the same as the reference, treat it as consistent.
- If extra or missing arrows cause the overall process logic to change — e.g., the flow direction is completely reversed, or a model's input gains something not present in the reference — this can be judged inconsistent.

Step 3 — Apply the Inconsistency Rules
Only the following situations may be judged as inconsistent:
- A clearly extra arrow or connection not present in the reference
- A clearly missing arrow or connection that exists in the reference
- A connection clearly linked to the wrong module
- A clearly reversed flow direction

**Output Format:**
Return JSON only. No Markdown fences or other text.
```json
{
  "correct": true,
  "differences": []
}
```
or
```json
{
  "correct": false,
  "differences": [
    "Specifically describe which relationship is inconsistent with the reference"
  ]
}
```
\end{lstlisting}

\subsection{Hallucination Benchmark}
\label{app:bench-hallucination}
\begin{lstlisting}[style=promptappendix,caption={Hallucination benchmark evaluation prompt.},label={lst:bench-hallucination}]
**Role:**
You are a content integrity auditor for flowchart diagrams.

**Inputs:**
You will be given:
1. The reference diagram (original)
2. The candidate diagram (to be checked)

**Task:**
Compare the two images and determine whether the pipeline output has introduced hallucinated content. Only evaluate content-completeness issues: whether the candidate adds text or images not present in the reference, is missing important blocks/modules from the reference, or introduces extra flowchart blocks that don't exist in the reference. Do NOT evaluate visual presentation differences — overall layout, arrow count, arrow direction, connection logic, font, color, size, border style, or design style are never grounds for flagging hallucination, even if they differ from the reference.

**Evaluation Criteria:**

1. **Image Check**
   - Scope: photos, illustrations, and charts within the flowchart.
   - Flag as hallucination: the candidate introduces image content that has no counterpart in, or is unrelated to, the reference.
   - Do NOT flag: gray "IMAGE" blocks, gray image-icon placeholder blocks, cropping or resolution differences, differences in color depth, differences in size or position, or vague visual elements whose subject can't be clearly identified.

2. **Flowchart Structure Check**
   - Scope: blocks, modules, containers, and groups in the flowchart, including the count and type of elements within each group.
   - Flag as hallucination: the candidate adds multiple blocks/modules that don't exist in the reference; the candidate is missing clearly present and important blocks/modules from the reference; or existing reference blocks are excessively duplicated, resulting in an unreasonable count.
   - Do NOT flag: differences in block position, size, spacing, alignment, shape detail, border style, arrow count, arrow direction, connection logic, group appearance, or overall layout.

3. **Text Check**
   - Scope: descriptive/label text within the flowchart.
   - Flag as hallucination ONLY when the candidate contains a substantial block of text that is entirely unrelated to the reference, or is missing an entire substantial passage of text present in the reference.
   - Do NOT flag: whitespace, punctuation, capitalization, font, font size, color, rotation, line-break/re-wrapping differences, sub/superscripts, math-notation-related differences, blurry/hard-to-read text, minor truncation, text overlapping with lines, single-character errors, or minor text differences that don't affect the core information.

**Output Format:**
Return JSON only. No Markdown fences or other text.
```json
{
  "correct": true,
  "differences": []
}
```
or
```json
{
  "correct": false,
  "differences": ["describe which text, images, or blocks appear in the candidate but not in the reference, or are missing from the candidate but present in the reference"]
}
```

**Before finalizing your answer, double-check:**
- If your answer is `false`: re-examine the reason. If the only issues are minor text errors, typos, small (non-substantial) text differences, or a few missing minor labels/titles (not a large block), change the answer to `correct: true`.
- If your answer is `false`: re-examine the reason. If the only issues are connection errors, logic errors, or arrow errors, change the answer to `correct: true`.
- If your answer is `true`: re-check whether every image appearing in the candidate is identical to the corresponding image in the reference. If any image differs in content, has been compressed, rotated, is new, or has otherwise been altered, change the answer to `correct: false`.
\end{lstlisting}

\subsection{Layout Benchmark}
\label{app:bench-layout}
\begin{lstlisting}[style=promptappendix,caption={Layout benchmark evaluation prompt.},label={lst:bench-layout}]
**Role:**
You are a "Flowchart Relayout Layout Evaluator." Your task is limited to analyzing block placement, arrow direction, and overall layout.

**Inputs:**
- Image 0 is the original flowchart.
- If severe-error reference images are provided, these only illustrate typical severe layout error scenarios — they are NOT candidates and must never be placed in the ranking.
- Images 1..N are candidate re-layout results. Each candidate re-lays out the original flowchart onto a new canvas. You must rank the candidates by how well each one lays out the original content on the new canvas.
- Gray boxes labeled "IMAGE" in a diagram represent a photo crop from the original — never treat these as a flaw.

**Please Note:**
You should ONLY evaluate layout — meaning block placement, flowchart flow direction, canvas utilization, visual balance, and alignment.
Do NOT evaluate image/icon content, text content correctness, aesthetic quality, font style, or visual design style. The only scoring criterion is whether the candidate has designed a reasonable relayout on the new canvas based on Image 0's blocks — nothing else should factor into the evaluation.

Your task has two steps.

**Step 1: Identify which candidates have severe layout errors**

Every candidate must output the following three fields under `severe_layout_errors`. Each field's value can only be `true` or `false`, and each field name must be a single letter:

1. **a** — This candidate is oriented incorrectly: it is not directly readable as-is and must be flipped or rotated to be read. If the candidate's `is_ocr_rotated_90_270` is `true`, this candidate requires rotation to be read properly, so `a` must be `true`; otherwise `false`.
2. **b** — This candidate has excessively large blank areas and does not make full use of the canvas. If one or more sides (top, bottom, left, right) of the image show clearly large amounts of blank space, giving the clear impression that the full canvas isn't being used, or if the input's `empty_area_ratio` is greater than 0.6, output `true`; otherwise `false`.
3. **c** — This candidate's block placement and arrow positions are identical to the Image 0 reference, and it has simply been stretched or scaled into a new aspect ratio without any real relayout. If the input's `direct_resize_lpips` is less than 0.5, or the candidate otherwise matches this pattern of pure stretching/scaling without genuine relayout, output `true`; otherwise `false`.

If any of a candidate's `a`, `b`, or `c` fields is `true`, that candidate has a severe layout error — but it must still appear in the ranking.

**Step 2: Rank all candidates**

The most important factor in ranking is whether the overall layout is usable as a reference — that is, whether it has reorganized Image 0's main block relationships and flow structure onto the new canvas into a clear, reasonable, readable overall layout. Arrow style, font size, and how loosely/tightly elements are arranged can be used as secondary observations, but must not dominate the ranking — the primary basis remains overall layout, flow direction, alignment, canvas utilization, and visual balance.

If two or more candidates are very close in severity of layout errors and overall layout quality, you do not need to force a distinction — you may assign them the same rank.

Lower rank numbers are better. When candidates tie, the next rank must skip the occupied position(s) — e.g., two tied for 1st: `1, 1, 3`; two tied for 2nd: `1, 2, 2, 4`.

`descriptions` must describe every candidate, and every `candidate_index` must have exactly one description entry.

Judge severity in the order a, b, c: candidates with `a = true` should be ranked last overall; candidates with `b = true` should rank after ordinary candidates but generally ahead of candidates with `a = true`; candidates with `c = true` should rank after candidates with genuine relayout, but generally ahead of candidates with `a` or `b` true.

For all remaining candidates, consider whether the deduction criteria below apply, and rank by the number and severity of such issues. If several candidates show no clear severe errors, or the remaining candidates have similar error levels, they may be given the same rank.

**Primary Deduction Criteria:**

1. **Flow Arrangement**
   - The flowchart does not form a clear, easily readable primary flow direction — looking at the diagram's major components and overall process, arrow flow lacks consistency.
   - Reading requires frequently switching between looking left-right and up-down, making the flow feel disjointed and hard to follow in one consistent direction.

2. **Partially Flipped Components**
   - Part of the canvas contains components whose text and structure clearly need to be flipped to be read. Severity can be judged by the proportion of the canvas this affects.

3. **Space Utilization and Visual Balance**
   - The flowchart is overly concentrated in one part of the canvas, leaving large blank areas elsewhere — e.g., the bottom-left is mostly background color while the flowchart is concentrated in the upper-right.
   - If block groups or the overall flow could clearly be centered, aligned, or more evenly distributed across the available space, but the candidate fails to do so — leaving excess blank space on one side or elsewhere — this should also count as a visual balance / space utilization flaw.
   - If two candidates' `empty_area_ratio` values differ by more than 0.2, this may be treated as a meaningful difference in space utilization and used as a ranking basis.

**Reason / Description Writing Requirements:**
- All descriptions must be written in Chinese.
- `severe_layout_errors` must NOT output a `reason` field — only `candidate_index`, `a`, `b`, `c`.
- `a`, `b`, `c` may only be set based on the three severe-error definitions in Step 1 — never mark `true` for any other reason.
- Descriptions must explicitly map to the layout criteria above — e.g., arrow arrangement, space utilization, visual balance, alignment, canvas fit.
- Do not introduce scoring criteria not covered in these guidelines.
- Do not judge quality based on image content, gray "IMAGE" placeholder blocks, text content, color, or visual style.
- If a candidate has flaws but none of the three severe-error conditions apply, `a`, `b`, and `c` must all be `false` — the candidate must still remain in the ranking, with its layout flaws noted only in `descriptions`.

**Output Rules:**
- Output valid JSON only.
- Do not output Markdown.
- Do not output any text outside the JSON.
- `candidate_index` must use the image number: Image 1 → 1, Image 2 → 2, and so on.
- `severe_layout_errors` must include every candidate, and each entry must have all four fields: `candidate_index`, `a`, `b`, `c`.
- `a`, `b`, `c` must be output as JSON booleans (`true`/`false`) — never as strings.
- `ranking` must include every candidate, ordered from best to worst layout quality, with ties allowed.
- `descriptions` must include every candidate; order should match `ranking`.
- Do not output any field indicating a candidate was excluded from ranking.

**Output format:** Valid JSON only, no Markdown, nothing outside the JSON. `candidate_index` = image number. `a`/`b`/`c` are JSON booleans.

{
  "severe_layout_errors": [
    { "candidate_index": 1, "a": false, "b": false, "c": false }
  ],
  "ranking": [
    { "candidate_index": 1, "rank": 1 }
  ],
  "descriptions": [
    { "candidate_index": 1, "description": "" }
  ],
  "overall_summary": ""
}
\end{lstlisting}

\subsection{Style Benchmark}
\label{app:bench-style}
\begin{lstlisting}[style=promptappendix,caption={Style benchmark evaluation prompt.},label={lst:bench-style}]
**Role:**
You are a flowchart visual style evaluation expert. Your task is limited to evaluating the Style of the flowchart — whether the candidate diagram looks like it was drawn by the same author as the original, preserving the same visual style.

**Inputs:**
You will be given a set of images:
- Image 0 is the original flowchart (reference).
- Images 1..N are candidate re-layout results, referred to as Candidate 1, Candidate 2, Candidate 3, Candidate 4, Candidate 5.

**Special Notes:**
- Gray boxes labeled "IMAGE" in a diagram represent a photo/image crop from the original diagram. Do NOT judge these as a Style mismatch — always treat them as if they contain the exact same photo content as in Image 0. This rule applies at every stage of evaluation.
- You should ONLY evaluate whether Style matches the reference — meaning only Style-related elements: object color, font, line style, shape proportions, and the overall rotation orientation of the diagram.
- Do NOT evaluate icon/text content correctness, connector direction or logic, overall layout, or other non-Style elements. Things like incorrect text content, wrong connections, incorrect element geometry/size, or logical errors in the diagram must never be used as scoring grounds.

**Evaluation Process:**
First observe each candidate individually. Based on the Style evaluation dimensions below, determine which candidates are closest to the reference in Style and which are most inconsistent, then rank and explain.

**Evaluation Dimensions:**
1. Background color — whether the overall background color matches the reference.
2. Blocks:
   - Block fill color
   - Block border style (e.g. solid, dashed, border color)
   - Whether a gradient is present
3. Typography:
   - Text color
   - Text style (font family, bold, italic, etc.)
   - Font weight
4. Connector style:
   - Edge color
   - Edge style (dashed or solid)
   - Arrowhead style (e.g. arrow, dot, none, thickness)

Judge consistency with the reference across all the above dimensions, and rank accordingly.

**Ranking and Deduction Rules:**
- Weight deductions by the canvas area proportion of the erroneous element — the larger the proportion of the canvas an inconsistent element occupies, the greater the deduction.
- If two candidates have similar error proportions and similar severity, treat them as tied: assign the same rank and skip the next rank accordingly (e.g. with 5 candidates, if two tie for 1st, ranks should be 1, 1, 3, 4, 5).

**Key Exclusions (do not deduct for):**
- All icons/images replaced with gray "IMAGE" blocks — this is a special case and must never be a deduction basis.
- Differences between a candidate's gray placeholder block and the reference's actual photo/icon — always treat the placeholder as identical in content to Image 0.
- Any gray block labeled "IMAGE" should be treated immediately as equivalent to the reference's photo content; never deduct for it, and always evaluate Style as if it were the same photo as in Image 0.
- Do not mention any non-Style-related scoring criteria in the description — e.g. content errors, logic errors, connector errors, layout errors, or gray "IMAGE" placeholder blocks.
- Do not compare the reference's actual image against a candidate's gray "IMAGE" placeholder — always treat them as identical.
- Image/diagram rotation orientation is out of scope — never deduct for it.
- Incorrect text content or wrong connections are out of scope — never deduct for them.

**Important:** If multiple candidates score identically across every dimension, treat them as tied and give them the same rank.

**Description Writing Requirements:**
- All descriptions must be written in Chinese.
- Each description must explicitly map to the evaluation dimensions above (e.g. block color, shape errors, etc.).
- Do not introduce scoring criteria that aren't part of these guidelines.
- Do not judge quality based on a candidate's gray "IMAGE" placeholder blocks, content errors, logic errors, or connector errors — none of these are Style-related.

**Overall Summary Writing Requirements:**
- overall_summary must be written in Chinese.
- overall_summary should identify which Style elements were the key differentiators (e.g. color, font, line style).
- Do not mention any non-Style-related scoring criteria, such as content errors, logic errors, connector errors, or layout errors.
- Specifically state which candidates are closest to the reference in Style, which are most inconsistent, and the specific Style-element differences that led to these judgments.

**Output Rules:**
- Output valid JSON only.
- Do not output Markdown.
- Do not output any text outside the JSON.
- candidate_index must use the image number: Image 1 → 1, Image 2 → 2, and so on.
- ranking must be a number from 1 to N, ordered from best to worst Style quality. Tied candidates receive the same rank, and the next rank is skipped accordingly (e.g. with 5 candidates, if two tie for 1st, ranks should be 1, 1, 3, 4, 5).
- The order of entries in descriptions must match the ranking order.

Strictly use the following JSON format:

Return only:
{
  "descriptions": [
    {
      "candidate_index": 1,
      "ranking": 1,
      "description": ""
    },
    {
      "candidate_index": 3,
      "ranking": 1,
      "description": ""
    },
    {
      "candidate_index": 4,
      "ranking": 3,
      "description": ""
    }
  ],
  "overall_summary": "..."
}
\end{lstlisting}

\end{document}